\documentclass[12pt]{article}
\usepackage[a4paper,margin=1in]{geometry}

\usepackage[utf8]{inputenc}
\usepackage{authblk}
\usepackage{bbold}
\usepackage{latexsym,amsmath,amscd,amssymb,graphics}
\usepackage{enumerate}
\usepackage{wrapfig}
\usepackage{graphicx}
\usepackage{framed}
\usepackage{color}
\usepackage{comment}
\usepackage[sort&compress]{natbib}
\usepackage{url}

\usepackage[all]{xy}

\makeatletter

\newtheorem{theorem}{Theorem}

\newtheorem{definition}[theorem]{Definition}

\newtheorem{lemma}[theorem]{Lemma}

\newtheorem{remark}[theorem]{Remark}

\numberwithin{theorem}{section}

\def\be{\begin{equation}}
\def\ee{\end{equation}}
\def\bea{\begin{eqnarray}}
\def\eea{\end{eqnarray}}
\def\ba{\begin{array}}
\def\ea{\end{array}}

\def\bp{\mathbf{p}}
\def\bx{\mathbf{x}}

\newcommand{\rem}[1]{}

\newcommand{\bq}{\boldsymbol{q}}
\newcommand{\bu}{\boldsymbol{u}}

\newcommand{\bmu}{\boldsymbol{\mu}}
\newcommand{\bPi}{\boldsymbol{\Pi}}

\newcommand{\pp}[2]{\frac{\partial #1}{\partial #2}}

\newcommand{\mso}{\mathfrak{so}}

\title{Structure-preserving learning and prediction in optimal control of collective motion}
\author[1]{Sofiia Huraka; \thanks{email: huraka@ualberta.ca}}
\author[1,2]{Vakhtang Putkaradze \thanks{Corresponding author, email: vputkaradze@ua.edu}} 
\affil[1]{Department of Mathematical and Statistical Sciences, University of Alberta, 
Edmonton, AB, T6G 2G1 Canada} 
\affil[2]{Department of Mathematics, University of Alabama, 
Box 870350
Tuscaloosa, AL 35487-0350 (Permanent address)
} 
\date{\today}
\begin{document}
\maketitle

\abstract{
Wide-spread adoption of unmanned vehicle technologies requires the ability to predict the motion of the combined vehicle operation from observations. While the general prediction of such motion for an arbitrary control mechanism is difficult, for a particular choice of control, the dynamics reduces to the Lie-Poisson equations \cite{justh2010extremal,justh2015optimality}. Our goal is to learn the phase-space dynamics and predict the motion solely from observations, without any knowledge of the control Hamiltonian or the nature of interaction between vehicles. 
To achieve that goal, we propose the \emph{Control Optimal Lie-Poisson Neural Networks (CO-LPNets)} for learning and predicting the dynamics of the system from data. Our methods learn the mapping of the phase space through the composition of Poisson maps, which are obtained as flows from Hamiltonians that could be integrated explicitly. CO-LPNets preserve the Poisson bracket and thus preserve Casimirs to machine precision. We discuss the completeness of the derived neural networks and their efficiency in approximating the dynamics. To illustrate the power of the method, we apply these techniques to systems of $N=3$ particles evolving on ${\rm SO}(3)$ group, which describe coupled rigid bodies rotating about their center of mass, and  ${\rm SE}(3)$ group, applicable to the movement of unmanned air and water vehicles. 
Numerical results demonstrate that CO-LPNets learn the dynamics in phase space from data points and reproduce trajectories, with good accuracy, over hundreds of time steps. The method uses a limited number of points ($\sim200$/dimension) and parameters ($\sim 1000$ in our case), demonstrating potential for practical applications and edge deployment. 
}
\section*{List of Main Notations}
\begin{tabular}{lp{10cm}}
  \textbf{Symbol} & \textbf{Description} \\
  $G$    & Lie group on which an individual particle evolves \\
    $\mathfrak{g}$     & Lie algebra of that Lie group\\
$\Gamma^s_{ij}$    &  Structure constants of the Lie algebra\\
  $N$ & Total number of interacting particles (vehicles) \\
    $n$ & Dimension of the Lie group/algebra \\
        $m$ & Number of controls \\
  $\mu_k$    &  Symmetry reduced co-state (momentum), $\mu_k \in \mathfrak{g}$\\
  $\mu$    &  Total Symmetry reduced co-state (momentum), $\mu = (\mu_1, \ldots, \mu_N) \in \mathfrak{g}$\\
    $\bmu_k$    &  A vector representing all components of momenta in a given basis for each $k=1, \ldots N$\\
        $\mu_{ki}$    &  The $i$-th component of vector $\bmu_k$\\
$\widehat{\mu}_k$ & An antisymmetric matrix composed from components $\mu_{ki}$ with components $\alpha$, $\beta$ given as 
$\widehat{\mu}_{k, \alpha, \beta} = - \sum_{s=1}^{n} \mu_{ks} \Gamma^s_{\alpha \beta}$ \\
$\breve{\mu}$ & A vector of dimension $N n$ consisting of all vectors $\bmu_k$, $k=1, \ldots, N$ stacked together.
\\
$\{ \cdot, \cdot\}$ & Poisson bracket 
\\ 
$\Lambda(\mu)$ & Poisson tensor (usually for Lie-Poisson systems) 
\\
$\sigma$ & Activation function 
\\ 
$\mathbf{P}_\alpha$ & Poisson transformations used in machine learning; $\alpha$ is the index of that transformation in composition  
\end{tabular}

\section[Problem setup and general approach]{Problem setup and general approach}
\label{sec:Introduction}
\paragraph{Background: controlled dynamics of interacting autonomous vehicles}
Understanding and predicting the collective motion of interacting unmanned vehicles is becoming increasingly important for civilian and military uses.  Mathematically, the problem we consider can be formulated as follows. 
Suppose we observe the evolution of a system of $N$ interacting elements, each of them evolving in its own phase space, which is assumed to be a Lie group $G$. For the unmanned aerial vehicles (UAVs) or underwater drones, the individual phase space is $G={\rm SE}(3)^N$, where ${\rm SE}(3)$ is the group of rotations and translations in three-dimensional space. For land-based drones or drones operating on the sea surface, $G={\rm SE}(2)^N$, and for a set of rigid bodies that are rotating about their center of mass, $G={\rm SO}(3)^N$. Suppose that we have observations of each particle of the system, and we know that the objects are connected and controlled, but we have limited information about the actual control procedure and the connection between them. The goal is to predict the collective motion of the system from data, without the need to decipher the actual information about the method of control.   

In its most general formulation, without any information about the control, the prediction of motion is likely impossible. However, with some limited assumptions on the nature of control, a solution to this problem can be found. In what follows, we assume that the collective behavior comes from the optimal control procedure. The process of symmetry reduction of problems of collective behavior using optimal control was developed in Justh and Krishnaprasad \cite{justh2010extremal, justh2015optimality}.  These papers show that the control Hamiltonians are of the type allowing complete reduction to coupled Lie-Poisson systems. We are going to predict the motion in exactly that case of complete Lie-Poisson reduction as described in \cite{justh2010extremal, justh2015optimality}. Without any knowledge of the system, 
the actual control Hamiltonian, depending on the directions of control and the control amplitudes,  cannot be recovered completely. However, we show that instead of learning the Hamiltonian and the nature of control, we can recreate phase space trajectories and predict dynamics of the system in the future, based on the available data observations only, for an arbitrary control Hamiltonian. 
 In order to do that, we propose a new type of Neural Networks, which we call CO-LPNets (Control Optimal Lie--Poisson Neural Networks), which are a generalization of Lie-Poisson Neural Networks \cite{eldred2024lie} for the problems of optimal control of multiple interacting vehicles. 
 
 \paragraph{Data-driven discovery of differential equations with structure}
Many recent studies have focused on the data-driven discovery of differential equations and the prediction of their evolution. One of the most popular methods is Machine Learning (ML) augmented by physical insights. 
Purely data-based ML methods have been successful in interpreting large amounts of unstructured data; however, they have experienced difficulties in physics and engineering applications, where the equations of motion are, to some extent, known, but data is often scarce. Suppose a system's evolution as a function of time is described by a quantity $\bu (t)$ belonging to some phase space, and the system's motion $ \bu (t)$ is described by the governing equation:
\begin{equation} 
\dot{\bu } =\boldsymbol{f} ( \bu )\,\, , \quad \bu(0) = \bu_0 \, . 
\label{general_eq}
\end{equation} 
 Here, $\bu$ and $\boldsymbol{f} $ can be either finite-dimensional, forming a system of ODEs, or infinite-dimensional, forming a system of PDEs where they depend, for example, on additional spatial variables $\bx$ and derivatives with respect to these variables, involving some boundary conditions in the latter case. In this proposal, we focus on autonomous equations, so $\mathbf{f}(\bu)$ in \eqref{general_eq} does not depend explicitly on time. If $\boldsymbol{f}(\bu)$ is known exactly, the traditional numerical simulation approach provides good control over errors, a clear understanding of the computations, and the reliability required for practical applications. An alternative approach, based on the theory of neural networks, was recently derived in  \citep{raissi2019physics} under the name of \textit{Physics Informed Neural Networks (PINNs)}, see the following reviews for a thorough overview \citep{karniadakis2021physics,cuomo2022scientific,kim2024review}. In this approach, the solution of the system $\bu(t)$ over time is approximated by a neural network, minimizing the loss function comprising the equations \eqref{general_eq} itself, computed through automatic differentiation  \citep{baydin2018automatic}, the boundary conditions, and available data, which yields the loss function for the network. PINNs offer computational efficiency, with reported speedups exceeding 10,000 times 
for evaluations of solutions in complex systems like weather \citep{bi2023accurate,pathak2022fourcastnet}, accompanied by a drastic reduction of the energy consumption needed for computation. In spite of the popularity of PINNs, the judgment on the applicability, advantages, and disadvantages of PINNs is still a subject of active discussions \cite{karniadakis2021physics,kim2024review,krishnapriyan2021characterizing}. 
In particular, PINNs approach struggles with long-or even medium-term predictions when equation \eqref{general_eq} can be written from some equations with structure (like canonical Hamiltonian, Lie-Poisson \emph{etc.}),   due to accumulation of errors. This is exactly the case we will consider. 
 We are interested in the case where the function  $\boldsymbol{f}(\bu)$ will be considered universally unknown through this proposal, except for some structure of the equation \eqref{general_eq}. 

When $\boldsymbol{f}(\bu)$ is not known,  one can use the method of equation discovery from given data. Suppose we have some information about the states of the system $\bu_i = \bu(t_i)$, for some time points $(t_1, \ldots, t_N)$. 
In that approach, differential equations are represented by a neural network, with the phase flow of that differential equation approximating, in some optimal sense, the available data \cite{schaeffer2017learning,raissi2018deep,rackauckas2020universal}. However, these approaches do not, in general, take into account the structure of equations, which is important for predicting the solutions over the intermediate and long term. 

\paragraph{Discovering equations vs mappings in phase space.} It is also worth discussing two different approaches to data-based computing. Some of the works for data-based approximation of general system \eqref{general_eq} compute the equations of motion, or, in formal language, the infinitesimal flow of the system \eqref{general_eq}. 
In data-based computing applied to systems with structure, such as canonical Poisson systems, there are also works that compute the infinitesimal flow using the Hamiltonian structure \cite{greydanus2019hamiltonian}. After the equation is derived via neural network approximation, one still needs to compute the solution, for example, using a structure-preserving integrator \cite{marsden2001discrete}. One could call such an approach a \emph{continuous} method. 

In contrast, other works have approximated the mapping $\boldsymbol{\Phi}_h$ of phase space of \eqref{general_eq} after a given time interval $\Delta t=h$,  taking the flow from the initial conditions to the time $t=h$, $\bu(h) =\boldsymbol{\Phi}_h (\bu_0)$. The solution at times $t = h, 2h \ldots nh$ starting at $\bu(0) = \bu_0$ can be obtained through an iterative sequence of mappings $\bu_{k} =\boldsymbol{\Phi}_h (\bu_{k-1}) $, $k=1, \ldots n$. When \eqref{general_eq} describes a canonical or Poisson system, these previous works used the properties of mappings in the phase space of canonical equations \cite{jin2020sympnets} or more general Poisson systems \cite{jin2022learning,vaquero2023symmetry,vaquero2024designing,eldred2024lie,eldred2025clpnets}.  We call that approach \emph{discrete} data-based computing. 
\\
{\bf \emph{All research in this paper addresses discrete data-based computing.} } In other words, the primary goal is to discover mappings in the phase space, not the equations themselves. 

The derivation of the mapping of phase space is advantageous as it allows a much more direct computation of solutions, without the need to compute the solution using a 'continuous' solver based on the learned function $\mathbf{f}(\bu)$ in \eqref{general_eq}. This is especially important for the long-term computations of systems with structure, where it is imperative to introduce as little artificial dissipation as possible.  
The methods we will describe below conserve special quantities called \emph{Casimirs} with machine precision for all times. They also provide excellent energy stability and, for chaotic systems, provide the growth of errors consistent with Lyapunov's exponent for that system. 

\paragraph{Data-based, structure-preserving computing of Poisson systems } 
There are several ways to apply physics-informed data methods to computations of Hamiltonian systems, respecting the structure of equations. Most of the work has been focused on \emph{canonical} Hamiltonian systems, \emph{i.e.}, the systems where the law of motion \eqref{general_eq} has the following particular structure: $\bu$ is $2n$-dimensional,  $\bu=(\bq, \bp)$, and there is a function $H(\bq,\bp)$, the Hamiltonian, such that $\mathbf{f} $ in \eqref{general_eq} becomes 
\vspace{-2mm} 
\begin{equation}
\mathbf{f}  = \mathbb{J} \nabla_{\bu} H \, , \quad  
\mathbb{J} = 
\left( 
\begin{array}{cc}
0 & \mathbb{I}_n  
\\
- \mathbb{I}_n & 0
\end{array}
\right) \,, \quad \Leftrightarrow \quad \dot \bq =  \pp{H}{\bp} \, , \quad 
\dot \bp = - \pp{H}{\bq} \,.
\label{canonical_system_gen}
\end{equation}
with $\mathbb{I}_n$ denoting the $n \times n$ identity matrix. 
The evolution of an arbitrary function $F(\bq,\bp)$ along a solution of \eqref{canonical_system_gen} is then described by the canonical Poisson bracket : 
\begin{equation}
\frac{dF}{dt} = \left\{ F, H \right\} = 
\pp{F}{\bq}\pp{H}{\bp}
- 
\pp{H}{\bq}\pp{F}{\bp}.
\label{canonical_bracket}
\end{equation}
The bracket \eqref{canonical_bracket} is a mapping sending two smooth functions of $(\bq,\bp)$ into a smooth function of the same variables. This mapping is bilinear, antisymmetric, acts as a derivative on both arguments (Leibnitz rule), and satisfies the Jacobi identity: for all functions $F,G,H$
\begin{equation}
\{ \{ F, G \}, H \} + \{ \{ H, F \}, G \} + 
\{ \{ G, H \}, F \} =0 \, . 
\label{Jacobi_identity}
\end{equation}
Brackets that satisfy all the required properties, \emph{i.e.}, are bilinear, antisymmetric, act as a derivative (\emph{i.e.}, satisfy Leibnitz rule), and satisfy the Jacobi identity \eqref{Jacobi_identity}, but are not described by the canonical bracket \eqref{canonical_bracket}, are called \emph{general} Poisson brackets (also known as \emph{non-canonical} Poisson brackets). The corresponding equations of motion are called Hamiltonian (or Poisson) systems. Often, these brackets have a nontrivial null space leading to the conservation of certain quantities known as the Casimir invariants, or simply  \emph{Casimirs}. The Casimirs are properties of the Poisson bracket and are independent of a particular realization of a given Hamiltonian. This paper will focus on the data-based approaches for computations of an important class of non-canonical Poisson systems. 

There is an avenue of thought that focuses on learning the actual Hamiltonian for the system from data, or, in the more general case, the Poisson bracket. 
This approach was explicitly implemented for canonical Hamiltonian systems in \citep{greydanus2019hamiltonian} under the name of \textit{Hamiltonian Neural Networks (HNN)}, which approximated the Hamiltonian function $H(\bq,\bp)$ fitting the evolution of the particular data sequence through equations \eqref{canonical_system_gen}. It was shown that embedding the dynamics with the knowledge of the data allows a much more accurate and robust way to approximate the solution compared to a general Neural Network (NN). This work was further extended to include the adaptive learning of parameters and transitions to chaos \citep{han2021adaptable}. The mathematical background guaranteeing the existence of a Hamiltonian function sought in HNN was derived in \citep{david2021symplectic}. 
An alternative method of learning the equations is given by the \textit{Lagrangian Neural Networks (LNNs)} \citep{cranmer2020lagrangian} which approximates the solutions of Euler-Lagrange equations, \emph{i.e.}, the equations in the coordinate-velocity space $(\bq, \dot \bq)$ \emph{before} Legendre-transforming to the momentum-coordinate representation $(\bq, \bp)$ given by equations \eqref{canonical_system_gen}. More generally, learning a vector field for the non-canonical Poisson brackets was suggested in \citep{vsipka2023direct}. The main challenge in that work was enforcing the Jacobi identity \eqref{Jacobi_identity} for the learned equation structure. Finally, a method for learning the Hamiltonian function for a given Poisson system from noisy data was recently proposed in \cite{hu2025global}, using a structure preserving method of kernel ridge regression, optimized for the flow of Poisson systems. These methods could reproduce the vector fields well, although it is difficult to reproduce the actual Hamiltonian since it is only defined up to an addition of an arbitrary function of the Casimir(s). 

In these and other works on the topic, one learns the vector field governing the system, with the assumption that the vector field can be solved using appropriate numerical methods. However, one needs to be aware that care must be taken in computing the numerical solutions for Hamiltonian systems, especially for long-term computations, as regular numerical methods lead to the distortion of quantities that should be conserved, such as total energy and, when appropriate, the momenta. 
To compute the long-term evolution of systems obeying Hamiltonian vector fields, whether exact or approximated by Hamiltonians derived from neural networks, one can use variational integrator methods \citep{marsden2001discrete,leok2012general,hall2015spectral} that conserve momentum-like quantities with machine precision. However, these integrators may be substantially more computationally intensive compared to non-structure preserving methods. 

We focus on an alternative approach, exploring the learning transformations in phase space that satisfy appropriate properties. For the canonical Hamiltonian systems, a well-known result of Poincar\'e states that the flow $\boldsymbol{\phi}_t(\bu)$ of the canonical Hamiltonian system \eqref{canonical_system_gen}, sending initial conditions $\bu$ to the solution at time $t$, is a symplectic map \citep{arnol2013mathematical,MaRa2013}. In terms of explicit expressions, with $\bu = (\bq, \bp)$, $\boldsymbol{\phi}_t(\bu)$ satisfies 
\begin{equation}
\left( \pp{\boldsymbol{\phi}}{\bu} \right)^T 
\mathbb{J}
\left( \pp{\boldsymbol{\phi}}{\bu} \right) = 
\mathbb{J} \, . 
\label{symplectic_map_def}
\end{equation}
Several authors pursued the idea of searching directly for the symplectic mappings obtained from the data, instead of finding actual equations of the canonical systems and then solving them. 

Perhaps the first work in the area of computing the symplectic transformations directly was done in  \citep{chen2020symplectic}, where \emph{Symplectic Recurring Neural Networks (SRNNs)} were designed. The SRNNs computed an approximation to the symplectic transformation from data using the appropriate formulas for symplectic numerical methods. Additionally, this paper demonstrates that, at least numerically, preserving the canonical structure can lead to improved results in cases where the data contains noise. An alternative method of computation of canonical Hamilton equations for non-separable Hamiltonians was done in \citep{xiong2020nonseparable}, building approximations to symplectic steps using a symplectic integrator suggested in \citep{tao2016explicit}. This technique was named \textit{Non-Separable Symplectic Neural Networks (NSSNNs)}. In terms of the Lagrangian approach,  \cite{sharma2024lagrangian}  includes the mappings in the phase space $(\bq, \dot{\bq})$ obtained by structure-preserving discretization, and the approach that we have extended in \cite{eldred2025variational} for thermodynamic systems and friction forces.

A more direct computation of symplectic mappings was done using three different methods in \citep{jin2020sympnets,chen2021data}. 
The first approach derived in \citep{jin2020sympnets} computes the dynamics through the composition of symplectic maps of a certain type, which was implemented under the name of \emph{SympNets}. Another approach \citep{chen2021data} derives the mapping directly using a generating function approach for canonical transformations, implemented as \textit{Generating Function Neural Networks (GFNNs)}. The approach using GFNNs enables an explicit estimation of the error in long-term simulations. In contrast, error analysis in SympNets focuses on the local approximation error.   Finally, \cite{burby2020fast} developed \emph{H\'enonNets}, Neural Networks based on the H\'enon mappings, capable of accurately learning Poincar\'e maps of a Hamiltonian system while preserving the symplectic structure. SympNets,  GFNNs, and H\'enonNets showed the ability to accurately simulate long-term behavior of simple integrable systems like a pendulum or a single planet orbit, and satisfactory long-term simulation for chaotic systems like the three-body plane problem. Thus, learning symplectic transformations directly from data shows great promise for long-term simulations of Hamiltonian systems. 

The method of SympNets was further extended for non-canonical Poisson systems by transforming the non-canonical form to local canonical coordinates using the Lie-Darboux theorem and subsequently using SympNets \citep{jin2022learning} by assuming that the dynamics occur within a neighborhood in which the Poisson structure has constant rank. This method was named \emph{Poisson Neural Networks (PNNs). } While this method can, in principle, treat any Poisson system by learning the transformation to the canonical variables and their inverse, we shall note that there are several difficulties associated with this approach. First, the Lie-Darboux transformation is typically only defined locally, and second, the Casimirs are not preserved exactly under the evolution. 
The preservation of Casimirs is especially important for predicting the probabilistic properties for the long-term evolution of many trajectories in the Poisson system \citep{Dubinkina2007}. 

Yet another alternative approach was suggested in \cite{vaquero2024designing}, where learning a Poisson system was suggested by finding approximate solutions for Hamilton-Jacobi equations for symmetry-reduced systems. Using these approximate solutions, transformations can be constructed that exactly conserve properties such as Casimirs or momentum maps, and have very good stability criteria. Unfortunately, the solutions of Hamilton-Jacobi equations for general systems are difficult. While powerful for some systems, the methods of \cite{vaquero2024designing} are difficult to apply for arbitrary systems when one cannot readily find the solution of the Hamilton-Jacobi equations. 

The  SympNets and PNN approach was further extended in \cite{bajars2023locally}, where volume-preserving neural networks \emph{LocSympNets} and their symmetric extensions \emph{SymLocSympNets} were derived, based on the composition of mappings of a certain type. These papers demonstrate consistently good accuracy of long-term solutions obtained by  LocSympNets and SymLocSympNets for several problems, including a discretized linear advection equation, rigid body dynamics, and a particle in a magnetic field. Although the methods of \cite{bajars2023locally} did not explicitly appeal to the Poisson structure of equations, the efficiency of the methods was demonstrated as applied to several problems that are essentially Poisson in nature, such as rigid body motion and the motion of a particle in a magnetic field. However, the extension of the theory to more general problems was hampered by the fact that the completeness of activation matrices suggested in \cite{bajars2023locally} was not yet known. 

\paragraph{Data-based predictions for Lie-Poisson systems} 
As we discuss in Section~\ref{sec:Lie_Poisson_OC_theory} below, the questions of optimal control for the coupled autonomous systems acquires the Lie-Poisson structure of physical systems. Following \cite{justh2010extremal,justh2015optimality}, symmetry reduction of the controlled Hamiltonian systems leads to the complete Lie-Poisson reduction to the direct product of the individual Lie algebras which represent the reduced phase space for each vehicle. In general, symmetry reduction of many integrating particles is incomplete and should involve the motion on the Lie algebra and also the motion of the relative Lie group elements \cite{ellis2010symmetry}. In that case, the Poisson bracket will not be the 'pure' Lie-Poisson bracket, but will involve the components for the evolution of relative group elements. However, such incomplete reduction is not realized in actual Lie-Poisson reduction of the optimal control due to the form of the Hamiltonian \cite{justh2010extremal,justh2015optimality}. 

In Lie-Poisson systems, the Poisson bracket is known exactly: it has an explicit expression stemming from the expression of the Lie algebra bracket. The choice of the actual Hamiltonian is a secondary step coming from physics or control.

To address the question of data-based computing in Lie-Poisson systems, in \cite{eldred2024lie,eldred2025clpnets}, we have developed the methods of constructing the activation maps directly from the brackets, predicting dynamics built out of maps computed by particular explicit solutions of Lie-Poisson systems. The method is applicable to \emph{all} finite-dimensional Lie-Poisson systems, as well as any Poisson system where explicit integration of appropriate equations for appropriate transformations is available, as was shown in \cite{eldred2024lie}.

  The advantage of utilizing explicit integration in Lie-Poisson equations to drastically speed up calculations on every time step was already noticed in \citep{mclachlan1993explicit}, although the application was limited to Hamiltonians of a certain form depending on the Poisson bracket. Our method is  applicable to arbitrary Hamiltonians; in fact, the Hamiltonian does not need to be known for the construction of the Neural Network; we only assume that the underlying symmetry and the appropriate Lie-Poisson bracket are known. That is a reasonable assumption; the symmetry and the Lie-Poisson brackets come from the symmetry embedded in the formulation of the problem, which is known exactly. In contrast, the Hamiltonian comes from physics or engineering, is often known only approximately and is driven by the modeling choices for the particular physical system \citep{Tonti2013}, in contrast to the Lie-Poisson bracket itself. 
  
  For the Lie-Poisson systems, the configuration manifold of the system is defined on some Lie group $G$, and, in the simplest case we describe here for brevity,  the system is invariant with respect to the action of that group. For example, the motion of a satellite about its center of mass is described by the group of rotations in space ${\rm SO}(3)$ ($3\times3$ orthogonal matrices). The dynamics is obtained from the canonical bracket in coordinates $\mathbf{g} \in G$ and momenta $\mathbf{p} \in T^*G$ (cotangent bundle to G),  using left or right symmetry-reduced momentum $\bmu =\mathbf{g}^{-1}  \bp$ or $\bmu =  \bp \mathbf{g}^{-1}$,  depending on physics and geometry of the problem. Then, $\bmu$ is an element of the dual to the Lie algebra. For the case of satellites, $\bmu \in \mso(3)^*$ -- antisymmetric matrices which can be viewed as vectors. The Poisson bracket and the equations of motion for the momenta $\bmu$ are then: 
  \begin{equation} 
 \dot \bmu = \left\{ \bmu, H \right\} \, , \quad \Leftrightarrow \quad  \dot \bmu = \Lambda(\bmu) \nabla H(\bmu) \, , \quad 
 \{ F, G \} = \pm \left< \bmu , \left[ \pp{F}{\bmu}, \pp{G}{\bmu} \right] \right>
 \label{LP_bracket} 
  \end{equation} 
where $[\cdot, \cdot]$ is the commutator of Lie algebra, and plus or minus in \eqref{LP_bracket} is chosen depending on whether the system is either right or left invariant, respectively. One can see that in the case of all Lie-Poisson systems, 
 $\Lambda(\bmu)$ is a matrix-linear function of $\bmu$ with the actual structure depending on the underlying Lie group.   A Lie-Poisson bracket may have Casimir invariants, or Casimirs, discussed above: these are special functions that make the Poisson bracket vanish $\{ C, H \}=0$ for any $H$. The Casimirs are going to play a special role in our considerations. 
 
The idea of the works \cite{eldred2024lie,eldred2025clpnets} is to use the fact that the phase flow of \emph{any} Hamiltonian, not necessarily having a physical meaning, is the Poisson transformation preserving the Poisson bracket exactly \cite{MaRa2013}. We thus generate 'test' Hamiltonians that have the form $F(\bmu \cdot \mathbf{a})$ where $F(\xi)$ is an unknown function to be approximated and $\mathbf{a}$ is an unknown vector of parameters. 
For such Hamiltonians, Lie–Poisson reduced dynamics equations \eqref{LP_bracket} are greatly simplified and can be solved analytically, forming the basis for Lie–Poisson transformations used in Machine Learning.

Then, \cite{eldred2024lie} proceeds by building a neural network out of sequences of these Poisson transformations, fitting the parameters of the functions $F(\xi)$ and the parameters $\mathbf{a}$,  so they approximate all available data in the phase space. The neural network learns the phase flow with high precision for the whole phase space, using a very small number of parameters and data points. Moreover, irrespective of the data used for learning, the neural network reproduces the conservation of Casimirs with machine precision. 

In spite of the simplicity and efficiency of this method, the completeness of Poisson transformations suggested in  \cite{eldred2024lie} was not proven, which hampered further development of the theory. In this paper, we show that the LPNets from \cite{eldred2024lie}  can be made complete under some generalizations, but the actual implementation of the complete system is impractical. Instead, we suggest a practical generalization of the Poisson transformations, allowing efficient and practical learning of the evolution in the whole phase space. 

\paragraph{Contributions of this paper} In this paper, we consider systems arising from the optimal control of the collective motion of a group of particles. The methods we develop allow us to derive explicit Poisson transformations for arbitrary Lie-Poisson dynamics, for an arbitrary number of particles, on arbitrary Lie groups, using a generalization of LPNets. We explicitly develop these techniques for interacting identical particles, with each particle evolving on the groups ${\rm SO}(3)$ of rotations and ${\rm SE}(3)$ of rotations and translations. The first case is relevant to the collective orientation of objects like satellites, or perhaps other objects like antennas, that rotate about their centers. The second case is relevant to the motion of unmanned vehicles in air or water, allowing three-dimensional rotation and translation.   
 For each case, we assume that we know the symmetry group, which leads us to the knowledge of the Poisson bracket for each case. Our goal is to predict the collective motion of these particles from observations. Our algorithm proceeds as follows.  

 \noindent 
 {\bf Step 1} We derive the Poisson transformations approximating the mappings in phase space as the flows from the test Hamiltonians, which are computed in the explicit form, as shown in Section~\ref{sec:Data_based_computing}. For a given Poisson bracket, a control Hamiltonian generates a phase flow that generates a Poisson map of the phase space into itself. That Poisson map is parameterized by a neural network depending on the initial conditions for each step. To learn the dynamics in phase space, the Poisson map will be approximated as a superposition of the Poisson maps coming from the phase flow of a test Hamiltonian. Since every map approximating the full motion is Poisson, it preserves the Poisson bracket and hence the Casimir invariants (Casimirs) to machine precision; hence, the superposition of the mappings of the phase flow generated by the test Hamiltonian will also preserve the Casimirs with machine precision. 
 
 \noindent 
 {\bf Step 2} Given a control Hamiltonian coming from the physical system, we generate the phase flow using a high-precision numerical method. That simulation generates data in phase space that are considered the ground truth data.  
 The test Hamiltonians have several parameters; the parameters are optimized in such a way that the superposition of mappings generated by the test Hamiltonians approximates the flow, on average, for all available data. 
 
 \noindent 
 {\bf Step 3} After the optimization, we compare numerical solutions, obtained via solving differential equations coming from the true control Hamiltonian (the ground truth), with predictions based on the approximation of the phase flow by a sequence of Poisson transformations as flows from test Hamiltonians. 

As an example of our technique, we study the flow for two types of control Hamiltonians, which we call 'Dictatorship' and 'Democracy'.
 By 'Dictatorship', we denote the case when only one particle is interacting with all other particles, and there are no other connections. The 'Democracy' means that all the evolving particles are connected to each other. Following \cite{justh2010extremal, justh2015optimality}, we obtained Lie–Poisson reduced dynamics equations for ${\rm SO}(3)$ and ${\rm SE}(3)$ groups in both cases. Interestingly, analytic results for some properties are possible for both of these cases, which we present below in Section~\ref{sec:Lie_Poisson_OC_theory}, extending the results in \cite{justh2010extremal, justh2015optimality}. 
For reference, we also present the results of Lie--Poisson reduced dynamics equations in the case of one particle, for general reference for these groups, as they have not been presented before in the literature, making the structure of equations for several particles clearer. 

\paragraph{Novelty of results} This work is an extension of the earlier results in \cite{eldred2024lie}, which is devoted to the data-based computing of the Lie--Poisson system with the use of LPNets to the problems of optimal control of collective motion of particles. The novelty of our work, compared to previous results in this field, lies in developing new, more efficient Poisson transformations that are highly efficient and complete, especially when applied to the control of interacting vehicles. 
No information about the type of control, the control directions, and the number of control dimensions is necessary for learning the whole phase space and predicting trajectories, as long as the optimal control procedure allows appropriate symmetry reduction.
More precisely, our results learn the flow for \emph{any} control Hamiltonian,  arising from a Lagrangian, which depends only on the controls $u$, so that the system admits the Lie-Poisson symmetry reduction as described in \cite{krishnaprasad1993optimal,justh2010extremal}.

Note that in our method, we never compute either the Hamiltonian $H$, its gradients, or the equations of motion. Instead, we compute only the composition of Poisson transformations, reproducing the dynamics in phase space of some Poisson system, 
coming from an unknown control Hamiltonian.

\paragraph{Plan of the paper} The rest of this paper is organized as follows. Section~\ref{sec:Lie_Poisson_OC_theory} presents the theoretical background material necessary for this article. Section~\ref{sec:dictatorship_Democracy} discusses Lie--Poisson reduced dynamics equations for ${\rm SO}(3)$ and ${\rm SE}(3)$ groups for two specific control Hamiltonians, namely, 'Dictatorship' and 'Democracy'. Section~\ref{sec:Data_based_computing} gives a detailed explanation of how to construct the Poisson transformations out of the test Hamiltonians, which will be the main tool for data-based computing for optimal control problems. Section~\ref{sec:Particular_groups} demonstrates application of CO-LPNets towards two Lie--Poisson systems, one of which evolves on ${\rm SO}(3)^N$ Lie group, which is described in Subsection~\ref{sec:SO(3) group}, and the other one on the group ${\rm SE}(3)^N$ Lie group, which is described in Subsection~\ref{sec:SE(3) group} (here $N$ is the number of particles). Section~\ref{sec:Conclusions} provides the Conclusions and directions for future work. 
In the Appendix, Section~\ref{sec:single_particle} provides Lie--Poisson reduced dynamics equations for the case of "single particle" for reference, in order to present the whole theory more clearly. Section~\ref{sec:Hamiltonian_gradients} provides the exact values of the derivatives of the Hamiltonians used for data generations and ground truth computations. Section~\ref{app_sec:Partial_deriv_transformations} of the Appendix provides partial derivatives of the transformations, and Section~\ref{app:sec_LLM} provides the documentation for prompts and LLM responses when these tools were used to provide some of the codes in this article. 

\section[Introduction]{Background work in the area}
\subsection{The Lie-Poisson theory of optimal control of coupled interacting particles}
\label{sec:Lie_Poisson_OC_theory}
In this section, we describe the theory of the symmetry-reduced optimal control of $N$ interacting particles that was developed in \cite{justh2010extremal,justh2015optimality}. We refer the reader to these papers for details of computations and derivations. 

We assume that all the interacting particles are identical, and a configuration of each particle is described by a Lie group $G$. 
The configurational manifold for the system of interacting particles is then $G^{N}$, which is the direct product of $N$ copies of the Lie group $G$. 

Let us start by describing the symmetry reduction for mechanical systems, and then proceed to describing applications of this theory to optimal control. A \emph{mechanical} system on $G^N$ can be described in terms of the canonical Poisson bracket on $(T^{*}G)^{N}$, the cotangent bundle of $G^N$, which can be written in canonical Poisson bracket in the coordinates $(g_1, \ldots g_N)$ and momenta $\mathbf{p}_1,\ldots,\mathbf{p}_N$. 

As we mentioned in Section~\ref{sec:Introduction}, the complete symmetry reduction of the canonical systems leads to the Lie-Poisson systems \eqref{LP_bracket}, see \cite{marsden2013introduction}, Chapter~10. For a general system defined on a Lie group $G$, one naturally has the associated Lie algebra $\mathfrak{g}$ with Lie bracket denoted as $[\alpha, \beta]$ for $\alpha, \beta \in \mathfrak{g}$. Let us denote by $\langle \mu, \alpha \rangle$ the duality pairing between vectors $ \alpha $ in the Lie algebra  $ \mathfrak{g} $ and co-vectors (or momenta) $ \mu = g^{-1} p$ in the dual space $ \mathfrak{g} ^* $ to $ \mathfrak{g}$. The partial derivatives of functions $F, H: \mathfrak{g} ^* \rightarrow \mathbb{R}$ with respect to $\mu$ thus belong to $ \mathfrak{g}$, leading to the \emph{Lie--Poisson bracket} as in \eqref{LP_bracket}. We only consider left-invariant systems so we have to choose the '-' sign in the definition of the Lie-Poisson bracket \eqref{LP_bracket}.

\paragraph{Pontryagin's maximum principle: Connection of optimal control problems with canonical Hamiltonian equations}
We start with Pontryagin's maximum principle, which maps the problem of Optimal Control to a problem that is close to the canonical Hamilton's equations. Suppose we have a system which is described by a differential equation $\dot{x}=a(x(t),u(t),t)$ with controls $u(t)$, where we apply control $u(t)$ towards the state vector $x(t)$. We also impose boundary conditions $x(t_{0})=x_{0}^{*}$, $x(t_{1})=x_{1}^{*}$. The optimal control problem aims to minimize the functional 
\begin{equation}   \eta=\int_{t_{0}}^{t_{1}}L(x(\tau),\dot{x}(\tau),  u(\tau), \tau)d\tau \, , 
\label{Optimal_control_def}
\end{equation} where $F$ is a function that specifies some optimality goal such as minimum time, effort, energy \emph{etc.}, and $t_{0}$ and $t_{1}$ are, respectively, the initial and final times. We want to find the optimal trajectory $x(t)$ and the optimal control $u(t)$ such that $\eta\rightarrow min$. The function $L$ under the integral in \eqref{Optimal_control_def} is called the control Lagrangian.
Pontryagin's maximum principle \cite{gamkrelidze2013principles,boltyanskii1961theory} proceeds by defining the control Hamiltonian $H(x,u,t)=\lambda(t) a(x,u,t)-L(x,u,t)$, where $\lambda$ is the co-state. The necessary extremal conditions can be expressed as follows:
\begin{equation}
\left\{
    \begin{aligned}
    0 & =  \pp{H}{u}   \quad 
    \mbox{ Optimality in control variable $u$ } \\
    \dot x & = \pp{H}{\lambda } \quad \mbox{ Equation for the state variable} \\
    \dot \lambda &= - \pp{H}{x} \quad \mbox{Equation for co-state (conjugate equation)}
    %
    \end{aligned}
    \right.
    \label{Pontryagin_max_principle}
\end{equation}
with additional transversality conditions at the initial and final points. 
We see that the second and third equations of \eqref{Pontryagin_max_principle} are exactly the canonical Hamiltonian equations in variables $x$ as coordinates and $\lambda$ as momenta.

In this paper, we assume that the control function $L$ in \eqref{Optimal_control_def} is not known, and thus we do not know the Hamiltonian in \eqref{Pontryagin_max_principle}. 
Furthermore, the actual value of control $u(t)$, the type of control, \emph{i.e.}, along which axes the control is applied, directions of applications, number of control directions, or other features of control leading to the Hamiltonian are assumed unknown. 
However, we assume that we know the symmetry of the configuration manifold, which is a natural physical assumption: after all, we know in what space the particles are evolving. Our goal is to compute the future motion of the system based on the knowledge of symmetry and the data about the past evolution of the system.  To achieve this goal, we first describe the theory of symmetry reduction in optimal control following \cite{justh2010extremal,justh2015optimality}. 

\paragraph{Symmetry reduction applied to optimal control}
Lie--Poisson Reduction Theorem for
${\rm GL}(n)$, see 
\cite{marsden2013introduction}, Section 13.2, deduces Lie--Poisson bracket from the general Poisson bracket under the process known as \emph{symmetry reduction}.
In mechanics, symmetry reduction is used to reduce the evolution from the full phase space (cotangent bundle of the configuration manifold, which does not need to be a linear space) to the Lie algebra, which is a linear space. For example, the evolution of individual satellites, computed through Euler's equations of motion, is formulated in the space of angular momenta, and not in the space of orientation matrices and corresponding momenta. 

The symmetry reduction of optimal control problem, similar to Lie-Poisson reduction for the canonical Hamiltonian systems to the Lie-Poisson systems \cite{krishnaprasad1989,ratiu2005crashcourse, marsdenratiureduction}, described in  \cite{krishnaprasad1993optimal}, see also \cite{Jurdjevic1999}, pp.~227-267, performs the reduction of \eqref{Pontryagin_max_principle} to the Lie--Poisson system for the case of Lie group symmetry. For the optimal control problem addressing a single system, one assumes that the state variable $x$ in \eqref{Pontryagin_max_principle} belongs to some Lie group $G$, and then the geometric meaning of the co-state $\lambda \in T^*G$ is the analogue of a mechanical momentum variable. If Pontryagin's Lagrangian depends only on the control $u$, and the control directly affects the velocities, as demonstrated in \eqref{control_vel} below, the Pontryagin Hamiltonian $H$ takes a particular form which allows complete symmetry reduction to the Lie-Poisson system. 

This theory can be further extended to joint dynamics of interacting controlled objects. Taking mechanical analogy, we start with the canonical Hamiltonian $H(g_{1}; \dots g_{N}; p_{1}; \dots p_{N})$, where each $g_{i} \in G$ and each of the canonical momenta $p_{i}$ belong to cotangent space $T^{*}_{g_i}G$. If we only have a symmetry with respect to the Lie group $G$, and not the full configuration manifold $G^N$, there is no complete reduction to the Lie-Poisson system. Instead, we have additional Lie group-valued variables of the type $\alpha_{ij} = g_i^{-1} g_j$. The resulting equations contain Lie-Poisson parts and equations for the evolution of $\alpha_{ij}$ \cite{ellis2010symmetry}, which pose a substantial complication. 
For example, the overall rigid body motion symmetry in a coupled two-body problem considered in \cite{grossman1998, krishnaprasad1989} allows Poisson reduction from the phase space
\begin{gather*}
T^{*} SO(3) \times T^{*} SO(3) \cong SO(3) \times SO(3) \times \mathfrak{so}^{*}(3) \times \mathfrak{so}^{*}(3)
\end{gather*}
to the reduced space $\rm{SO}(3) \times \mathfrak{so}^{*}(3) \times \mathfrak{so}^{*}(3)$. Physically, such reduction results in the coupled evolution equations for two momenta and relative orientation of two objects. 
Thus, in mechanical problems with the configuration manifold $G^N$ and $G$-symmetry reduction, it is unusual to have a complete reduction to Lie--Poisson dynamics, in this example, on $(\mathfrak{g}^{*})^{N}$, unless a very particular form of the Hamiltonian is assumed.  

Similar considerations extend to the optimal control theory of interacting particles \cite{justh2010extremal,justh2015optimality}. However, because of the particular form of the control Hamiltonian, there is a complete symmetry reduction to the pure Lie-Poisson case, without the need to consider the relative orientations.

Below we show how the collective dynamics of $N$ controlled particles  is described mathematically. This exposition will be needed for the  examples of control Hamiltonians in the cases of 'Dictatorship' and 'Democracy', which are considered in more details in Section~\ref{sec:dictatorship_Democracy}. While the focus of the paper is on the data-based approach, it is important to give a detailed exposition of the general theory to demonstrate how the ground truth data is obtained for learning. 

\paragraph{The general theory of controlled dynamics of $N$ interacting particles}
In the exposition below, we follow the notation of \cite{justh2010extremal,justh2015optimality}.
We describe the connection of the particles through a connected, undirected graph with vertices $v_{1}, v_{2}, \dots, v_{N} \in V$, without self loops, and denote the degree of vertex $v_{i}$ by $d(v_{i})$.  We introduce couple of useful definitions below.

\begin{definition}
Let $a_{ij}, \quad i,j=1,\dots, N$ be elements of the adjacency matrix $A$.
Then $a_{ij}=1$ if vertices $v_{i}$ and $v_{j}$ are connected, and $a_{ij}=0$ otherwise. 
\end{definition}

\begin{definition}
The degree matrix $D$ for the graph is defined as \\
$D=\operatorname{diag}(d(v_{1}),\dots,d(v_{N}))$.
\end{definition}

\begin{definition}
The graph Laplacian is defined as $B=D-A$. The elements of matrix $B$ are denoted as $b_{ij}, \quad i,j=1,\dots, N$. 
\end{definition}

By definition, $A$, $D$, and $B$ are symmetric matrices \cite{fax2004information}. Also,  $B\boldsymbol{1}_{N}=\boldsymbol{0}$, where $\boldsymbol{1}_{N}=[1 \quad 1 \quad \cdots \quad 1]^{\mathsf T}$, \emph{i.e.}, the $N \times 1$ vector of ones, so $\boldsymbol{1}_{N}$ is an eigenvector of $B$ with eigenvalue $0$. In other words, the sum of elements in every row and in every column of $B$ is equal to $0$.

Let us assume that we observe $N$ particles, which are connected with one another through some kind of control procedure. The graph can serve as a model of interaction between particles, where particles are vertices, and edges mean connection between particles, if and only if such a connection exists. Let us assume that each our particle evolves on a finite dimensional Lie group $G$, thus, we work on $G^{N}$ - the direct product of $N$ copies of the Lie group $G$.
Let $n$ be the number of basis elements (dimension) of the Lie group $G$ and of the corresponding Lie algebra $\mathfrak{g}$. We use the trace norm $|\xi|^{2}=\operatorname{Tr}(\xi^{\mathsf T}\xi)$, and inner product $\langle \langle \xi,\eta \rangle \rangle=\operatorname{Tr}(\xi^{\mathsf T}\eta)$, $\xi, \eta \in \mathfrak{g}$. Let us introduce the following basis of the Lie algebra
$\{ X_{1}, X_{2} , \dots, X_{n}\}$, which we assume to be orthonormal with respect to the trace inner product. Also let $\langle \cdot, \cdot \rangle$ denote duality pairing between original and dual spaces \cite{justh2015optimality}.

The analogue of the symmetry reduction for Pontryagin's maximum principle was derived in \cite{justh2010extremal, justh2015optimality, krishnaprasad1993optimal}. We present this computation briefly for completeness, as it will allow us to introduce some notation and investigate the limits of applicability of the method. 
We consider a particular case where $x = g \in G$, where $G$ is some Lie group, and consider \eqref{Optimal_control_def} with the control Lagrangian $L =L(u)$, in other words, we look at the minimization problem 
\begin{gather}
{\rm Min}_{u} \eta =  {\rm Min}_{u}  \int_{0}^{T} L(u) dt \,.
\label{minimization_problem}
\end{gather}
Clearly, $L(u)$ is a ${G}$-invariant Lagrangian, as it does not depend on $g \in {G}$ and depends only on controls $u$. Each control $u(t)$ determines a curve $\xi_{u(t)}=:\xi$ in the Lie algebra $\mathfrak{g}$. The evolution equation for $g$ is given by 
\begin{equation}
\dot{g}=g\xi \, .     \label{control_equations_invariant}
\end{equation}
with fixed endpoints $g(0)=g_{0}$,  $g(T)=g_{1}$, $g \in G$. 
Canonical equations for state and co-state variables, which are the last two equations in \eqref{Pontryagin_max_principle}, are replaced with the equations coming from the Lie--Poisson bracket from the general Poisson bracket in the Lie--Poisson Reduction Theorem, see for example 
\cite{marsden2013introduction}, Section 13.2. The first condition of \eqref{Pontryagin_max_principle} connects the controls with the symmetry-reduced momenta according to the following computation. 
We define the pre-Hamiltonian according to \eqref{Pontryagin_max_principle}
\begin{gather}
H(p,g,u)=\langle p, g\xi \rangle - L(u)=\langle \mu, \xi \rangle - L(u),
\label{general_pre-Hamiltonian}
\end{gather}
where $p \in T^{*}_{g} G^{N}$, the cotangent space at $g$, and $\mu \in (\mathfrak{g}^{*})^{N}$, the dual space of $N$ copies of Lie algebra $\mathfrak{g}^{N}$, sometimes called coalgebra. Using the mechanical analogy, we will just call $\mu$ the 'momentum', although the more complex term 'symmetry-reduced co-state' for that variable is more  rigorous and appropriate. We hope no confusion arises from this notation.

We will only be working with Lie groups having a matrix representation. In that representation, we denote $g$ (the group elements) and $\xi$ (controls) to be block-diagonal matrices: 
\begin{gather*}
 g=diag(g_{1},g_{2},\dots,g_{N}), \quad  \xi=diag(\xi_{1},\xi_{2},\dots,\xi_{N}),
\end{gather*}
where $u=(u_{1}, u_{2}, \dots, u_{N})$ is the control vector of length $mN$. 
In this notation, the pre-Hamiltonian \eqref{general_pre-Hamiltonian} can be rewritten as
\begin{gather}
H(p,g,u)=\sum_{k=1}^{N} \langle p_{k}, g_{k}\xi_{k} \rangle - L(u)=\sum_{k=1}^{N}\langle \mu_{k}, \xi_{k} \rangle - L(u).
\label{particular_pre-Hamiltonian}
\end{gather}
The optimality in control variable $u$ (a vector which is the collection of all controls), which is the first necessary extremal condition in \eqref{Pontryagin_max_principle}, is rewritten as follows:
\begin{equation}
\pp{H(\mu,u)}{u} =0 \quad \Leftrightarrow \quad 
\biggl \langle \mu, \frac{\partial \xi}{\partial u} \biggr \rangle - \frac{\partial L}{\partial u}=0, 
\label{necessary_minimization_condition}
\end{equation}
where $\mu =  (\mu_1, \ldots, \mu_N)$  is the vector indicating the collection of all $\mu_k$. 
According to the implicit function theorem \cite{krantz2002implicit}, the system of equations \eqref{necessary_minimization_condition} can be solved for the controls $u= u(\mu)$, if the Hessian of the Hamiltonian, taken with respect to $u$ variables only, $\operatorname{Hess}_{\breve u} H(\mu,u)$ is non-degenerate. In the case of $\xi$ being a linear function of $u$, that condition is equivalent to the non-degeneracy of the Hessian of the control Lagrangian:   $\operatorname{det} \operatorname{Hess}_u L(u) \neq 0$. In what follows, we assume that such non-degeneracy of $\operatorname{Hess}_u H(\mu,u)$ is satisfied.

Following \cite{justh2010extremal, justh2015optimality}, we  discuss optimal control problem \eqref{minimization_problem} for the following control Lagrangian describing the motion of interacting particles:
\begin{gather}
L=L (\xi_{1}, \dots, \xi_{N})= \frac{1}{2} \biggl(\sum_{k=1}^{N} |\xi_{k}|^{2}+\chi\sum_{k=1}^{N}\sum_{j=1}^{N}a_{kj}|\xi_{k}-\xi_{j}|^{2} \biggr),  
\label{particular_Lagrangian1}
\end{gather}
with $\chi \geq0$ is a constant. Using the symmetry of A, according to the definition of adjacency matrix $A$ and graph Laplacian $B$, the relationship between elements of matrices $A$ and $B$ are the following:
\begin{align*}
b_{ii}&=\sum_{j=1}^{N} a_{ij} \quad \mbox{for all digonal elements of matrix $B$};\\
b_{ij}&=-a_{ij} \quad \mbox{for all non-diagonal elements of matrix $B$}.
\end{align*}
Thus, the Lagrangian \eqref{particular_Lagrangian1} can be rewritten as follows:
\begin{gather}
L=L(\xi_{1}, \dots, \xi_{N})= \frac{1}{2} \sum_{k=1}^{N} |\xi_{k}|^{2}+\chi\sum_{k=1}^{N} \biggl \langle \biggl \langle \xi_{k}, \sum_{j=1}^{N} b_{kj}\xi_{j} \biggr \rangle \biggr \rangle. 
\label{particular_Lagrangian2}
\end{gather}
We define control velocities $\xi_{k}$ to be affine in the control vector $u_{k}$ for each $k$, i.e., we take
\begin{align}
\xi_{k}=X_{q}+\sum_{i=1}^{m}u_{ki}X_{i}, \quad k=1,\dots,N,
\label{control_vel}
\end{align}
where $u_{k}=(u_{k1},\dots,u_{km}) \in \mathbb{R}^{m}$, $\{X_{1},X_{2},\dots,X_{n}\}$ is a basis in the Lie algebra, which is assumed to be orthonormal with respect to the trace inner product, with $m<n$ (the number of controls is less than the number of dimensions). 
The drift component $X_q$ is acting on the non-controlled direction, so $m+1 \leq q \leq n$. Thus, the system is underactuated, has a drift in $X_{q}$ and is controlled in $m$ directions. Setting $X_{q}$ in \eqref{control_vel} to zero yields a driftless system.  With the substitution \eqref{control_vel}, we can write \eqref{particular_Lagrangian2} as $L=L(u_1,\dots,u_N)$.

As we mentioned above, the exact implementation of \eqref{control_vel} is not assumed to be known to us during the system discovery, but it is important for the implementation of the system providing the ground truth data.

We introduce the basis of the coalgebra $\mathfrak{g}^{*}$, defined as $\{ X_{1}^{b}, X_{2}^{b}, \dots, X_{n}^{b}\}$, dual to the basis of Lie algebra $\mathfrak{g}$. Let us denote $\mu_{ki}$ to be the coefficients of expansion of $\mu_k$ in this basis: $\mu_{k}=\sum_{i=1}^{n} \mu_{ki}X_{i}^{b}$. Then, due to the structure of control \eqref{control_vel}, \eqref{necessary_minimization_condition} is rewritten as follows:
\begin{gather}
\biggl\langle \mu_{k}, \frac{\partial \xi_{k}}{\partial u_{ki}} \biggr\rangle - \frac{\partial L}{\partial u_{ki}}=0 \,\Leftrightarrow \, 
\biggl\langle \mu_{k}, X_{i} \biggr\rangle - \frac{\partial L}{\partial u_{ki}}=0 \,\Leftrightarrow \,
\mu_{ki}  - \frac{\partial L}{\partial u_{ki}}=0 \,  
\label{particular_minimization_condition}
\end{gather}
for $k=1, \dots, N$ and $i=1,\dots, m$.  
 
 From the expression of the Lagrangian \eqref{particular_Lagrangian1} we obtain the connection between the reduced co-state $\mu$ and controls 
\begin{equation}
 \frac{\partial L}{\partial u_{ki}}=u_{ki}+\chi\sum_{j=1}^{N}b_{jk}u_{ji} = \mu_{ki} \, , \quad k=1, \dots, N, \quad i=1, \dots, m \, . 
 \label{mu_u_connection_quadratic}
\end{equation}
To write \eqref{mu_u_connection_quadratic} in vector form, let us introduce the vector of all momenta $
\breve{\mu}=\begin{bmatrix}\mu_{1}^{\mathsf T} & \mu_{2}^{\mathsf T} & \cdots & \mu_{N}^{\mathsf T}\end{bmatrix}^{\mathsf T}$, 
of length $Nn$. In contrast, we will write the vector of the first $m$ (controlled) components of 
the momenta $\mu_k$ for each $k=1,\dots,N$ as 
$\tilde{\mu}_{k}=(\mu_{k1}, \dots , \mu_{km})$, $k=1,\dots, N$.

Then, for the tilde components of the momenta, we have the connection between the controls and momenta in the component and vector form: 
\begin{equation}
\tilde{\mu}_{k}=u_{k}+2\chi\sum_{j=1}^{N} b_{jk}u_{ji}    
\, \Leftrightarrow \, 
\left[\begin{matrix}
\tilde{\mu}_{1}\\
\vdots\\
\tilde{\mu}_{N}\\\end{matrix}\right] =  ((\mathbb{I}_{N}+2\chi B)\otimes \mathbb{I}_{m})
\left[\begin{matrix}
u_{1}\\
\vdots\\
u_{N}\\\end{matrix}\right],
\label{mu_control_connection}
\end{equation}
where $\otimes$ denotes the Kronecker product. 
Notice that all of the eigenvalues of $B$ are real and non-negative, including (at least) one zero eigenvalue (see, e.g., \cite{chung1997spectral, fax2004information}). Therefore, the inverse matrix $((\mathbb{I}_{N}+2\chi B)\otimes \mathbb{I}_{m})^{-1}$ is guaranteed to exist for $\chi\geq0$. For convenience, we will define this auxiliary matrix, obtained by solving \eqref{mu_control_connection}, which connects momenta and controls, by $\Psi$: 
\begin{gather}
\Psi=((\mathbb{I}_{N}+2\chi B)\otimes \mathbb{I}_{m})^{-1}=(\mathbb{I}_{N}+2\chi B)^{-1}\otimes \mathbb{I}_{m}.\label{matrix_PSI}
\end{gather}
We thus obtain the expression of the controls $(u_1, \ldots, u_N)$ as functions of momenta $\tilde{\mu}_1, \ldots \tilde{\mu}_N$. 
\begin{gather*}
\left[\begin{matrix}
u_{1}\\
\vdots\\
u_{N}\\\end{matrix}\right]= \Psi \left[\begin{matrix}
\tilde{\mu}_{1}\\
\vdots\\
\tilde{\mu}_{N}\\\end{matrix}\right],  
\end{gather*}
and substituting the found values of optimal controls back into the pre-Hamiltonian \eqref{particular_pre-Hamiltonian} yields, up to a constant, the control Hamiltonian \cite{justh2010extremal,justh2015optimality}: 
\begin{equation}
h=\sum_{k=1}^{N}\mu_{kq}+\frac{1}{2}\begin{bmatrix}\tilde{\mu}_{1} & \cdots & \tilde{\mu}_{N}\end{bmatrix}\Psi\left[\begin{matrix}
\tilde{\mu}_{1}\\
\vdots\\
\tilde{\mu}_{N}\end{matrix}\right].
\label{Reduced_ham_explicit}
\end{equation}
Mathematically, the reduced system is defined on $(\mathfrak{g}^{*})^{N}$, without any variables encoding the relative orientation of matrices, as we discussed above in Section~\ref{sec:Lie_Poisson_OC_theory} 
%
\cite{justh2010extremal,justh2015optimality}. 

The symmetry-reduced dynamics for $\breve{\mu}$ is given by the Lie--Poisson equations 
\begin{equation}
\dot{\breve{\mu}}=\Lambda(\breve{\mu})\nabla h \, , \quad 
\Lambda(\breve{\mu})=-\frac{1}{\sqrt{2}}\left[\begin{matrix}
\widehat{\mu}_{1}&0&\cdots&0\\
0&\widehat{\mu}_{2}&\cdots&0\\
\vdots&\vdots&\ddots&\vdots\\
0&0&\cdots&\widehat{\mu}_{N}\\\end{matrix}\right],
\label{LP_reduced_dynamics_equations}
\end{equation}
where we have defined the $k$-th block of the Poisson tensor $\Lambda(\breve{\mu})$ in \eqref{LP_reduced_dynamics_equations} as the following $n \times n$ antisymmetric matrix:
\begin{gather}
\widehat{\mu}_{k}=-\sum_{s=1}^{n} \mu_{ks}  \left[\begin{matrix}
\Gamma_{11}^{s}&\Gamma_{12}^{s}&\cdots&\Gamma_{1n}^{s}\\[3pt]
\Gamma_{21}^{s}&\Gamma_{22}^{s}&\cdots&\Gamma_{2n}^{s}\\[3pt]
\vdots&\vdots&\ddots&\vdots\\[3pt]
\Gamma_{n1}^{s}&\Gamma_{n2}^{s}&\cdots&\Gamma_{nn}^{s}
\end{matrix}\right],
\label{Poisson_tensor_one_particle}
\end{gather}
and $\Gamma_{ij}^{s}$ are the structure constants for the corresponding Lie algebra~$\mathfrak{g}$, that is, 
\begin{equation}
[X_{i}; X_{j}]=\sum_{s=1}^{n}\Gamma_{ij}^{s}X_{s}.
\label{structure_const_def}
\end{equation}
\begin{remark}[Geometric structure of equations]
{\rm 
Due to the complete symmetry reduction of equations \eqref{LP_reduced_dynamics_equations}, they can be written in the coordinate-free form as 
$\dot \mu_k = - {\rm ad}^*_{\pp{h}{\mu_k}} \mu_k$, where ${\rm ad}^*$ is the co-adjoint action of the Lie algebra $\mathfrak{g}$, and we have taken the left invariance \cite{holm2009geometric,marsden2013introduction}. 
}
\end{remark}

\begin{remark}
{\rm Due to the structure of the Poisson tensor in \eqref{LP_reduced_dynamics_equations}, $\breve{\mu}=0$ is a stationary point for an \emph{arbitrary} control Hamiltonian $h$.   
}
\end{remark}
In order to derive the reduced control Hamiltonian $h(\breve{\mu})$ in \eqref{LP_reduced_dynamics_equations}, we will need to consider particular cases of the graph Laplacian $B$, with the particles moving on particular Lie groups.

\paragraph{Generating the ground truth data and predictions: the two cases}
Given the group $G$, the adjacency matrix $B$ and the controls, one derives the control Hamiltonian according to \eqref{Reduced_ham_explicit} and generates the trajectories according to the Lie-Poisson equations \eqref{LP_reduced_dynamics_equations}.  
In this paper, we perform detailed explicit calculations for the cases of $G={\rm SO}(3)$ and $G={\rm SE}(3)$ groups which describe particles with fixed centers of mass, and drones moving in space, respectively.

We present our calculations for two particular cases for the adjacency matrix, which we call \emph{Dictatorship} (one particle connected to all others) and \emph{Democracy} (all particles connected to each other), where some analytical results can be obtained. While the general case was considered earlier in \cite{justh2010extremal,justh2015optimality}, the computations for the particular cases we consider here are, as far as we know, novel. These computations are not the centerpiece of this paper, which focuses on the predictions of the system from data. Still, we believe that these computations are not without interest and we therefore present them in some detail. In particular, we present analytic formulas for the Hamiltonians for these cases.  
We solve the motion of particles in both cases numerically and generate the ground truth data for learning.  
 
In addition, in  Section~\ref{sec:single_particle} of the Appendix, we present the case of evolution for a single particle. Of course, such calculations do not describe coupled systems. Still, these computations are useful, as they allow us to clarify the structure of the system and obtain analytical solutions in certain cases. 

\subsection[Different cases]{Two different cases of adjancency matrix}
\label{sec:dictatorship_Democracy}

In what follows, to be concrete, we consider two cases of an adjacency matrix. The first case refers to the adjacency matrix $A$ when one 'leading' particle is connected to all other particles, which we call 'Dictatorship'. Another case happens when the adjacency matrix $A$ is such that all particles are connected with all other particles, which we will denote as 'Democracy'. In these two cases, we can explicitly compute the inverse matrix $(\mathbb{I}_{N}+2\chi B)^{-1}$ in the corresponding part of the matrix $\Psi$ defined by \eqref{matrix_PSI} in its explicit form. The matrix $(\mathbb{I}_{N}+2\chi B) $ is always invertible, but the explicit formula in \eqref{main_part_of_matrix_PSI_dictatorship} is useful for analytic considerations. 

\paragraph{The case of 'Dictatorship' governance}
At first, we consider a 'Dictatorship' case, that is, when particle~1 is connected to all other particles. There are no other connections between particles (edges of the graph).
In this case the adjacency matrix $A$ is given by the $N\times N$ matrix
\begin{gather*}
A =\left[\begin{matrix}
0&1&1&\cdots&1&1\\
1&0&0&\cdots&0&0\\
1&0&0&\cdots&0&0\\[-4pt]
\vdots&\vdots&\vdots&\ddots&\vdots&\vdots\\
1&0&0&\cdots&0&0\\
1&0&0&\cdots&0&0\end{matrix}\right],
\end{gather*}
The degree matrix is the $N\times N$ matrix $D=\operatorname{diag}(N-1,1,1,\dots,1,1)$, and the graph Laplacian is the $N\times N$ matrix
\begin{gather*}
B=D-A =\left[\begin{matrix}
N{-}1&-1&-1&\cdots&-1&-1\\
-1&1&0&\cdots&0&0\\
-1&0&1&\cdots&0&0\\[-4pt]
\vdots&\vdots&\vdots&\ddots&\vdots&\vdots\\
-1&0&0&\cdots&1&0\\
-1&0&0&\cdots&0&1\end{matrix}\right].
\end{gather*}
Moreover, for this particular case of 'Dictatorship' control, we can, after some algebra, obtain the exact formula for the inverse of $(\mathbb{I}_{N}+2\chi B)$ as: 
\begin{gather}
(\mathbb{I}_{N}+2\chi B)^{-1}=
\left(\left[\begin{matrix}
1{+}2\chi(N{-}1)&-2\chi&-2\chi&\cdots&-2\chi&-2\chi\\
-2\chi&1{+}2\chi&0&\cdots&0&0\\
-2\chi&0&1{+}2\chi&\cdots&0&0\\[-4pt]
\vdots&\vdots&\vdots&\ddots&\vdots&\vdots&\\
-2\chi&0&0&\cdots&1{+}2\chi&0\\
-2\chi&0&0&\cdots&0&1{+}2\chi\end{matrix}\right]\right)^{-1} \nonumber\\
=\left[\begin{matrix}
\frac{1{+}2\chi}{1{+}2N\chi}\!\!&\!\!\frac{2\chi}{1{+}2N\chi}\!\!&\!\!\frac{2\chi}{1{+}2N\chi}&\cdots&\frac{2\chi}{1{+}2N\chi}&\frac{2\chi}{1{+}2N\chi}\\
\frac{2\chi}{1{+}2N\chi}\!\!&\!\!\frac{1{+}2N\chi{+}4\chi^{2}}{(1{+}2N\chi)(1{+}2\chi)}\!\!&\!\!\frac{4\chi^{2}}{(1{+}2N\chi)(1{+}2\chi)}&\cdots&\frac{4\chi^{2}}{(1{+}2N\chi)(1{+}2\chi)}\!\!&\!\!\frac{4\chi^{2}}{(1{+}2N\chi)(1{+}2\chi)}\\
\frac{2\chi}{1{+}2N\chi}\!\!&\!\!\frac{4\chi^{2}}{(1{+}2N\chi)(1{+}2\chi)}\!\!&\!\!\frac{1{+}2N\chi{+}4\chi^{2}}{(1{+}2N\chi)(1{+}2\chi)}&\cdots&\frac{4\chi^{2}}{(1{+}2N\chi)(1{+}2\chi)}\!\!&\!\!\frac{4\chi^{2}}{(1{+}2N\chi)(1{+}2\chi)}\\[-3pt]
\vdots&\vdots&\vdots&\ddots&\vdots&\vdots\\
\frac{2\chi}{1{+}2N\chi}\!\!&\!\!\frac{4\chi^{2}}{(1{+}2N\chi)(1{+}2\chi)}\!\!&\!\!\frac{4\chi^{2}}{(1{+}2N\chi)(1{+}2\chi)}&\cdots&\frac{1{+}2N\chi{+}4\chi^{2}}{(1{+}2N\chi)(1{+}2\chi)}\!\!&\!\!\frac{4\chi^{2}}{(1{+}2N\chi)(1{+}2\chi)}\\
\frac{2\chi}{1{+}2N\chi}\!\!&\!\!\frac{4\chi^{2}}{(1{+}2N\chi)(1{+}2\chi)}\!\!&\!\!\frac{4\chi^{2}}{(1{+}2N\chi)(1{+}2\chi)}&\cdots&\frac{4\chi^{2}}{(1{+}2N\chi)(1{+}2\chi)}\!\!&\!\!\frac{1{+}2N\chi{+}4\chi^{2}}{(1{+}2N\chi)(1{+}2\chi)}\end{matrix}\right].
\label{main_part_of_matrix_PSI_dictatorship}
\end{gather}

\paragraph{The case of 'Democratic' control}
Next, we consider a 'Democracy' case, that is, when each particle is connected to all of the others. In other words, we have a complete graph.
In this case, the adjacency matrix A is given by the $N\times N$ matrix
\begin{gather*}
A =\left[\begin{matrix}
0&1&1&\cdots&1&1\\
1&0&1&\cdots&1&1\\
1&1&0&\cdots&1&1\\[-4pt]
\vdots&\vdots&\vdots&\ddots&\vdots&\vdots\\
1&1&1&\cdots&0&1\\
1&1&1&\cdots&1&0\end{matrix}\right],
\end{gather*}
the degree matrix is the $N\times N$ matrix $D=\operatorname{diag}(N-1,N-1,N-1,\dots,N-1,N-1)$, and the graph Laplacian is the $N\times N$ matrix
\begin{gather*}
B=D-A =\left[\begin{matrix}
N{-}1&-1&-1&\cdots&-1&-1\\
-1&N{-}1&-1&\cdots&-1&-1\\
-1&-1&N{-}1&\cdots&-1&-1\\[-4pt]
\vdots&\vdots&\vdots&\ddots&\vdots&\vdots\\
-1&-1&-1&\cdots&N{-}1&-1\\
-1&-1&-1&\cdots&-1&N{-}1\end{matrix}\right].
\end{gather*}
In this particular case, the inverse of the matrix $(\mathbb{I}_{N}+2\chi B)$ is also given explicitly as 
\begin{gather}
(\mathbb{I}_{N}+2\chi B)^{-1} =
\rem{ 
\left(\left[\begin{matrix}
1+2\chi(N-1)&-2\chi&-2\chi&\cdots&-2\chi&-2\chi\\
-2\chi&1+2\chi&-2\chi&\cdots&-2\chi&-2\chi\\
-2\chi&-2\chi&1+2\chi&\cdots&-2\chi&-2\chi\\
\vdots&\vdots&\vdots&\ddots&\vdots&\vdots\\
-2\chi&-2\psi&-2\chi&\cdots&1+2\chi&-2\chi\\
-2\chi&-2\chi&-2\chi&\cdots&-2\chi&1+2\chi\end{matrix}\right]\right)^{-1} \nonumber\\
} 
\left[\begin{matrix}
\frac{1{{+}}2\chi}{1{+}2N\chi}&\frac{2\chi}{1{+}2N\chi}&\frac{2\chi}{1{+}2N\chi}&\cdots&\frac{2\chi}{1{+}2N\chi}&\frac{2\chi}{1{+}2N\chi}\\
\frac{2\chi}{1{+}2N\chi}&\frac{1{+}2\chi}{1{+}2N\chi}&\frac{2\chi}{1{+}2N\chi}&\cdots&\frac{2\chi}{1{+}2N\chi}&\frac{2\chi}{1{+}2N\chi}\\
\frac{2\chi}{1{+}2N\chi}&\frac{2\chi}{1{+}2N\chi}&\frac{1{+}2\chi}{1{+}2N\chi}&\cdots&\frac{2\chi}{1{+}2N\chi}&\frac{2\chi}{1{+}2N\chi}\\[-3pt]
\vdots&\vdots&\vdots&\ddots&\vdots&\vdots\\
\frac{2\chi}{1{+}2N\chi}&\frac{2\chi}{1{+}2N\chi}&\frac{2\chi}{1{+}2N\chi}&\cdots&\frac{1{+}2\chi}{1{+}2N\chi}&\frac{2\chi}{1{+}2N\chi}\\
\frac{2\chi}{1{+}2N\chi}&\frac{2\chi}{1{+}2N\chi}&\frac{2\chi}{1{+}2N\chi}&\cdots&\frac{2\chi}{1{+}2N\chi}&\frac{1{+}2\chi}{1{+}2N\chi}\end{matrix}\right].\label{main_part_of_matrix_PSI_Democracy}
\end{gather}
The inversion of the matrix $(\mathbb{I}_{N}+2\chi B)$ does not allow explicit formulas as in the case of 'Dictatorship' \eqref{main_part_of_matrix_PSI_dictatorship} or 'Democracy' \eqref{main_part_of_matrix_PSI_Democracy}. Thus, it is advantageous to consider the dynamics for these particular cases of interactions between the particles, as they allow additional analytical insights into the problem. 

Analytical solutions may be obtained for a single particle dynamics for particular groups in terms of elliptic functions. Such solution for a particle evolving on ${\rm SE}(2)$ group was obtained in \cite{justh2015optimality}. In the Appendix, we derive these analytical solutions for ${\rm SO}(3)$ group in Section~\ref{app:sec_SO3_single_particle} and a particular realization of the control in ${\rm SE}(3)$ group in Section~\ref{app:sec_SE3_single_particle}. While the single-particle solution does not address the collective behavior, we present these solutions for reference since they allow analytical solutions, and, to our knowledge, have not been described before. 

\subsection{Algorithm for producing the data in phase space}
\label{sec:Producing_data_in_phase_space}
The algorithms presented in this Section are used for producing data for a particular realization of the phase space, which are taken to be $N$ interacting particles in either ${\rm SO}(3)$ or ${\rm SE}(3)$. 

We solve the Lie--Poisson reduced dynamics equations \eqref{LP_reduced_dynamics_equations} for two types of control Hamiltonians described above, which we call \emph{'Dictatorship'} and \emph{'Democracy'}. We generate $N_t=40$ short trajectories for ${\rm SO}(3)$ and $N_t = 80$ trajectories for ${\rm SE}(3)$ of the length $N_p = 51$ points each. We choose $\chi = 0.5$ for all simulations. The initial conditions for these trajectories are generated randomly with a uniform distribution in the box $\bmu_k \in  [-1; 1]^3$ for every particle $k=1, \ldots, N$. 
The numerical solutions are obtained using a high-precision Lie-Poisson integrator  \cite{zhong1988lie}, based on an implicit second-order midpoint rule, with second-order accuracy in the time step. The integrator conserves Casimir invariants with machine precision and Hamiltonian with the accuracy of about $10^{-14}$. The same integrator is used to construct ground truth for comparison with the reconstructed solutions obtained by the neural network. Such an accuracy may seem excessive as it far exceeds any reasonable accuracy of experimental data. However, it is important to have a structure-preserving integrator to avoid possible drift in solutions for the comparison to the reconstructed solutions over long times.

\section{Data-based computing of the coupled Lie-Poisson control systems}
\label{sec:Data_based_computing}
\subsection{General considerations and completeness result}
\label{sec:General_considerations_completeness}
The main goal of this article is to approximate and predict further evolution of the Lie--Poisson system. In what follows, we assume that we do not know the Hamiltonian itself, and thus we do not assume the knowledge of the Lagrangian \eqref{particular_Lagrangian1}, the form of the controls \eqref{control_vel}, or a particular form of the Hamiltonian \eqref{general_pre-Hamiltonian}.  The Lagrangian and controls are assumed to be arbitrary, as long as the non-degeneracy of the Hessian in \eqref{general_pre-Hamiltonian} is satisfied. Our goal is to predict the evolution of the Lie-Poisson control system based on available data observations only. 

For the data set, we consider a set of points corresponding to the start and end points of short trajectories obtained from the exact equations of motion as described in Section~\ref{sec:Producing_data_in_phase_space}. This procedure of data generation gives a set of data points in the phase space connected by the phase space transformation over a short time interval. Our goal is to learn the phase space transformation connecting the data points in some optimal sense that we quantify below.   

\begin{remark}[On data observability and applicability of our theory]
{\rm 
    The ground truth data obtained in Section~\ref{sec:Producing_data_in_phase_space} used for learning the system involves information about the co-states of the system $\mu$. Such information is not readily available from the data collected by an external observer. The data can, however, be obtained from the internal control methods of the system itself. The methods for data-based predictions we derive here can then be used to compare the actual evolution of the system to the predicted evolution in case there are inaccuracies in the control application or the analytic expression for the control Lagrangian/Hamiltonian. In these cases, the data-based prediction that we develop here can be a valuable tool for fine-tuning the system and predicting its evolution. 
    }
\end{remark}

As we mentioned in Section~\ref{sec:Introduction}, in this work, we do not seek the approximation of the actual control Hamiltonian from data. Instead, we are focusing on the approximation of the mappings of the phase space in the $\mu$ variables between the data points in time using a sequence of structure-preserving Poisson maps.  For this purpose, we consider Hamiltonians $h$ for which we can solve the equations of motion analytically. We call such Hamiltonians \textit{test Hamiltonians}. These test Hamiltonians consist of arbitrary functions of projections of the momenta $\mu$ on a certain vector.  We construct Poisson transformations as the flows of these test Hamiltonians, and apply these Poisson transformations sequentially to approximate the flow generated by the real control Hamiltonian.
We start our considerations with the following 
\begin{lemma}
\label{lem:Lin_transform_LPNets}
Suppose $h(\mu) = f(\xi)$, where $\xi=  \left\langle \alpha , \mu \right\rangle + \beta$,  $\alpha$ is a vector belonging to the Lie algebra $\mathfrak{g}^N$, dual to $\breve{\mu} \in (\mathfrak{g}^*)^N$, and $\beta$ is a constant. Then the Lie--Poisson reduced dynamics \eqref{LP_reduced_dynamics_equations} reduces to equations that are linear in $\mu$.
\end{lemma}
{\bf Proof}. By definition, any Hamiltonian $h$ is a constant, thus, $\xi=  \left\langle \alpha , \mu \right\rangle + \beta=const$ as well. Moreover, $\nabla h=af'(\xi)=const$. According to \eqref{LP_reduced_dynamics_equations}, $\Lambda(\breve{\mu})$ is a block-diagonal matrix. \eqref{LP_reduced_dynamics_equations} suggests that the Lie-Poisson reduced dynamics equations for $i$-th component of $k$-th particle will look as follows:
\begin{gather*}
\dot{\mu}_{ki}=-\sum_{j,s=1}^{n} \mu_{ks}  \Gamma_{ij}^{s}
\pp{h}{\mu_{kj}}
\rem{ 
\left[\begin{matrix}
\Gamma_{i1}^{k}&\Gamma_{i2}^{k}&\cdots&\Gamma_{in}^{k}
\end{matrix}\right] \left[\begin{matrix}
\partial h/\partial \mu_{m1}\\
\partial h/\partial \mu_{m2}\\
\vdots\\
\partial h/\partial \mu_{mn}\\\end{matrix}\right]
} 
\end{gather*}
that is, for every $k=1,\dots,N; \quad i = 1 \ldots n$, we have: 
\begin{gather}
\dot{\mu}_{ki}=-\sum_{j,s=1}^{n} \mu_{ks}\Gamma_{ij}^{s}a_{kj}f'(\xi).
\label{lemma_eq}
\end{gather}
We know that $f'(\xi)=const$ on solutions of \eqref{lemma_eq}. 
Hence, the term multiplying the function $f'(\xi)$ is constant on solutions in \eqref{lemma_eq}, and thus the equation \eqref{lemma_eq} can be solved as a linear system of ODEs in the unknown $\mu$.
$\blacksquare$

\paragraph{Connection to splitting methods in numerical analysis}
Any Hamiltonian $h=f(\xi)$ described in Lemma~\ref{lem:Lin_transform_LPNets} generates a phase flow that is a Poisson transformation which can be computed analytically. 
Suppose the total Hamiltonian $h$ can be separated into a finite number of terms described in Lemma~\ref{lem:Lin_transform_LPNets}, \emph{i.e.} $h = \sum_{i=1}^M f_i(\left< \alpha_i, \mu \right>+ \beta)$.  Then, the idea of \emph{splitting methods} \cite{mclachlan2002splitting,mclachlan2006geometric} of the traditional numerical analysis is to divide each time step $\Delta t$ into $M$ sub-steps $\tau_j$, with $\tau_1 + \tau_2 + \ldots \tau_M = \Delta t$, with each sub-step $j$ yielding the phase flow $\Phi^j_{\tau_j}$ according to the solution of \eqref{lemma_eq}.
The total phase flow after time $\Delta_t$ can be approximated as a composition of the phase flows by each sub-Hamiltonian $h_i = f_i(\left< \alpha_i, \mu \right>+ \beta)$: 
\begin{equation}
    \Phi_{\Delta t} = \Phi^{M}_{\tau_M} \circ \Phi^{M-1}_{\tau_{M-1}} \circ 
    \ldots \circ \Phi^{1}_{\tau_1} \, , \quad 
    \sum_{j=1}^M \tau_j = \Delta t \, .
    \label{splitting_methods_def}
\end{equation}
\paragraph{Completeness of transformations defined in Lemma~\ref{lem:Lin_transform_LPNets}}
To prove the completeness of transformations derived in Lemma~\ref{lem:Lin_transform_LPNets}, we use a corollary of the Universal approximation theorem for the function and derivatives from \cite{hornik1990universal}:  
\begin{theorem}
{\rm Take any smooth, monotonically increasing function $\sigma(\xi)$.
Consider any $\epsilon>0$ and any compact set $U$ of the phase space. Then, any smooth Hamiltonian $h(\mu)$ can be approximated in $U$ up to the accuracy $\epsilon$ in the second Sobolev norm $W_p^2(U)$, $1 \leq p < \infty$ (the function and its first and second derivatives belong to $L_p(U)$): 
}
\begin{gather}
\left\| h - \sum_{j=1}^M \left( \sigma(\left< \alpha_j , \breve{\mu} \right> + \beta_j \right) \right\|_{W_p^2(U)} < \epsilon \, .
\label{LPNets_complete}
\end{gather}
\end{theorem}
In fact, the universal approximation theorem in \cite{hornik1990universal} is stronger: it states that one can approximate all derivatives of a function up to order $m$ on a compact domain $U$ with multi-layer networks, not just a single-layer network as in \eqref{LPNets_complete}. However, multi-layer networks do not satisfy the conditions of Lemma~\ref{lem:Lin_transform_LPNets} and the resulting Hamiltonians for each sub-step will not be integrable in explicit form. Also, we do not need to approximate the derivatives of the Hamiltonian of order higher than two. 
Using Theorem~\eqref{LPNets_complete}, we formulate the following 
\begin{lemma}[On the approximation of phase flow]
\label{lem:Approx_phase_flow}
  {\rm Suppose a phase flow of the Poisson system $\frac{d \breve{\mu}}{d t}  =  \Lambda(\breve{\mu}) \nabla h $ \eqref{LP_reduced_dynamics_equations} is defined by a smooth Hamiltonian $h(\breve{\mu})$. Choose a compact set $U$ in the phase space, and some $T>0$. Then, take any $\epsilon>0$, and choose the set $U_\epsilon: \{ \mathbf{x} \in U: \, {\rm dist}(\mathbf{x}, \partial U > \epsilon \}$ \footnote{In other words, $U_\epsilon$ is a subset of $U$ with all the points of $U_\epsilon$ separated from the boundary $\partial U$ by at least $\epsilon$.}  Then, there is an integer $M>0$,  and constant vectors $\alpha_j$ in $\mathfrak{g}^N$, and $\beta_j \in \mathbb{R}$ such that the flow is approximated by the phase flow of the Hamiltonian 
  \begin{equation}
      h_{\rm approx} = \sum_{j=1}^M 
      \sigma(\left< \alpha_j, \breve{\mu} \right>  + \beta_j)
      \label{h_approx}
  \end{equation}
  up to the accuracy $\epsilon$, for all $0<t<T$ as long as the solution generated by the phase flow of the Hamiltonian $h$ does not leave  $U_\epsilon$. }
\end{lemma}
{\bf Proof} Let us choose arbitrary  $\epsilon>0$, and some $\epsilon_1 = \epsilon/D$, where $D>0$ is to be determined. We also choose $p>d$, where $d = {\rm dim} \breve{\mu} = N n$. Let us take $M$ and $(\alpha_j,\beta_j)$  such that the true Hamiltonian $h$ and $h_{\rm approx}$ are $\epsilon_1$-close in $W^2_p$ norm. 
\eqref{LPNets_complete} is satisfied for $\epsilon = \epsilon_1$. Then, the Poisson vector fields defined by $\frac{d \breve{\mu}}{d t}  = - \Lambda(\breve{\mu}) \nabla h $ for $h=h(\breve{\mu})$ and $h=h_{\rm appr}(\breve{\mu})$ are $\epsilon_1$-close in $W^2_p$, and, since $p>d$, by Sobolev's embedding theorem, they are $C^1$-close uniformly in $U$. 

Suppose $L$ is the Lipshitz constant of the Poisson vector field $\mathbf{f}(\breve{\mu}) = \Lambda({\breve{\mu}}) \nabla h$.  We define $\mathbf{f}_1(\breve{\mu}) = \Lambda({\breve{\mu}}) \nabla h_{\rm approx}$ to be the vector field generated by the approximate Hamiltonian. Then, the solutions $\mathbf{x}(t)$ and $\mathbf{x}_1(t)$ satisfy the integral equations 
\begin{equation}
   \mathbf{x} = \mathbf{x}(0) + \int_0^t 
\mathbf{f}(\mathbf{x}(s)) \mbox{d} s \, , \quad 
\mathbf{x}_1 = \mathbf{x}_1(0) + \int_0^t 
\mathbf{f}_1(\mathbf{x}(s)) \mbox{d} s
\, . 
\label{x_0_x_1}
\end{equation}
Assuming that the initial conditions are the same, $\mathbf{x}_1(0) = \mathbf{x}(0)$, from \eqref{x_0_x_1} we obtain the estimate for the error 
$e(t) = \left|\mathbf{x}(t)- \mathbf{x}_1(t) \right| $, for all $t$ such that $\mathbf{x}(t)$ and $\mathbf{x}_1(t)$ are in $U_\epsilon$: 
\begin{equation}
\begin{aligned} 
e(t) & = \left|\mathbf{x}(t) - \mathbf{x}_1(t) \right| = 
\left| \int_0^t \mathbf{f}(\mathbf{x}(s)) - \mathbf{f}_1(\mathbf{x}_1(s)) \mbox{d} s
\right| 
\\ 
& = \left| \int_0^t \mathbf{f}(\mathbf{x}(s)) - \mathbf{f}_1(\mathbf{x}(s)) + 
\mathbf{f}_1(\mathbf{x}(s))
\mathbf{f}_1(\mathbf{x}_1(s)) \mbox{d} s
\right| \leq \epsilon_1 + \int_0^t L e(s) \mbox{d} s \, . 
\end{aligned}
    \label{error_eq}
\end{equation}
By the integral Gronwall's inequality \cite{hartman2002ordinary}, equation \eqref{error_eq} leads to the estimate for $e(t)$ as 
\begin{equation}
    e(t) \leq \frac{\epsilon_1}{L}
    \left( e^{L t} - 1\right) \, . 
\end{equation}
Thus, for any $0<t<T$, such that the solution does not leave $U_\epsilon$, we can take 
\begin{equation}
    \epsilon_1 = \epsilon \frac{L}{e^{L T}-1}  \, , 
    \label{epsilon_1_choice}
\end{equation}
which guarantees that $e(t) = |\mathbf{x}(t) - \mathbf{x}_1(t)|< \epsilon$, for $0<t<T$ as long as  $\mathbf{x}(t)$, $\mathbf{x}_1(t)$ staying in $U$. If the solution leaves $U$ at  $T_1<T$, then \eqref{epsilon_1_choice} still works and $e(t) < \epsilon$, but only on the interval $0<t<T_1$. 
$\blacksquare$
\\ 
\paragraph{Practical applications of completeness results} In principle, Lemma~\ref{lem:Approx_phase_flow}, yielding the completeness result, allows us to construct a splitting method based on the approximation of the Hamiltonian by the separable ansatz \eqref{h_approx}. One would then take $M$ sub-steps on each time step and approximate the flow uniformly in the compact domain $U$. 

In reality, such a straightforward application of the method is not practical. Since only the approximation by single-layer networks is allowed for a separable ansatz \eqref{h_approx}, the number of terms $M$ needed for accurate approximations can easily be very large. Estimates for deep ReLU networks \cite{yarotsky2017error}, for example, yield the estimates for the function and first derivative with accuracy $\epsilon$ $M \sim \epsilon^{-d}$, where $d$ is the number of dimensions, although one-layer networks tend to provide even worse approximations. 
Taking a modest value $\epsilon = 0.1$ and the number of dimensions $d=18$ which we will use for three interacting particles in  ${\rm SE}(3)$ in Section~\ref{sec:SE(3) group}, we arrive to the astonishing number of $M \sim 10^{18}$ substeps per time step. Clearly, writing a machine learning scheme with that number of parameters and having that number of substeps is impractical to say the least. Even if that number is reduced substantially, there will never be enough data in practical applications to construct such a machine learning scheme, which in addition will be extremely slow and memory-demanding.  In what follows, we develop a numerical scheme that has the number of substeps growing quadratically with $d$, the dimension of the phase space, which is highly efficient. Our scheme will be able to address a high number of dimensions with a relatively small number of data points (only a few thousand), to learn the complete dynamics of the phase space.

\begin{remark}[On polynomial Hamiltonians]
  {\rm   A theorem on polynomial expansion \cite{mclachlan2002splitting}, Chapter 3, says that if the Hamiltonian $H(\breve{\mu})$ is a polynomial, there exist an expansion $h = \sum_{j=1}^M f_j(\left< \alpha_j, \breve{\mu} \right>)$ where each $f_j$ is a polynomial of one variable. 
  The Hamiltonian \eqref{Reduced_ham_explicit} is quadratic, but that only happened because, following \cite{justh2010extremal,justh2015optimality} we made a very particular choice of the Lagrangian and controls. In general, there is no guarantee that the Hamiltonian is polynomial in components of $\breve{\mu}$. Our goal will be to describe the motion without any restriction on the type of Hamiltonians considered. One may be tempted to use a Taylor expansion of the Hamiltonian $h(\breve{\mu})$ to approximate it by a polynomial. Unfortunately, such expansion will fail, for example, if we multiply the quadratic Hamiltonian defined by \eqref{Reduced_ham_explicit} by a prefactor $(1+  \| \breve{\mu}\|^2)^{-1}$ and try to predict the motion of solutions in the domain $\| \breve{\mu}\| \leq 2$, as the Taylor expansion will fail beyond $\| \breve{\mu} \| \geq 1$. We thus proceed without making any assumptions on how well the Hamiltonian can be approximated by polynomials. 
  }
\end{remark}

\subsection{CO-LPNets: general consideration and design}
\label{sec:CO-LPNets}
\paragraph{The design of Poisson transformations}
We construct the solutions of the equations \eqref{lemma_eq} with the test Hamiltonian; these solutions will generate the transformations that we will use to learn the dynamics of the phase space. For $N$ interacting particles, each evolving on a Lie group of dimension $n$, we choose the test Hamiltonians $h_k = a_k \mu_{j_k}$, where $j_k$ is the chosen component for the $k$-th case. This test Hamiltonian generates the phase flow according to the equation 
\begin{equation}
h_k = w_k \mu_{j_k} \, , \quad \frac{d\breve{\mu}}{dt} = \Lambda(\breve{\mu}) w_k \mathbf{e}_{j_k} \, , \quad \breve{\mu}(t; \breve{\mu}_0) = \mathbf{P}_k(t, \breve{\mu}_0; w_k)\, , 
    \label{OC_PLNets_def}
\end{equation}
where $\mathbf{e}_{j_k}$ is the basis vector with $1$ at the $j_k$ coordinate and $0$ otherwise. The solution of this linear equation depends on the structure of the group, as well as the scalar $a_k$ setting the time scale for the particular equation. 
Usually, we take $K$ to be a multiple of $N n$, the total number of particles, and take $j_k = k \, \mbox{ mod} (N n)$, so the index $j_k$ runs repeatedly through $1, 2, \ldots N n$ a certain number of times. 

The crucial observation is that the transformations $\mathbf{P}_k$ defined by \eqref{OC_PLNets_def} remain Poisson if we make $w_k$ a function of the initial conditions $\breve{\mu}_0$. It is important to note that we set the value of parameter $w_k = w_k (\breve{\mu}_0)$ \emph{after} we have defined the Poisson transformation, which does not break the Poisson structure of the transformation itself. 

\begin{figure}
    \centering
\includegraphics[width=0.75\linewidth]{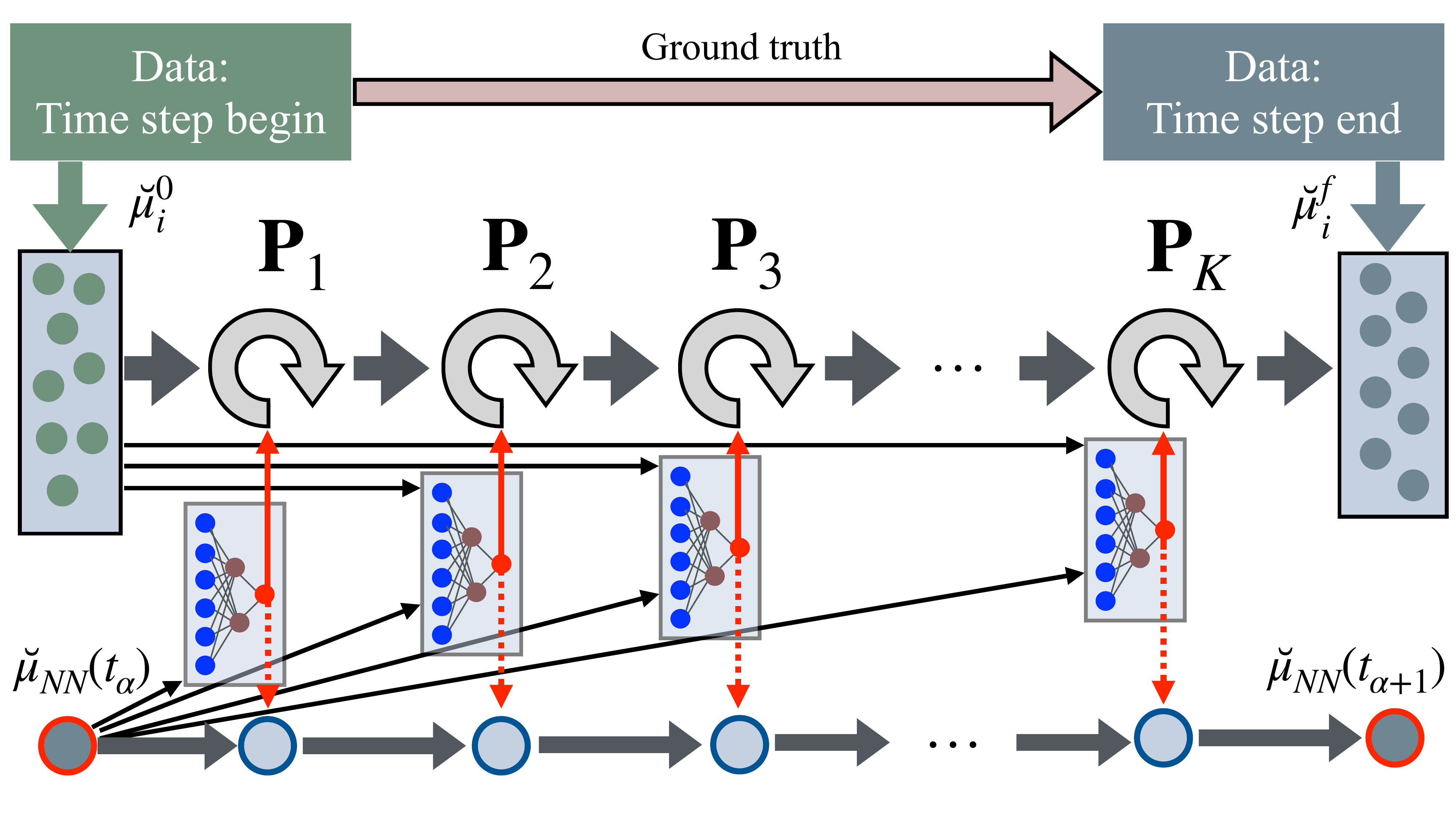}
    \caption{A schematic of proposed neural network. The top of the Figure illustrates the learning procedure. The ground truth data involves sets of short pieces of trajectories with the beginning and end points. The neural network consists of a sequence of Poisson transformations $\mathbf{P}_1, \mathbf{P}_2, \ldots, \mathbf{P}_K$, depending on parameters that are applied in sequence to the original points. The Poisson transformations are obtained from the test Hamiltonian $h_k = w_k \mu_k$, $k = 1, \ldots K$, with the index $k$ running through all the indices of the vector $\breve{\mu}$.  The parameters $w_k$ are defined by the neural network depending on the initial conditions for each time step $w_k = w_{k,NN}(\breve{\mu_0})$, indicated by the black arrows pointing to the network.  The network parameters for $w_{k,NN}(\breve{\mu_0})$ defining each sequence of transformation are optimized to the output of the mapping is as close as possible to the ground truth data for the end of the interval. On the bottom of the Figure, procedure for reconstructing the next step starting from $\breve{\mu}_{NN}(t_\alpha)$. The reconstruction consists of $K$ intermediate Poisson transformations with the learned parameters.  }
    \label{fig:Schematics}
\end{figure}

\begin{definition}{CO-LPNets}
{\rm A composition of Poisson mappings of phase space  $\mathbf{P}_k$ generated by Hamiltonians $h_k$ defined by \eqref{OC_PLNets_def} over time $\Delta t$ forms a Poisson transformation. We make the parameter $a_k$ to be defined as a function of the initial condition for every step $\breve{\mu}_0$ through a neural network $w_k = w_{k,NN}(\breve{\mu}_0)$. The total neural network is constructed as the composition of $\mathbf{P}_K \circ \mathbf{P}_{k-1} \circ \ldots \mathbf{P}_1$, with the parameter $a_k$ in each transformation depending on an individual neural network. The neural network constructed by such composition is called \emph{CO-LPNets}. The construction of \emph{CO-LPNets} is illustrated on Figure~\ref{fig:Schematics}. 
}
\end{definition}

\begin{remark}[On the type of networks for $w_k$]
    {\rm We take a shallow network with the input being all components of $\breve{\mu}_k$ and the output being one-dimensional parameter $w_k$, and one hidden layer of width $W$. There are $K$ identical networks, one predicting each  value of $w_k = w_{k,NN}(\breve{\mu})$, $k=1, \ldots, N$.
   The number of parameters in each network is thus 
    $N_{p}^0=(N \cdot n \cdot W)+2W+1$, and the total number of parameters is 
    $N_{p,\mbox{tot}}=K N_{p}^0$. We will be taking $K$ to be a multiple of the total dimensions $N n$ (usually equal to $Nn$), so the number of parameters in the network grows with the number of particles and dimensions of the group as $n^2 N^2  W$. }
\end{remark}
In our observations, we take $W$ to be quite small. In particular, in our simulations for $N=3$ particles, we take $W=3$, and $K=N n$. For particles evolving in ${\rm SO}(3)$ group, the number of parameters in each network is 34, with the total number of parameters being 306. For ${\rm SE}(3)$ group, each network contains 61 parameters, with the total number of parameters equal to 1098.  

\begin{remark}[On completeness of transformations]
    {\rm One can see that making index $j_k$ in \eqref{OC_PLNets_def} run through all the components of $\breve{\mu}$ creates $Nn$ independent vector fields, forming the basis for all possible vector fields in the phase space. Thus, one can say that the flows generating the phase space transformations defined by \eqref{OC_PLNets_def} are complete in the infinitesimal sense, describing all possible Poisson flows $\frac{d \breve{\mu}}{dt} = \Lambda(\breve{\mu}) \nabla h$ generated by all possible Hamiltonians $h(\breve{\mu})$. } 
\end{remark}

\paragraph{On possible simplifications}
Several simplifications of the CO-LPNets can be constructed. One can remove the middle layer and make the output neuron dependent only on one input neuron. Then, each network will contain only two parameters (three in the case of additional scaling), and the total number of parameters will be proportional to $K$. This was the idea of G-LPNets in \cite{eldred2024lie}. Unfortunately, for Lie-Poisson systems with block-diagonal matrices, such systems may have problems with convergence as Poisson transformations $\mathbf{P}_k$ will be isolated to different particles. A solution to this problem is to consider the output neuron depending on a pair of input neurons. Such mapping was implemented for this problem as a test case. However, we believe that the convergence of that algorithm will be slower, since it is not completely clear \emph{a priori} how many Poisson transformations of this type we will need to achieve convergence.  

\paragraph{The algorithm of CO-LPNets}
\begin{enumerate}
    \item Generate several short trajectories of ground truth data. The data will define the ground truth transformations $\mu^0_i \rightarrow \mu^f_i$, $i=1,\ldots, N_p$. 
    \item We construct the transformations $\mathbf{P}_k(w_k,\breve{\mu}_0)$ and compute the loss function 
    \begin{equation}
\begin{aligned}
    L(\overline{\mathbf{W}}) & = \sum_{j=1}^M 
    \left\| \breve{\mu}^f_j - \mathbf{P}_K(w_K) \circ 
    \mathbf{P}_{K-1}(w_{K-1})
    \circ \ldots \circ \mathbf{P}_{1}(w_{1})(\breve{\mu}^0_j)
    \right\|^2\, , 
    \\ w_j & = w_{j,NN}(\mathbf{W}_j)\, , \quad 
    \overline{\mathbf{W}}:= 
    (\mathbf{W}_1, \mathbf{W}_2, \ldots, \mathbf{W}_K) \, , 
\end{aligned}
\label{Loss_function_def}
\end{equation}
where $\overline{\mathbf{W}}$ stands for weights and biases of all networks defining $w_{k,NN}$, $k=1, \ldots K$. 
\item Find the optimal weights 
\begin{equation}
    \overline{\mathbf{W}}_* = 
    \mbox{arg min} L(\overline{\mathbf{W}})
    \label{Optimal_W}
\end{equation}
\item Start with an initial condition $\breve{\mu}_0$, and construct $N_r$ reconstruction steps:
\begin{equation}
\begin{aligned} 
\breve{\mu}_{\alpha+1} &= \mathbf{P}_K(w_K) \circ 
    \mathbf{P}_{K-1}(w_{K-1})
    \circ \ldots \circ \mathbf{P}_{1}(w_{1})(\breve{\mu}_\alpha), \quad 
    0 \leq \alpha \leq N_{r} \, ,\\
    w_k & = w_{k,NN}(\overline{\mathbf{W}}_*) \, . 
    \label{Reconstruction}
    \end{aligned}
\end{equation}
Compare the reconstructed solution with the ground truth solution obtained by the Lie-Poisson integrator.  
\end{enumerate}

\begin{remark}[On the structure of Poisson transformations]
    {\rm In what follows, we demonstrate that for the cases we consider, namely $G={\rm SO}(3)$ and $G={\rm SE}(3)$, the Poisson transformations $\mathbf{P}_k(w_k, \breve{\mu}_0)$ have the form $\mathbb{A}(a_k,w_k) \breve{\mu}_0$, where $\mathbb{A}_k$ is a matrix of a certain form. The transformation is nonlinear since $w_k = w_{k, NN}(\breve{\mu}_0)$. That structure of transformation is a consequence of the solutions to \eqref{OC_PLNets_def}. 
    Since every $\mathbb{A}_k(\breve{\mu}, w_k)$ only depends on its own network through a single parameter $w_k=w_{k,NN}(\breve{\mu})$, the optimization of the loss function \eqref{Loss_function_def} can be made  highly computationally efficient.   }
\end{remark}

\section{Specialization to particular groups}
\label{sec:Particular_groups}
\subsection{${\rm SO}(3)$ group}
\label{sec:SO(3) group}
\subsubsection{Equations for the ground truth calculations}
Let us now apply the general theory to a particular group. We start with the collective motion of objects, with each object evolving on the Lie group of rotations ${\rm SO}(3)$. The ${\rm SO}(3)$ group consists of $3 \times 3$ orthogonal matrices $\mathbb{Q}$ such that $\mathbb{Q}^{\mathsf T} \mathbb{Q} = \mathbb{Q} \mathbb{Q}^{\mathsf T} = \mathbb{I}_3$, where $\mathbb{I}_3$ is a $3 \times 3$ identity matrix, and $\operatorname{det} \mathbb{Q} = 1$. This example can be applied to a set of rigid objects rotating to achieve the desired orientation, for example, for observations of a particular target in space. 

To find the evolution of particles on the $N$ copies of Lie algebra $\mathfrak{so}(3)$, we defined the scalar product $<X, Y> = \operatorname{Tr} (X^{\mathsf T} Y)$ and use the following orthogonal basis: 
\begin{gather*}
X_{1}=\frac{1}{\sqrt{2}}\left[\begin{matrix}
0&0&\hphantom{-}0\\
0&0&-1\\
0&1&\hphantom{-}0\\\end{matrix}\right], \quad
X_{2}=\frac{1}{\sqrt{2}}\left[\begin{matrix}
0&0&1\\
0&0&0\\
-1&0&0\\\end{matrix}\right], \quad
X_{3}=\frac{1}{\sqrt{2}}\left[\begin{matrix}
0&-1&0\\
1&0&0\\
0&0&0\\\end{matrix}\right],
\end{gather*}
which is normalized so that $\|X_{j}\|=1$, $j = 1,2,3$ with respect to the trace norm. 
The structure constants for the $\mathfrak{so}(3)$ Lie algebra are $\Gamma_{ij}^{k}=\frac{1}{\sqrt2}\varepsilon_{ijk},$ where $\varepsilon_{ijk}$ is the Levi-Civita symbol.

For each particle, we introduce the vector of $\bmu_k$ which consists of the components of $\widehat{\mu}_k$ in the basis $\left\{ X_1, \ldots X_n\right\}$:
\begin{gather}
\bmu_k=(\mu_{k1}, \mu_{k2}, \mu_{k3})^{\mathsf T}, \quad k=1,\dots, N\, , \quad \Leftrightarrow \quad 
\widehat{\mu}_k = \sum_{j=1}^3 \mu_{kj} X_j \, . 
\label{notation_SO(3)}
\end{gather}
Thus, according to \eqref{LP_reduced_dynamics_equations}, $\Lambda(\breve{\mu})$ is a block-diagonal matrix of the following form:
\begin{gather*}
\Lambda(\breve{\mu})=\frac{1}{\sqrt{2}}\left[\begin{matrix}
\widehat{\mu}_{1}&0&\cdots&0\\
0&\widehat{\mu}_{2}&\cdots&0\\
\vdots&\vdots&\ddots&\vdots\\
0&0&\cdots&\widehat{\mu}_{N}\\\end{matrix}\right]
\end{gather*}
with $k$-th block equal to the following $3 \times 3$ antisymmetric matrix:
\begin{gather}
\widehat{\mu}_k=\left[\begin{matrix}
0&-\mu_{k3}&\mu_{k2}\\
\mu_{k3}&0&-\mu_{k1}\\
-\mu_{k2}&\mu_{k1}&0\\\end{matrix}\right], \quad k=1,\dots, N,
\label{one_block_of_Poisson_tensor_for_SO(3)_group}
\end{gather}
which is just the familiar \emph{hat map}, mapping vectors in $\mathbb{R}^3$ to elements of Lie algebra $\mathfrak{so}(3)$.
As a result of the decoupled Poisson tensor, there is a Casimir $c_k$ for each particle with the index $k = 1, \ldots N$:
\begin{align*}
c_{k}=\mu_{k1}^{2}+\mu_{k2}^{2}+\mu_{k3}^{2}=||\bmu_{k}||^{2}, \quad k=1,\dots, N.
\end{align*}
The Lie--Poisson equations \eqref{LP_reduced_dynamics_equations} are written in vector form as:
\begin{gather*}
\dot{\bmu}_{k}=\frac{1}{\sqrt{2}}\bmu_{k} \times \frac{\partial h}{\partial \bmu_{k}}, \quad k=1,\dots, N,
\end{gather*}
which are (formally) $N$ copies of the equations of motion a rigid bodies \cite{arnol2013mathematical}, with the Hamiltonian $h=h(\bmu_1, \ldots, \bmu_N)$ providing the interaction between the particles.

We assume that the control is applied in the $X_1$ direction and the drift is in the $X_2$ direction. Then, \eqref{control_vel} becomes 
\begin{equation}
\xi_{k}=X_{2}+u_{k1}X_{1}, \quad k=1,\dots, N,
\label{control_SO3}
\end{equation}
After some algebra, the Hamiltonian \eqref{Reduced_ham_explicit} for the cases of 'Dictatorship' yields: 
\rem{ 
References \cite[formula~(17), p.~3]{justh2010extremal} and \cite[formula~(2.19), p.~6]{justh2015optimality} lead us to $\Psi=(\mathbb{I}_{N}+2\chi B)^{-1}\otimes \mathbb{I}_{1}=(\mathbb{I}_{N}+2\chi B)^{-1}$, thus, $\Psi$ is the same as the matrix in \eqref{main_part_of_matrix_PSI_dictatorship}.
Then, according to \cite[formula~(21), p.~3]{justh2010extremal} and \cite[formula~(3.3), p.~9]{justh2015optimality}, the Hamiltonian looks like
} 
\begin{equation}
\begin{aligned}
h={}&\sum_{k=1}^{N}\mu_{k2}+\frac{1}{2}\biggl(\frac{1+2\chi}{1+2N\chi}\mu_{11}^{2}
+\frac{1+2N\chi+4\chi^{2}}{(1+2N\chi)(1+2\chi)}\sum_{j=2}^N \mu_{j1}^{2}  \\
&+\frac{4\chi}{1+2N\chi}\mu_{11}\sum_{j=2}^N \mu_{j1}+\frac{8\chi^{2}}{(1+2N\chi)(1+2\chi)}\sum_{i,j=2, i<j}^{N}\mu_{i1}\mu_{j1}\biggr).
\label{h_SO3_dictatorship}
\end{aligned}
\end{equation} 
Similarly, for the case of 'Democracy', we obtain
\begin{equation}
\begin{aligned}
h=\sum_{k=1}^{N}\mu_{k2}+\frac{1}{2}\left(\frac{1+2\chi}{1+2N\chi}\sum_{j=1}^N \mu_{j1}^{2}+\frac{4\chi}{1+2N\chi}\sum_{i,j=1, i<j}^{N}\mu_{i1}\mu_{j1}\right).
\label{h_SO3_Democracy}
\end{aligned}
\end{equation}

These two cases ('Dictatorship' and 'Democracy') are simulated for $N=3$ particles, computing $40$ trajectories of length $51$, thus creating $2000$ intervals for learning of the phase space. The initial conditions for these trajectories are chosen randomly in the hypercube $[-1,1]^d$, where $d$ is the dimension of the phase space, equal to $N n=9$ in this case. As mentioned in Section~\ref{sec:Producing_data_in_phase_space}, we use a Lie-Poisson integrator to compute the ground truth data, which preserves the system's structure and Casimirs with machine precision. We choose the value of the parameter $\chi =0.5$ for all simulations. For reference, we present analytical formulas for the derivatives of the Hamiltonians in Appendix~\ref{app_sec:SO3_exact_deriv}.

Next, we describe how to construct the Poisson transformations that define the transformations of the phase space. We then build CO-LPNets to learn the motion of the system in the whole space where the data is available, and predict the trajectories of the system that are previously unseen by the algorithm.

\subsubsection{Poisson transformations through test Hamiltonians} 
Following Section~\ref{sec:CO-LPNets}, we consider test Hamiltonians which depend only on a single component $i$ of one particle $k$. Following \eqref{OC_PLNets_def}, we start with the Hamiltonian $h_{k i} = w_{k i} \mu_{k i}$, where $w_{ki}$ is some constant on that particular time step. 

The Lie--Poisson reduced dynamics equations $\dot{\breve{\mu}}=\Lambda(\breve{\mu})\nabla h$ \eqref{LP_reduced_dynamics_equations} simplify to give 
\begin{equation}
\dot{\bmu}_{k} = \bmu_{k} \times \mathbf{e}_{i} w_{k i},
\label{test_Ham_rotation_SO3}
\end{equation}
where $\mathbf{e}_{i}$ means the $i$-th standard basis vector in the 3-dimensional space. Assuming that we are given initial conditions for $\breve{\mu}(0) = (\bmu_1(0), \ldots , \bmu_N(0))$, the solution of such an initial value problem is simply 
\begin{equation}
    \bmu_k(t) = \mathbb{R}(\mathbf{e}_{i}, - w_{k i}t) \bmu_{k}(0)\, , \quad \bmu_j(t) = \bmu_j (0) , \qquad j \neq k \, . 
\end{equation}
where  $\mathbb{R}(\mathbf{n}, \varphi)$ denoting the rotation matrix with respect to the vector $\mathbf{n}$ by the angle $ \varphi$.

To speed up computation of derivatives of the loss function \eqref{Loss_function_def} with respect to the parameters, we have also analytically computed the partial derivatives of matrices $\mathbb{R}$, $ i = 1,2,3$ with respect to $w_{ki}$. We present these formulas in the Section~\ref{app_sec:SO3_exact_deriv} of the Appendix.

\subsubsection{Application of CO-LPNets for ${\rm SO}(3)$ group}
\label{sec:SO3_results}

We generate the data in phase space, as was described in
\eqref{sec:Producing_data_in_phase_space}. We consider $N=3$ coupled particles each evolving in ${\rm SO}(3)$, with the dimension of each component being $n=3$. The total dimension of the space is $9$, and the number of generated intervals is $2,000$ with the time step $\Delta t =0.1$. Data is produced in two cases of particle interaction: 'Dictatorship' and 'Democracy'. We then apply CO-LPNets with $N n = 9$ transformations, the minimum possible number. Each transformation is parameterized by a neural network with $9$ input neurons, $W = 3$ hidden neurons, and one output neuron. We then optimize the loss defined by \eqref{Loss_function_def}. We perform an optimization using the Adam algorithm with a learning rate of $0.005$ for 10,000 epochs. During that time, the loss decreases from about $\sim 0.1-0.01$ to $\sim 10^-6$, \emph{i.e.} by four to five orders of magnitude. After the optimization is finished, we simulate ten trajectories starting with random initial conditions in the cube $[-1,1]^{N n}$, both using the structure-preserving integrator the CO-LPNets with computed parameters. 

\begin{figure} 
\centering 
\includegraphics[width=0.9\textwidth]{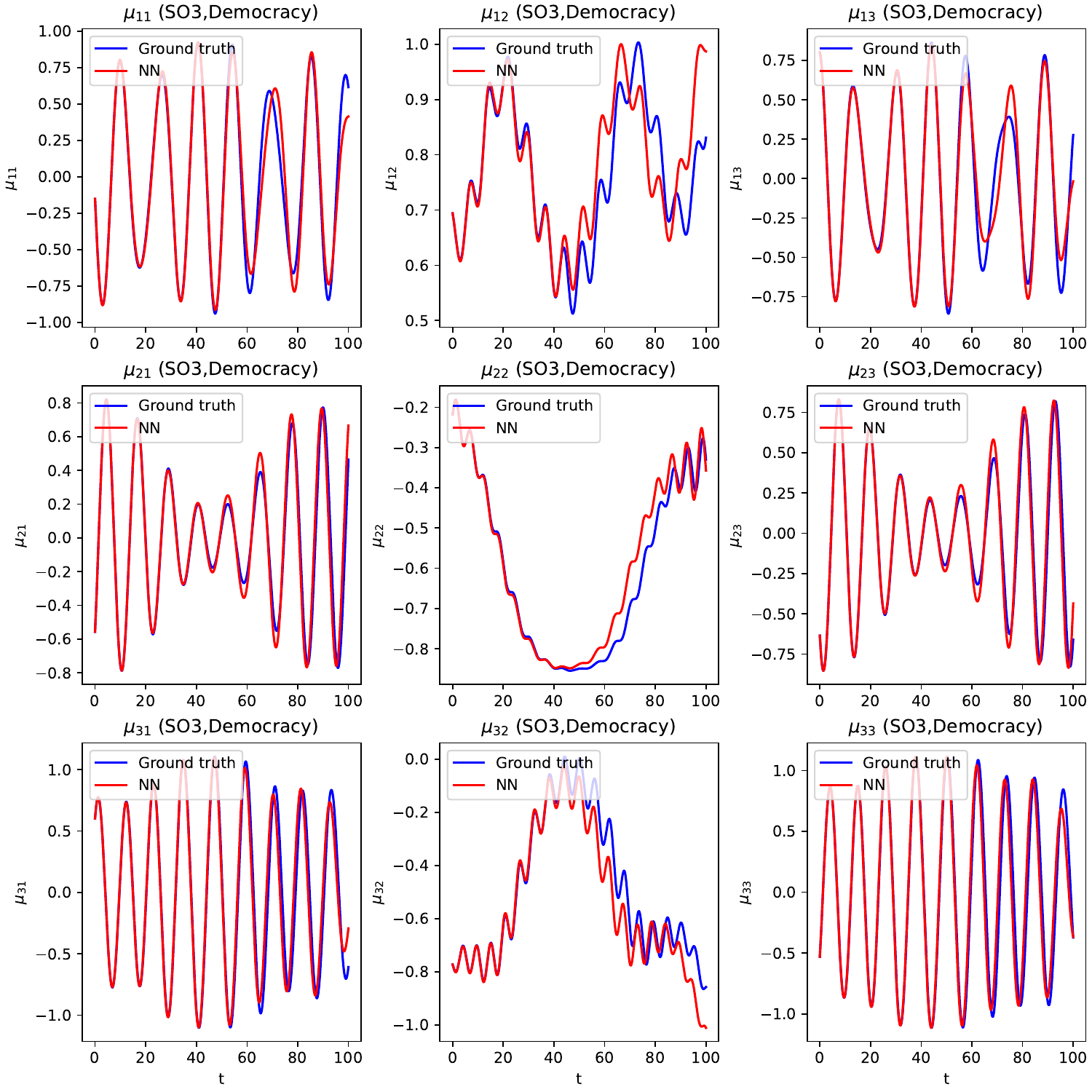}
\caption{Trajectory comparison in the case of three particles for ${\rm SO}(3)$ group, for the case of 'Democracy'. Ground truth case is denoted with blue colour, while CO-LPNets - with red colour, for one representative trajectory. All components $\mu_{k\alpha}$ are plotted, with particle index indicated by the row, and the component $\alpha$ by the column of the table. \label{fig:SO3_test_trajectory_Democracy}}
\end{figure}

\begin{figure} 
\centering 
\includegraphics[width=1\textwidth]{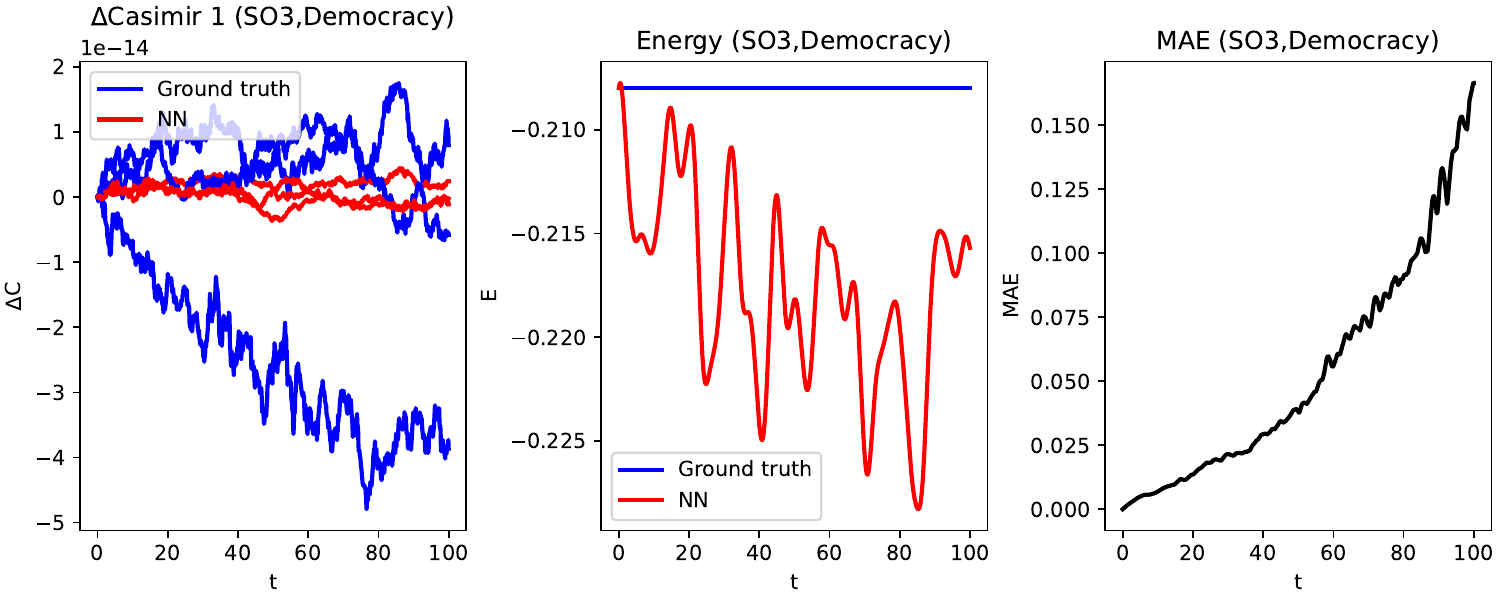}
\caption{Errors in the Casimir for each particle (left), Total energy (center) and Mean Absolute Error (MAE, right) for the trajectory presented on Figure~\ref{fig:SO3_test_trajectory_Democracy}. MAE is computed for all ten reconstructed solutions, not just for the one shown on Figure~\ref{fig:SO3_test_trajectory_Democracy}. Again, ground truth solution is denoted with blue colour, while the results of CO-LPNets are denoted with red colour. \label{fig:SO3_Casimir_Energy_error_Democracy}}
\end{figure}

\begin{figure} 
\centering 
\includegraphics[width=0.9\textwidth]{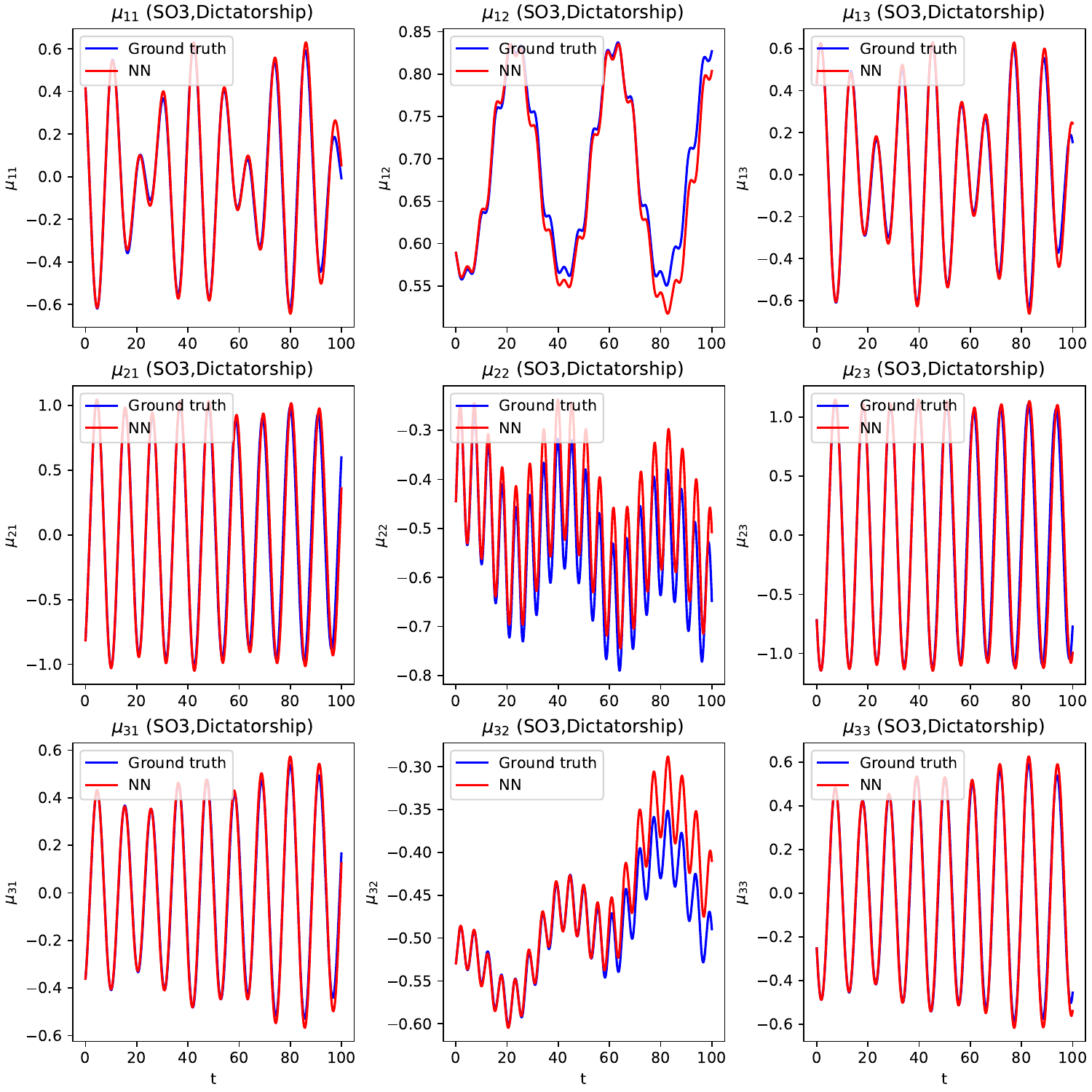}
\caption{
The comparison of solutions for the case of three particles for ${\rm SO}(3)$ group, the 'Dictatorship' case, with all notations and color scheme as in Figure~\ref{fig:SO3_test_trajectory_Democracy}. 
\label{fig:SO3_test_trajectory_dictatorship}}
\end{figure}

\begin{figure} 
\centering 
\includegraphics[width=1\textwidth]{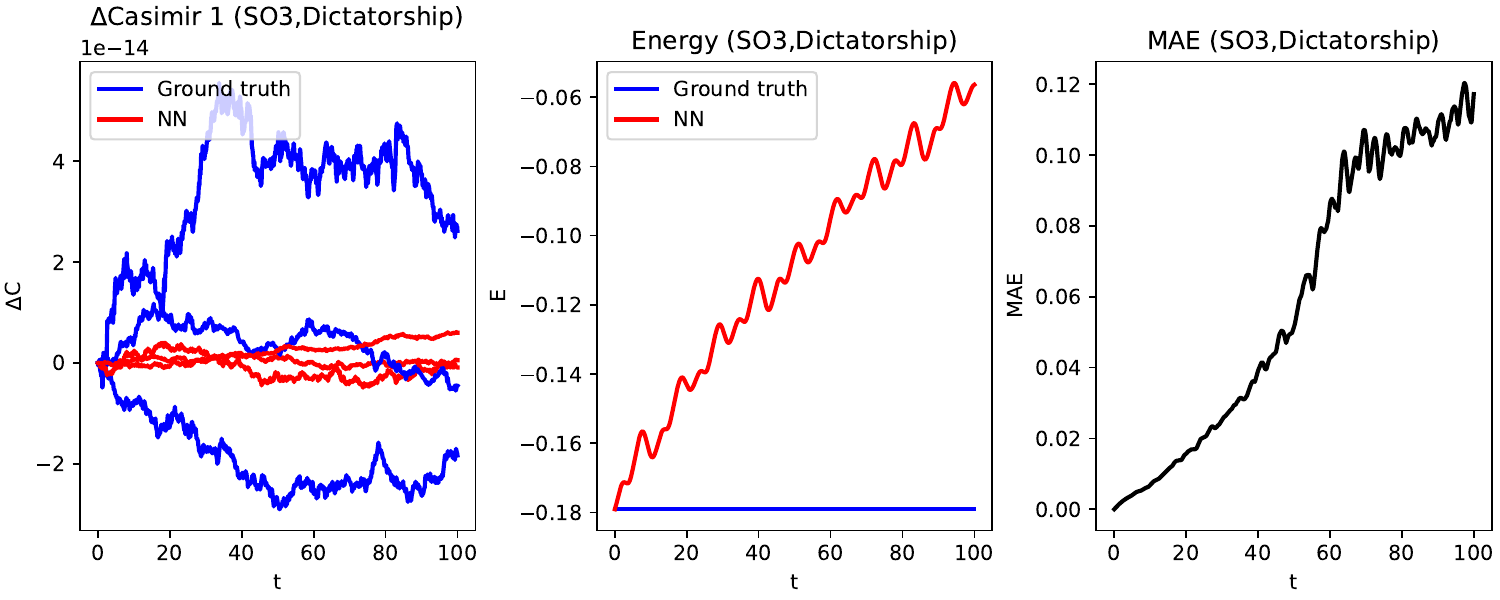}
\caption{Errors in the Casimir (left), Energy (center) and Mean Absolute Error (MAE, right) for the sample trajectory presented on Figure~\ref{fig:SO3_test_trajectory_dictatorship}. Similar to Figure~\ref{fig:SO3_Casimir_Energy_error_Democracy}, MAE is computed over all sample solutions. Color scheme and notations are the same as in Figure~\ref{fig:SO3_Casimir_Energy_error_Democracy}. 
\label{fig:SO3_Casimir_Energy_error_dictatorship}}
\end{figure}

The results of application of CO-LPNets are shown in Figure~\ref{fig:SO3_test_trajectory_Democracy} for the case of 'Democracy', with the corresponding summary of errors in Casimirs,  energy, and MAE of all components in Figure~\ref{fig:SO3_Casimir_Energy_error_Democracy}. The same information is shown on 
Figure~\ref{fig:SO3_test_trajectory_dictatorship} for the case of 'Dictatorship', with the corresponding summary of errors in Casimirs,  energy, and MAE of all components in Figure~\ref{fig:SO3_Casimir_Energy_error_dictatorship}. The trajectories are reproduced faithfully over $1000$ time steps for every particle, the variations of energy in trajectories by CO-LPNets are bounded, and the Casimirs for each particle are preserved with machine precision. 

In all Figures, when we compare the results of the CO-LPNets with the ground truth, we use a consistent color scheme:  the ground truth is marked with a solid blue line, and results obtained from CO-LPNets are presented with a solid red line. 

\subsection{${\rm SE}(3)$ group}
\label{sec:SE(3) group}
\subsubsection{Equations for ground truth calculations}
\paragraph{Definitions} We now proceed to the consideration of equations of motion for the Lie group of rotations and translations. This group gives a highly practical example as it describes the motion of several interacting objects in space, such as drones. 
We start with some definition of the semidirect product group of rotations and translations ${\rm SE}(3)$. For a rotation $\mathbb{Q} \in {\rm SO}(3)$ and translation $\mathbf{v} \in \mathbb{R}^3$, we form a $4 \times 4$ matrix $g \in {\rm SE}(3)$, with corresponding multiplication and inverse 
\begin{equation}
 g = 
    \left(
    \begin{array}{cc}
    \mathbb{Q} & \mathbf{v} \\
    \mathbf{0}^\mathsf{T} & 1 
    \end{array} 
    \right) 
    ; \, 
        g_1 g_2  = 
    \left(
    \begin{array}{cc}
    \mathbb{Q}_1\mathbb{Q}_2 & \mathbb{Q}_1 \mathbf{v}_2 + \mathbf{v}_1 \\
    \mathbf{0}^\mathsf{T} & 1 
    \end{array} 
    \right) 
    ; \,
     g^{-1} = 
    \left(
    \begin{array}{cc}
    \mathbb{Q}^\mathsf{T} & - \mathbb{Q}^\mathsf{T} \mathbf{v} \\
    \mathbf{0}^\mathsf{T} & 1 
    \end{array} 
    \right) \,, 
    \label{SE3_def}
\end{equation} 
where $\mathbf{0}^\mathsf{T}$ is the row of zeros. The Lie algebra of ${\rm SE}(3)$, which we call $\mathfrak{se}(3)$, is six-dimensional. 
We again employ the trace product of two matrices $<X,Y> = \operatorname{Tr} (X^{\mathsf T} Y)$. With this product, we select  the following orthogonal basis Lie algebra $\mathfrak{se}(3)$:
\begin{gather*}
G_{1}=\frac{1}{\sqrt{2}}\left[\begin{matrix}
0&0&1&0\\
0&0&0&0\\
-1&0&0&0\\
0&0&0&0\\\end{matrix}\right], \quad
G_{2}=\frac{1}{\sqrt{2}}\left[\begin{matrix}
0&-1&0&0\\
1&0&0&0\\
0&0&0&0\\
0&0&0&0\\\end{matrix}\right], \quad
G_{3}=\frac{1}{\sqrt{2}}\left[\begin{matrix}
0&0&0&0\\
0&0&-1&0\\
0&1&0&0\\
0&0&0&0\\\end{matrix}\right], \nonumber\\
G_{4}=\left[\begin{matrix}
0&0&0&0\\
0&0&0&1\\
0&0&0&0\\
0&0&0&0\\\end{matrix}\right], \quad
G_{5}=\left[\begin{matrix}
0&0&0&0\\
0&0&0&0\\
0&0&0&1\\
0&0&0&0\\\end{matrix}\right], \quad
G_{6}=\left[\begin{matrix}
0&0&0&1\\
0&0&0&0\\
0&0&0&0\\
0&0&0&0\\\end{matrix}\right], \nonumber\\
\end{gather*}
which is normalized so that $\|G_{j}\|=1$, $j = 1, \ldots, 6$ with respect to the trace norm.
The structure constants for the $\mathfrak{se}(3)$ Lie algebra are then computed as follows:
\begin{gather*}
\Gamma_{61}^{5}=\frac{1}{\sqrt{2}}, \quad \Gamma_{16}^{5}=-\frac{1}{\sqrt{2}}, \quad \Gamma_{62}^{4}=-\frac{1}{\sqrt{2}}, \quad \Gamma_{26}^{4}=\frac{1}{\sqrt{2}}, \quad \Gamma_{43}^{5}=-\frac{1}{\sqrt{2}}, \quad \Gamma_{34}^{5}=\frac{1}{\sqrt{2}},\\
\Gamma_{42}^{6}=\frac{1}{\sqrt{2}}, \quad \Gamma_{24}^{6}=-\frac{1}{\sqrt{2}}, \quad \Gamma_{53}^{4}=\frac{1}{\sqrt{2}}, \quad \Gamma_{35}^{4}=-\frac{1}{\sqrt{2}}, \quad \Gamma_{51}^{6}=-\frac{1}{\sqrt{2}}, \quad \Gamma_{15}^{6}=\frac{1}{\sqrt{2}},\\
\Gamma_{31}^{2}=\frac{1}{\sqrt{2}}, \quad \Gamma_{13}^{2}=-\frac{1}{\sqrt{2}}, \quad \Gamma_{32}^{1}=-\frac{1}{\sqrt{2}}, \quad \Gamma_{23}^{1}=\frac{1}{\sqrt{2}}, \quad \Gamma_{12}^{3}=\frac{1}{\sqrt{2}}, \quad \Gamma_{21}^{3}=-\frac{1}{\sqrt{2}}
\end{gather*}
with all other structure constants equal to 0.

Using the mechanical analogy from the \emph{mechanical} Lie-Poisson systems \cite{holm2009geometric,marsden2013introduction}, let us separate the full momentum vector for each particle $\bmu_k$ into two three-dimensional vectors $\bPi_k$ and $\mathbf{p}_k$ having the meaning of the angular and linear momenta:
\begin{gather}
\mbox{vector of angular momenta:} \quad \boldsymbol{\Pi}_{k}=(\mu_{k1}, \mu_{k2}, \mu_{k3})^{\mathsf T}, \quad k=1,\dots, N \nonumber\\
\mbox{vector of linear momenta:} \quad \boldsymbol{p}_{k}=(\mu_{k4}, \mu_{k5}, \mu_{k6})^{\mathsf T}, \quad k=1,\dots, N. 
\label{notation_SE(3)}
\end{gather}

Thus, according to \eqref{LP_reduced_dynamics_equations}, 
$\Lambda(\breve{\mu})$ is a block-diagonal matrix of the following form:
\begin{gather*}
\Lambda(\breve{\mu})=\frac{1}{\sqrt{2}}\left[\begin{matrix}
\widehat{\mu}_{1}&0&\cdots&0\\
0&\widehat{\mu}_{2}&\cdots&0\\
\vdots&\vdots&\ddots&\vdots\\
0&\cdots&0&\widehat{\mu}_{N}\\\end{matrix}\right]
\end{gather*}
with $k$-th block equal to the following $6 \times 6$ antisymmetric matrix:
\begin{gather}
\widehat{\mu}_{k}=\left[\begin{matrix}
0&-\mu_{k3}&\mu_{k2}&0&-\mu_{k6}&\mu_{k5}\\
\mu_{k3}&0&-\mu_{k1}&\mu_{k6}&0&-\mu_{k4}\\
-\mu_{k2}&\mu_{k1}&0&-\mu_{k5}&\mu_{k4}&0\\
0&-\mu_{k6}&\mu_{k5}&0&0&0\\
\mu_{k6}&0&-\mu_{k4}&0&0&0\\
-\mu_{k5}&\mu_{k4}&0&0&0&0\\\end{matrix}\right]=\left[\begin{matrix} \widehat{\Pi}_{k}&\widehat{p}_{k}\\
\widehat{p}_{k}&\mathbb{0}\end{matrix}\right], \quad k=1,\dots, N,
\label{one_block_of_Poisson_tensor_for_SE(3)_group}
\end{gather}
where $\mathbb{0}$ is a $3 \times 3$ matrix of zeros and $\widehat{\alpha}_{ij} = -\varepsilon_{ijk} \alpha_k$ is, as usual, the hat map between the components of the vector $\boldsymbol{\alpha}$ and the antisymmetric matrix $\widehat{\alpha}$, just as in \eqref{one_block_of_Poisson_tensor_for_SO(3)_group}. The matrix $\widehat{\alpha}$ defined this way posesses the property $\widehat{\alpha} \mathbf{v} = \boldsymbol{\alpha} \times \mathbf{v}$ for any vector $\mathbf{v} \in \mathbb{R}^3$. 
As a result of the decoupled Poisson tensor, there are $2N$ Casimirs, with two Casimirs for each particle: 
\begin{align*}
C_{1,k}=\mu_{k4}^{2}+\mu_{k5}^{2}+\mu_{k6}^{2}= ||\boldsymbol{p}_{k}||^{2}, \quad k=1,\dots, N,\\
C_{2,k}=\mu_{k1}\mu_{k4}+\mu_{k2}\mu_{k5}+\mu_{k3}\mu_{k6}=\boldsymbol{\Pi}_{k} \cdot \boldsymbol{p}_{k}, \quad k=1,\dots, N.
\label{Casimirs_for_SE(3)_group}
\end{align*}
The Lie--Poisson equations \eqref{LP_reduced_dynamics_equations} 
reduce to the well-known form:
\begin{gather*}
\dot{\boldsymbol{\Pi}}_{k}=\frac{1}{\sqrt{2}} \boldsymbol{\Pi_{k}} \times \frac{\partial h}{\partial \boldsymbol{\Pi}_{k}} + \frac{1}{\sqrt{2}} \boldsymbol{p_{k}} \times \frac{\partial h}{\partial \boldsymbol{p}_{k}}, \quad k=1,\dots, N\\
\dot{\boldsymbol{p}}_{k}=\frac{1}{\sqrt{2}} \boldsymbol{p_{k}} \times \frac{\partial h}{\partial \boldsymbol{\Pi}_{k}}, \quad k=1,\dots, N
\end{gather*}
which are the equations for $N$ underwater vehicles when the centers of mass and buoyancy coincide, extending the classical single-vehicle case developed by Kirchhoff  \cite{leonard1997stability,leonard1997stability2,holmes1998dynamics}. 
We assume that the drift is along the first component of linear momentum (the drone moves forward) and there are two controls in angular momentum in the first and second coordinate. The equation \eqref{control_vel} becomes 
\begin{equation}
\xi_{k}=G_{4}+u_{k1}G_{1}+u_{k2}G_{2}, \quad k=1,\dots, N\, . 
\label{control_vel_SE3}
\end{equation}
Using the general formula \eqref{Reduced_ham_explicit}, after some algebra, we compute the explicit formulas for the Hamiltonian for the case of 'Dictatorship' 

\begin{equation}
\begin{aligned}
\hspace{-2mm}h = \sum_{k=1}^{N}\mu_{k4}  & +\frac{1}{2}\biggl(\frac{1+2\chi}{1+2N\chi} \left(  \mu_{11}^{2}+\mu_{12}^{2}\right) 
 +\frac{1+2N\chi+4\chi^{2}}{(1+2N\chi)(1+2\chi)} \sum_{k=2}^N \left( \mu_{k1}^{2}+ \mu_{k2}^{2} \right)  \\
&+\frac{4\chi}{1+2N\chi}\biggl(\mu_{11}\sum_{k=2}^N\mu_{k1}+\mu_{12}\sum_{k=2}^N\mu_{k2}\biggr)
\\
& +\frac{8\chi^{2}}{(1+2N\chi)(1+2\chi)}\sum_{i,j=2, i<j}^{N}\left( \mu_{i1}\mu_{j1} +\mu_{i2}\mu_{j2}\right)\biggr) 
\label{h_SE3_dictatorship}
\end{aligned}
\end{equation}

For the case of 'Democracy', the Hamiltonian is given by 
\begin{equation}
\begin{aligned}
h=\sum_{k=1}^{N}\mu_{k4} 
& +\frac{1}{2}\biggl(\frac{1+2\chi}{1+2N\chi} \sum_{k=1}^N \left( \mu_{k1}^{2}+ \mu_{k2}^{2}\right) 
\\
& +\frac{4\chi}{1+2N\chi}\sum_{i,j=1, i<j}^{N} \left( \mu_{i1}\mu_{j1}+\mu_{i2}\mu_{j2} \right) \biggr).
\label{h_SE3_Democracy}
\end{aligned}
\end{equation}

The forms for the Hamiltonian \eqref{h_SE3_dictatorship} and \eqref{h_SE3_Democracy} are based on the explicit form of the matrix $\Psi$, which we present in Section~\ref{app_sec:SE3_exact_deriv}. 

\paragraph{Data generation for learning} We generate data for two cases, 'Dictatorship' and 'Democracy', by simulating the system with $N=3$ particles, creating $80$ trajectories of length $51$, thus creating $4000$ intervals for learning of the phase space. The initial conditions for these trajectories are also chosen randomly in the hypercube $[-1,1]^d$, where $d$, the dimension of the phase space, is now equal to $N n=18$ in this case. Again, as mentioned in Section~\ref{sec:Producing_data_in_phase_space}, we use a Lie-Poisson integrator to compute the ground truth data, which preserves the system's structure and Casimirs with machine precision. We choose the value of the parameter $\chi =0.5$ for all simulations. We also present analytical formulas for the matrices $\Psi$ and derivatives of the Hamiltonians in Appendix~\ref{app_sec:SE3_exact_deriv}.

Next, we construct Poisson transformations as flows generated by test Hamiltonians in order to create Poisson transformations for CO-LPNets.

\subsubsection{Poisson transformations through test Hamiltonians} 
To construct the Poisson transformations for CO-LPNets in the case of ${\rm SE}(3)$ group we consider Hamiltonians which depend only on a single component $i$, $1\leq i \leq 6$ of one particle $k$, $1 \leq k \leq N$; more precisely, we take $h_{ki} = w_{ki} \mu_{ki}$ To construct the Poisson transformations explicitly, we consider two possibilities: $1 \leq i \leq 3$ (test Hamiltonian corresponding to the angular momentum) and $4 \leq i \leq 6$ (test Hamiltonian corresponding to the linear momentum). 
\paragraph{Test Hamiltonians for angular momenta} This is the case $h_{ki} = w_{ki} \mu_{ki}$ with $1 \leq i \leq 3$. 
Equations  \eqref{LP_reduced_dynamics_equations} simplify as follows: 
\begin{equation}
\begin{aligned}
& \dot{\boldsymbol{\Pi}}_{k} = \boldsymbol{\Pi}_{k} \times \mathbf{e}_{i} w_{ki}, \quad 
\dot{\boldsymbol{p}}_{k} = \boldsymbol{p}_{k} \times \mathbf{e}_{i} w_{ki},
\\
& \dot{\boldsymbol{\Pi}}_{j} = \mathbf{0} \, , \quad 
\dot{\mathbf{p}}_{j} = \mathbf{0}\,, \quad 
j \neq k \, . 
\end{aligned}
\label{test_transformations_SE3_angmom}
\end{equation}
with $\bPi_k$ and $\mathbf{p}_k$ defined as in \eqref{notation_SE(3)}. 
Again, $\mathbf{e}_{i}$ means the $i$-th standard basis vector in the 3-dimensional space; $w_{ki}$ is a constant on every time step that will need to be approximated by the neural network. Suppose we are given initial conditions for $\bPi(t)$ and $\mathbf{p}(t)$ at the point $t=0$, which we denote as   $\boldsymbol{\Pi}_{k}(0)$ and $\boldsymbol{p}_{k}(0)$. The solution of the initial value problem \eqref{test_transformations_SE3_angmom} is then given by 
\begin{equation}
\begin{aligned}
    \mathbf{\Pi}_{k}(t) & = \mathbb{R}(\mathbf{e}_{i}, -w_{ki}t) \boldsymbol{\Pi}_{k}(0) \, , \quad  \mathbf{p}_{k}(t) = \mathbb{R}(\mathbf{e}_{i}, - w_{ki}t) \boldsymbol{p}_{k}(0) , 
    \\
        \mathbf{\Pi}_{j}(t) & =  \boldsymbol{\Pi}_{j}(0) \, , \quad  \mathbf{p}_{j}(t) =  \boldsymbol{p}_{j}(0) \, , \quad 
        j \neq k \, . 
    \end{aligned} 
    \label{sol_test_Ham_SE3_angmom}
\end{equation} 
\paragraph{Test Hamiltonians for linear momenta} This is the case $h_{ki} = w_{ki} \mu_{ki}$ with $4 \leq i \leq 6$.
Equations  \eqref{LP_reduced_dynamics_equations} are now written as follows: for the $k$th particle, we get the system of equations:
\begin{equation}
\begin{aligned}
& \dot{\boldsymbol{\Pi}}_{k} = \mathbf{p}_{k} \times \mathbf{e}_{i-3} w_{ki}, \quad 
\dot{\mathbf{p}}_{k} = \mathbf{0},
\\
& \dot{\boldsymbol{\Pi}}_{j} = \mathbf{0} \, , \quad 
\dot{\mathbf{p}}_{j} = \mathbf{0}\,, \quad 
j \neq k \, . 
\end{aligned}
\label{test_transformations_SE3_linmom}
\end{equation}
with the solution 
\begin{equation}
\begin{aligned}
    \mathbf{\Pi}_{k}(t) & =  \mathbf{p}_{k} \times \mathbf{e}_{i-3} w_{ki} t + \boldsymbol{\Pi}_{k}(0) \, , \quad  \mathbf{p}_{k}(t) =  \boldsymbol{p}_{k}(0) , 
    \\
        \mathbf{\Pi}_{j}(t) & =  \boldsymbol{\Pi}_{j}(0) \, , \quad  \mathbf{p}_{j}(t) =  \boldsymbol{p}_{j}(0) \, , \quad 
        j \neq k \, . 
    \end{aligned} 
    \label{sol_test_Ham_SE3_linmom}
\end{equation} 

The transformations \eqref{sol_test_Ham_SE3_angmom} and 
\eqref{sol_test_Ham_SE3_linmom} can be written as block-diagonal matrices with $j \neq k$ blocks being identities, and $k$-th block being a combination of rotations and translations. For brevity, we do not present them here. For the computations of the loss function \eqref{Loss_function_def}, we present partial derivatives of these transformations with respect to parameters $w_{ki}$ in Section~\ref{app_sec:SE3_exact_deriv} of the Appendix.

\subsubsection{Results from the application of CO-LPNets for particles evolving on ${\rm SE}(3)$ }
\label{sec:SE3_results}
We generate the data in phase space, as was described in
\eqref{sec:Producing_data_in_phase_space}. We consider $N=3$ coupled particles each evolving in ${\rm SE}(3)$, having dimension of $n=6$ with the total dimension of the phase space being $N n = 18$. We generate $4,000$ points, to match the number of points per dimension with the ${\rm SO}(3)$ case described in Section~\ref{sec:SO3_results}, with  $\sim 222$ points per dimension of the phase space. Again, the time step is chosen to be $\Delta t =0.1$.  We produce data for learning in two cases: 'Dictatorship' and 'Democracy'. We then apply CO-LPNets with $18$ rotations, the minimum number of rotations allowed to cover all possible motions in space, and optimize the loss defined by \eqref{Loss_function_def}. We perform the optimization using the Adam algorithm with a learning rate of $0.005$ for 10,000 epochs. During that time, the loss decreases from about $\sim 0.1$ to $\sim 10^{-5}$, \emph{i.e.} by four orders of magnitude. After the optimization is finished, we simulate ten trajectories starting with random initial conditions in the cube $[-1,1]^{N n}$, both using the structure-preserving integrator, the CO-LPNets with computed parameters. 

\begin{figure} 
\centering 
\includegraphics[width=0.9\textwidth]{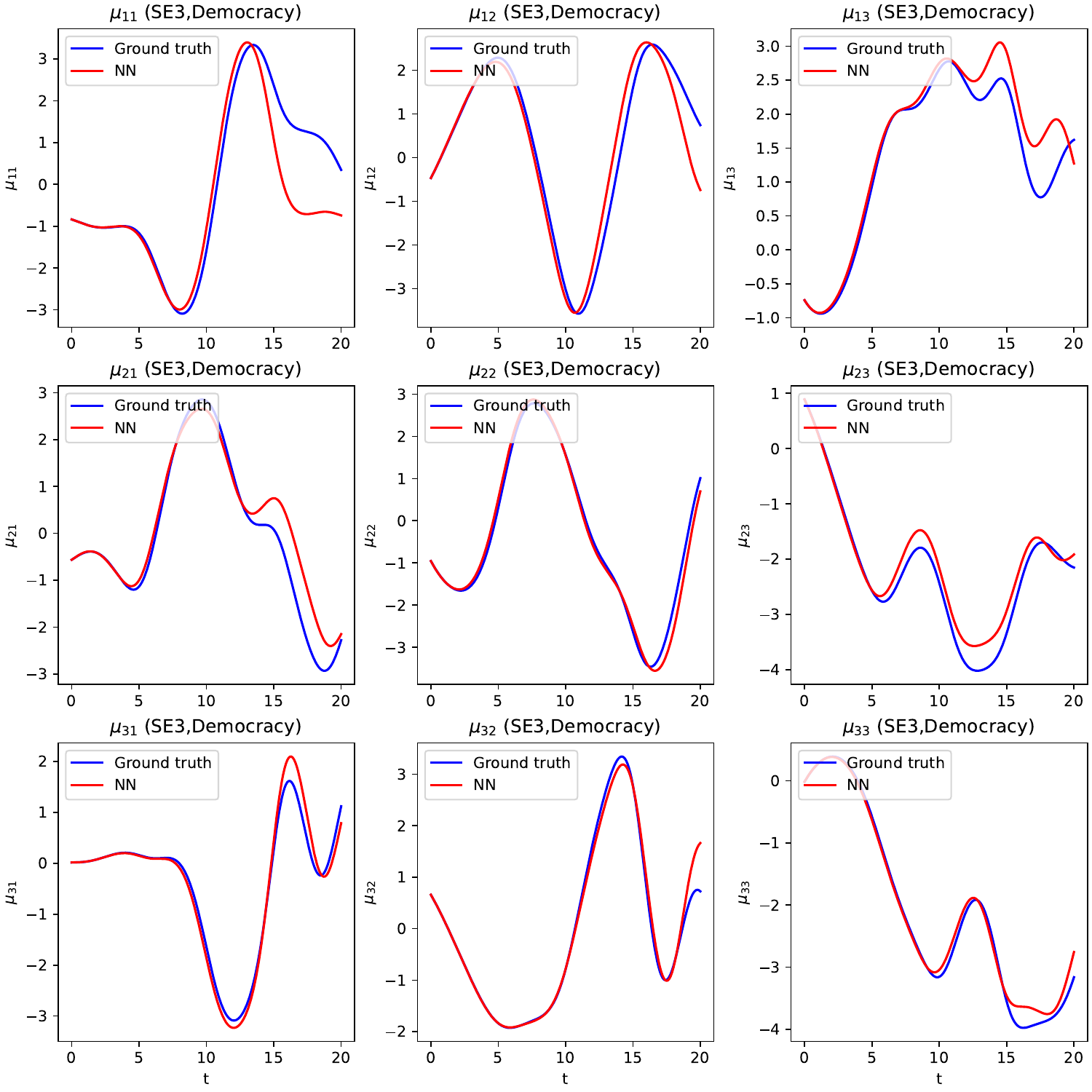}
\caption{Trajectory comparison in the case of three particles for ${\rm SE}(3)$ group, for the case of 'Democracy'. Ground truth case is denoted with blue colour, while CO-LPNets - with red colour, for one representative trajectory. Only the first three components $\mu_{k\alpha}$, $\alpha=1,2,3$ are plotted, with particle index $k=1,2,3$ indicated by the row, and the component $\alpha$ by the column of the table. \label{fig:SE3_test_trajectory_Democracy}}
\end{figure}

\begin{figure} 
\centering 
\includegraphics[width=1\textwidth]{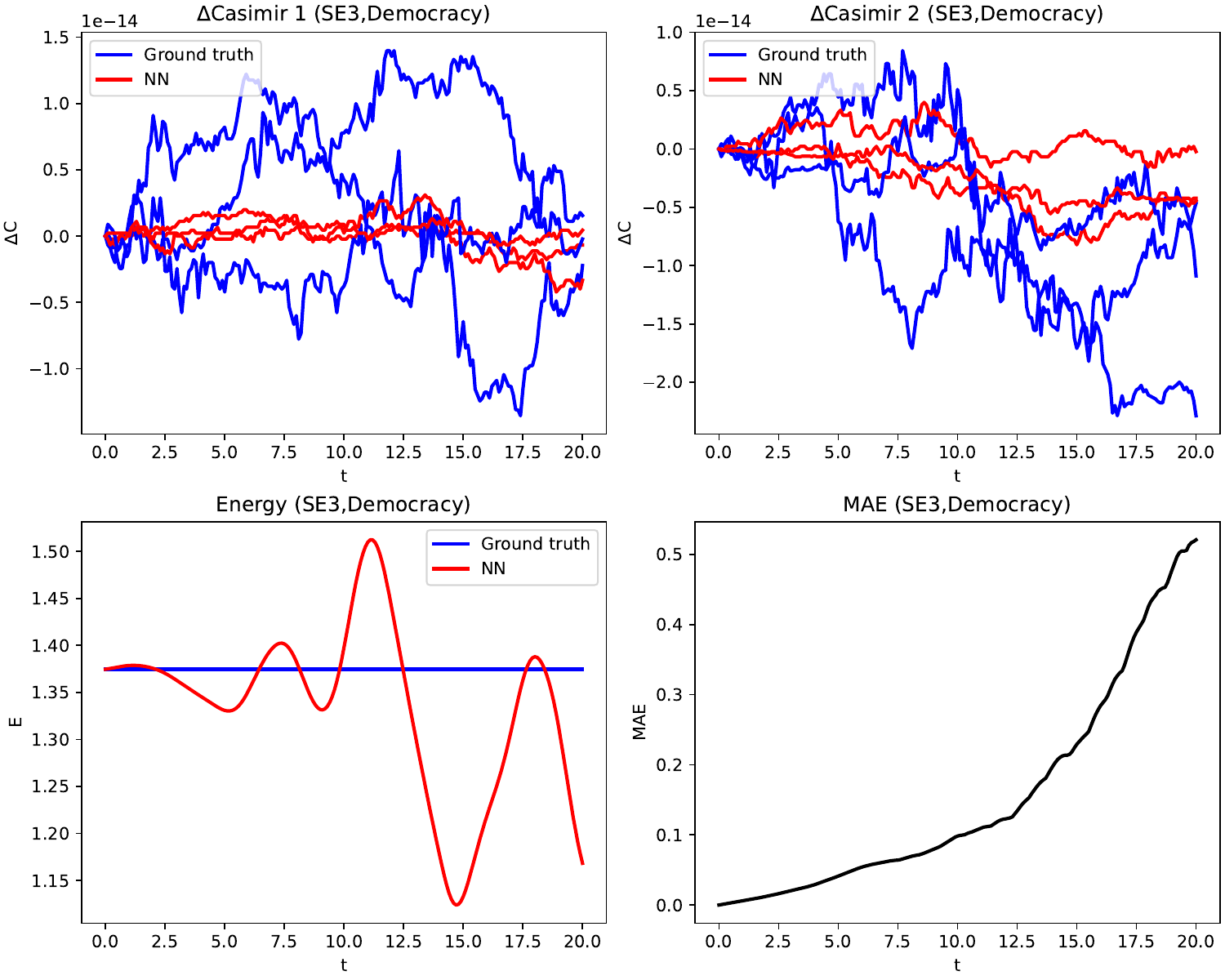}
\caption{Conservation of the Casimirs (top), Errors in the energy (bottom left) for a sample trajectory, Mean Absolute Error (MAE) for all ten test trajectories (bottom right), for the case of 'Democracy' for ${\rm SE}(3)$. Ground truth case is denoted with blue colour, while CO-LPNets are denoted with red colour. Notice, again, preservation of the Casimirs with machine precision, and quite accurate conservation of the energy on average by CO-LPNets.\label{fig:SE3_Casimir_Energy_error_Democracy} }
\end{figure}

\begin{figure} 
\centering 
\includegraphics[width=0.9\textwidth]{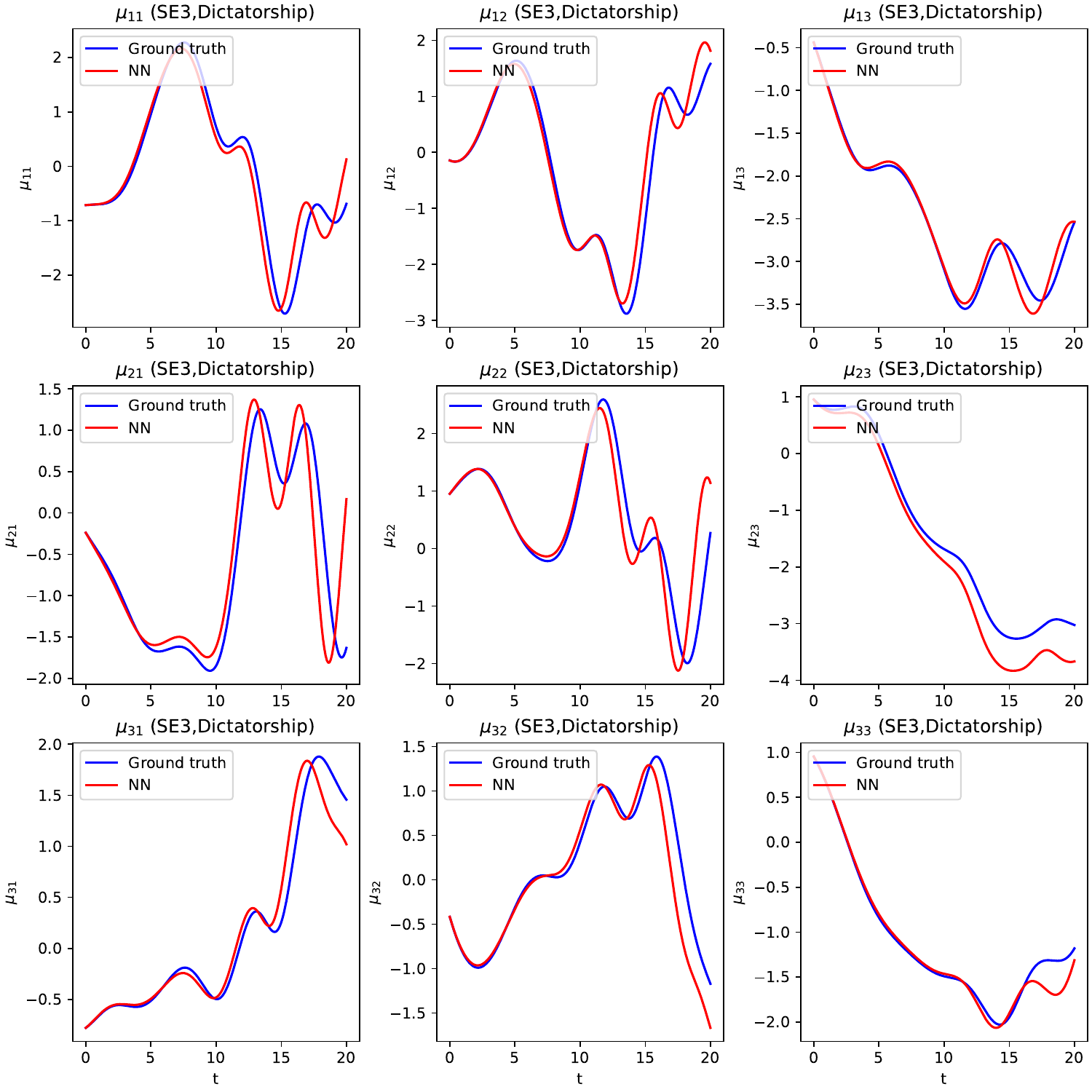}
\caption{Trajectory comparison in the case of three particles for ${\rm SE}(3)$ group, for the case of dictatorship.  Only first three components of six are plotted for every particle, with the number of the particle $k=1,2,3$ corresponding to the row in the table, and the component $\alpha=1,2,3$ to the column of the table. Color scheme and notations are identical to Figure~\ref{fig:SE3_test_trajectory_Democracy}.
\label{fig:SE3_test_trajectory_dictatorship}
}
\end{figure}

\begin{figure} 
\centering 
\includegraphics[width=1\textwidth]{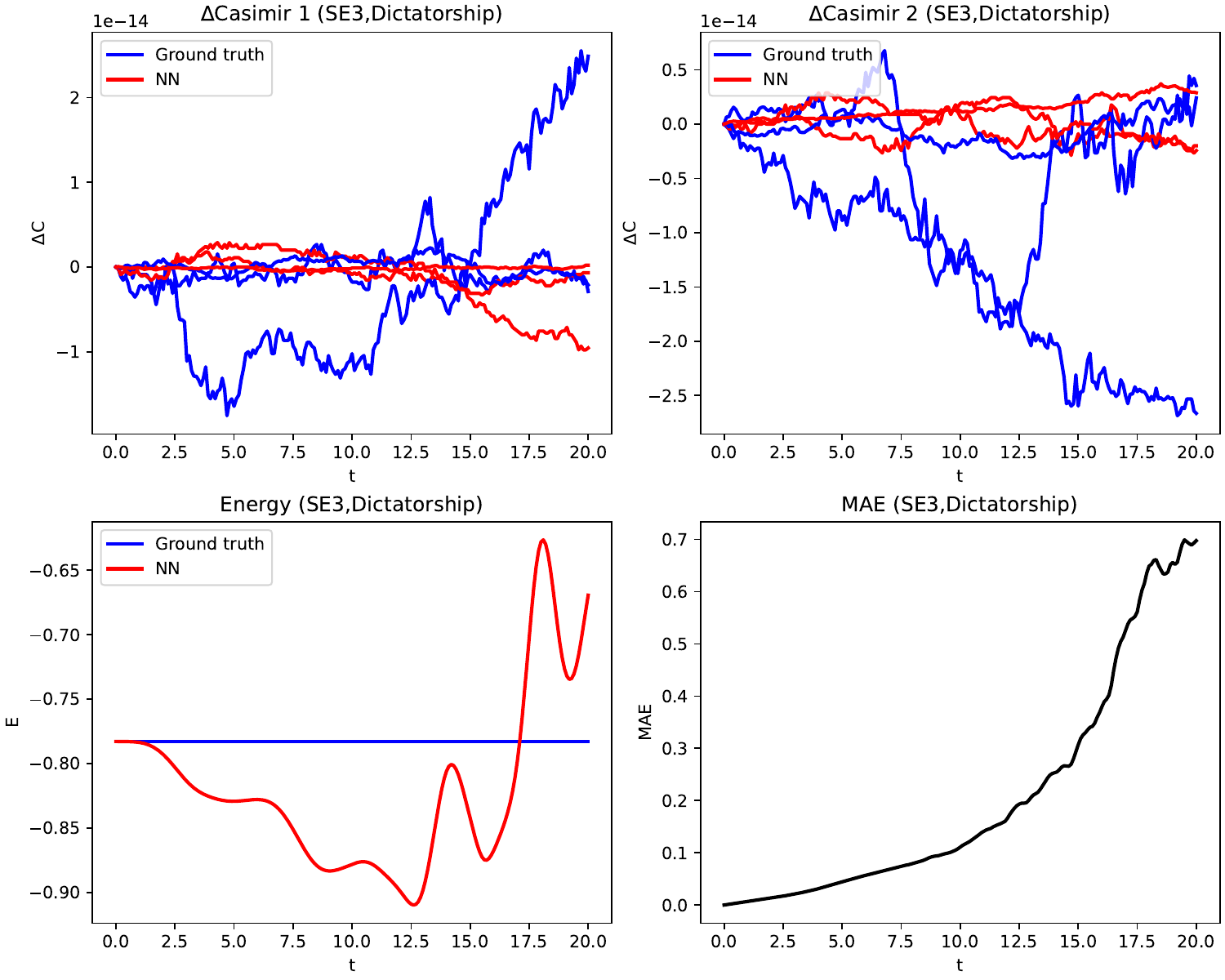}
\caption{Summary of 'Dictatorship' governance for three particles evolving on ${\rm SE}(3)$. Error in Casimirs (top), Errors in the energy (bottom left) and  Mean Absolute Error (MAE) for all ten test trajectories (bottom right).   Both Casimirs are conserved with machine precision for all particles, as expected. Color scheme and notations are identical to Figure~\ref{fig:SE3_Casimir_Energy_error_Democracy}. 
\label{fig:SE3_Casimir_Energy_error_dictatorship}}
\end{figure}

The results of application of CO-LPNets are shown in Figure~\eqref{fig:SE3_test_trajectory_Democracy} for the case of 'Democracy', with the corresponding summary of errors in Casimirs,  energy, and MAE of all components in Figure~\ref{fig:SE3_Casimir_Energy_error_Democracy}. The same information is shown on 
Figure~\eqref{fig:SE3_test_trajectory_dictatorship} for the case of , with the corresponding summary of errors in Casimirs,  energy and MAE of all components in Figure~\ref{fig:SE3_Casimir_Energy_error_dictatorship}. We see that CO-LPNets faithfully reproduce the trajectory over an intermediate time scale (200 steps), preserving Casimirs with machine precision. The Hamiltonian (energy) is not conserved by CO-LPNets, but the deviations of energy are bounded. 

\section[Conclusions]{Conclusions and future work}
\label{sec:Conclusions}
The main contribution of this paper is to present a novel method of learning the full phase space behavior of general Lie-Poisson systems, in particular, Lie-Poisson systems important for optimal control. Our systems are efficient as they use a relatively small number of parameters in the model and a small number of data points. We discuss the completeness of the transformations and connect our work to the established field of structure-preserving splitting methods. We call our method CO-LPNets. The illustrative examples of predicting the evolution of interacting particles in ${\rm SO}(3)$ and ${\rm SE}(3)$ group shows the potential of our method, and possible applications to higher-dimensional systems.  CO-LPNets preserve Poisson structure, and thus conserve Casimirs with machine precision. 

In the future, we would like to extend the current progress to the control of interacting particles on ${\rm SU}(2)$ group, essential for quantum computing. Another problem is the interaction of several particles evolving on different groups, with one group being a subgroup of another group. For example, a flying drone evolves on ${\rm SE}(3)$ group, and a land-based or a sea-surface drone evolves on ${\rm SE}(2)$  group, which is a subgroup of ${\rm SE}(3)$. Thus, we can consider the evolution of such a control system on the direct group product ${\rm SE}(3) \times  {\rm SE}(2)$. Theoretical studies for such work have been undertaken in \cite{bloch2017optimal}; it would be interesting if our machine learning methods extend to these more general, and practically important, cases. 

\section*{Acknowledgements}
We are grateful to Anthony Bloch, Denys Cherhykalo, Chris Eldred, Fran\c{c}ois Gay-Balmaz, Tristan Griffith, Anthony Gruber, Melvin Leok, Andrew Sinclair, Irina K. Tezaur, and  Dmitry Zenkov for fruitful and engaging discussions. SH was mainly supported by the Vanier Canada Graduate Scholarships (Vanier CGS) awarded by the Natural Sciences and Engineering Research Council (NSERC). 
VP and SH were partially supported by the NSERC Discovery grant.

\section*{Use of AI in this work}
No AI was used in the writing of the text of this manuscript. 

Part of the work designing the code was done using AI-assisted coding in order to evaluate the possibility of generating the results by a reader who wants to implement our method with a Large Language Model (LLM).  After writing initial codes in \emph{Numpy} for $N=2$ particles, we produced a program generating the data and producing the learning procedure for CO-LPNets using AI-assisted coding. We used the Large Language Model \emph{Gemini}, Pro version, to produce the code. A short summary of the experience is as follows: while some parts of the code were written correctly by LLM, the LLM has made numerous errors in deriving the analytical formulas for Poisson transformations, especially in constructing the Poisson tensor. The code had to be thoroughly checked for errors and misunderstandings by the LLMs, which resulted, for example, in non-conservation of energy and Casimirs for the ground truth data, and erroneous responses by the LLM to the nature of the problem. A more detailed description of the prompts used in programming is presented in Section~\ref{app:sec_LLM} in the Appendix. The conclusion is that at this point, AI-assisted programming of our method is possible, but needs to be used after a thorough understanding of the nature of the mathematics underpinning the method, and the algorithm has been achieved by the user.  

\section*{Conflict of interests} The authors declare that, to the best of their knowledge, they have no competing financial or personal interests that could have affected the work reported in this article.

\bibliographystyle{plain}
\bibliography{bibliography.bib}

\appendix
\section{Lie--Poisson reduced dynamics for a single particle}
\label{sec:single_particle}
\subsection{Single particle equations in the general case}
This section computes the equations of motion for the single particle, $N=1$. It is clearly not a coupled case, but it also has its own pedagogical merit. In this case, the Poisson tensor will comprise only one block and, thus, will simplify \eqref{Poisson_tensor_one_particle} for a single particle:

\begin{align}
\Lambda=\widehat{\mu}=-\sum_{s=1}^{n} \mu_{1s}  \left[\begin{matrix}
\Gamma_{11}^{s}&\Gamma_{12}^{s}&\cdots&\Gamma_{1n}^{s}\\[3pt]
\Gamma_{21}^{m}&\Gamma_{22}^{s}&\cdots&\Gamma_{2n}^{s}\\[3pt]
\vdots&\vdots&\ddots&\vdots\\[3pt]
\Gamma_{n1}^{s}&\Gamma_{n2}^{s}&\cdots&\Gamma_{nn}^{s}
\end{matrix}\right],
\label{Poisson_tensor_one_particle_init}
\end{align}
where $\Gamma_{ij}^{s}$, again, denote structure constants for the corresponding Lie algebra~$\mathfrak{g}$.
Moreover, now the matrix $\mathbb{I}_{N}=\mathbb{I}_{1}=1$, and we can take $B=0$.
Then $(\mathbb{I}_{N}+2\chi B)=1$, \quad $(\mathbb{I}_{N}+2\chi B)^{-1}=1$. Finally,
\begin{gather}
\Psi=(\mathbb{I}_{N}+2\chi B)^{-1}\otimes \mathbb{I}_{m}=1\otimes \mathbb{I}_{m}=\mathbb{I}_{m}.\label{matrix_PSI_one_particle}
\end{gather}
According to \eqref{Reduced_ham_explicit}, the Hamiltonian looks as follows, since the particle index always equals $k=N=1$:
\begin{align}
h =
\rem{
{}&\sum_{k=1}^{N=1}\mu_{kq}+\frac{1}{2}\begin{bmatrix}\tilde{\mu}_{1} & \cdots & \tilde{\mu}_{N}\end{bmatrix}\Psi\left[\begin{matrix}
\tilde{\mu}_{1}\\
\vdots\\
\tilde{\mu}_{N}\end{matrix}\right]
}
\mu_{1q}+
\frac{1}{2} \sum_{j=1}^m  \mu_{1j}^2 \, , 
\label{Hamiltonian_one_particle}
\end{align}
where $m$ is the number of controls. Reference \cite{justh2015optimality} considered the case of ${\rm SE}(2)$ group; we extend these considerations for ${\rm SO}(3)$ and ${\rm SE}(3)$ groups. 
\subsection{The case of single particle dynamics evolving on  ${\rm SO}(3)$}
\label{app:sec_SO3_single_particle}
In this Section, we consider the Lie--Poisson reduced dynamics for a single particle evolving on ${\rm SO}(3)$ group. We use $m=1$ (one control in the first element) and $q=2$ (drift in the second element). 
With the use of \eqref{Poisson_tensor_one_particle} and in terms of notation \eqref{notation_SO(3)} for k=1, the Poisson tensor is
\begin{gather*}
\Lambda=\frac{1}{\sqrt{2}}\left[\begin{matrix}
0&-\mu_{3}&\mu_{2}\\
\mu_{3}&0&-\mu_{1}\\
-\mu_{2}&\mu_{1}&0\\\end{matrix}\right]=\frac{1}{\sqrt{2}}\widehat{ \mu}.
\end{gather*}
\rem{ 
With the use of \eqref{matrix_PSI_one_particle}:
\begin{gather*}
\Psi=\mathbb{I}_{1}=1.
\end{gather*}
}
According to \eqref{Hamiltonian_one_particle}, the Hamiltonian becomes
\begin{equation}
\label{Ham_single_particle_SO3}
h =\mu_{2}+\frac{1}{2}\mu_{1}^{2}.
\end{equation}
The Casimir function is $c=\mu_{1}^{2}+\mu_{2}^{2}+\mu_{3}^{2}=||\boldsymbol{\mu}||^{2}$ which clearly satisfies $\Lambda \nabla c=0$.

For this single particle problem, the Lie--Poisson reduced dynamics equations $\dot{\breve{\mu}}=\Lambda(\breve{\mu})\nabla h$ \eqref{LP_reduced_dynamics_equations}, where $\nabla h=(\mu_{1}, 1 , 0)^T$, are given by
\begin{align*}
\frac{d}{dt}
\left[\begin{matrix}
{\mu}_{1}\\
{\mu}_{2}\\
{\mu}_{3}\\
\end{matrix}\right]=\frac{1}{\sqrt{2}}\left[\begin{matrix}
0&-\mu_{3}&\mu_{2}\\
\mu_{3}&0&-\mu_{1}\\
-\mu_{2}&\mu_{1}&0\\\end{matrix}\right]
\left[\begin{matrix}
\mu_{1}\\
1\\
0\\\end{matrix}\right]
\end{align*}
or, equivalently,
\begin{align}
\begin{cases}
\dot{\mu}_{1}=-\frac{1}{\sqrt{2}}\mu_{3}, \\
\dot{\mu}_{2}=\frac{1}{\sqrt{2}}\mu_{1}\mu_{3}, \\
\dot{\mu}_{3}=-\frac{1}{\sqrt{2}}\mu_{1}\mu_{2}+\frac{1}{\sqrt{2}}\mu_{1}.\label{SO(3)_group_equation_one_particle}
\end{cases}
\end{align}
One verifies that $h$ is a conserved quantity, as we have
\begin{align*}
\dot{h}=\dot{\mu}_{2}+\mu_{1}\dot{\mu}_{1}=\frac{1}{\sqrt{2}}\mu_{1}\mu_{3}-\mu_{1}\frac{1}{\sqrt{2}}\mu_{3}=0.
\end{align*}
Hence, the Hamiltonian and the Casimir function are conserved quantities.
From \eqref{SO(3)_group_equation_one_particle} we obtain the second-order equation
\begin{align*}
\ddot{\mu}_{1}=-\frac{1}{\sqrt{2}}\dot{\mu}_{3}=\frac{1}{2}\mu_{1}(\mu_{2}-1)=\frac{1}{2}\mu_{1}\left((h-1)-\frac{1}{2}\mu_{1}^{2}\right),
\end{align*}
that is,
\begin{equation}
\ddot{\mu}_{1}-\frac{(h-1)}{2}\mu_{1}+\frac{1}{4}\mu_{1}^{3}=0.
\label{Jacobi_SO3}
\end{equation}
In fact, we observe that \eqref{Jacobi_SO3} is the same as equation~(36) in \cite{justh2010extremal} and eq.~(3.24) in \cite{justh2015optimality} (which, we remind the reader, was written for ${\rm SE}(2)$ group), with the only difference $h \rightarrow (h-1)$. Equation \eqref{Jacobi_SO3} can be solved in terms of Jacobi elliptic functions. Interestingly enough, \eqref{Jacobi_SO3} also appears in the theory of elasticity \cite{jurdjevic1997}.

\subsection{The case of ${\rm SE}(3)$ with a particular Hamiltonian allowing analytical solution}

\paragraph{Lie--Poisson reduced dynamics for a single particle}
\label{app:sec_SE3_single_particle}
In this Section, we will introduce the Lie--Poisson reduced dynamics for a single particle, evolving on $ {\rm SE}(3)$ group. We use controls in the first two coordinates (rotations) and drift in 4-th coordinate (first linear momentum). 
Using \eqref{Poisson_tensor_one_particle} and notation \eqref{notation_SE(3)} for $k=1$, the Poisson tensor is
\begin{gather*}
\Lambda=\frac{1}{\sqrt{2}}\left[\begin{matrix}
0&-\mu_{3}&\mu_{2}&0&-\mu_{6}&\mu_{5}\\
\mu_{3}&0&-\mu_{1}&\mu_{6}&0&-\mu_{4}\\
-\mu_{2}&\mu_{1}&0&-\mu_{5}&\mu_{4}&0\\
0&-\mu_{6}&\mu_{5}&0&0&0\\
\mu_{6}&0&-\mu_{4}&0&0&0\\
-\mu_{5}&\mu_{4}&0&0&0&0\end{matrix}\right]=\frac{1}{\sqrt{2}}\left[\begin{matrix} \widehat{\Pi}&\widehat{p}\\
\widehat{p}&\mathbb{0}\end{matrix}\right].
\end{gather*}
With the use of \eqref{matrix_PSI_one_particle}:
\begin{gather*}
\Psi=\mathbb{I}_{2}=\left[\begin{matrix}
1&0\\
0&1\\\end{matrix}\right].
\end{gather*}
According to \eqref{Hamiltonian_one_particle}, the Hamiltonian looks like
\begin{equation}
h 
\rem{ {}&\mu_{4}+\frac{1}{2}\begin{bmatrix}\tilde{\mu}_{1} & \cdots & \tilde{\mu}_{N}\end{bmatrix}\Psi\left[\begin{matrix}
\tilde{\mu}_{1}\\
\vdots\\
\tilde{\mu}_{N}\end{matrix}\right]=\mu_{4}+\frac{1}{2}\begin{bmatrix}\mu_{1} & \mu_{2}\end{bmatrix} \left[\begin{matrix}
1&0\\
0&1\\\end{matrix}\right] \left[\begin{matrix}
\mu_{1}\\
\mu_{2}\end{matrix}\right]\\
}
=\mu_{4}+\frac{1}{2}\mu_{1}^{2}+\frac{1}{2}\mu_{2}^{2}.
\label{h_single_particle_SE3}
\end{equation}
There are two Casimir functions:  $c_{1}=\mu_{4}^{2}+\mu_{5}^{2}+\mu_{6}^{2}= ||\boldsymbol{p}||^{2}$ and $c_{2}=\mu_{1}\mu_{4}+\mu_{2}\mu_{5}+\mu_{3}\mu_{6}=\boldsymbol{\Pi} \cdot \boldsymbol{p}$ which clearly satisfy $\Lambda \nabla c_{1}=0$ and $\Lambda \nabla c_{2}=0$ .

For this single particle problem the Lie--Poisson reduced dynamics equations $\dot{\breve{\mu}}=\Lambda(\breve{\mu})\nabla h$  \eqref{LP_reduced_dynamics_equations}, where $\nabla h=(\mu_{1} , \mu_{2}, 0 , 1 , 0 , 0)^T$, are given by
\begin{align*}
\frac{d}{dt}
\left[\begin{matrix}
{\mu}_{1}\\
{\mu}_{2}\\
{\mu}_{3}\\
{\mu}_{4}\\
{\mu}_{5}\\
{\mu}_{6}\\
\end{matrix}\right]=\frac{1}{\sqrt{2}}\left[\begin{matrix}
0&-\mu_{3}&\mu_{2}&0&-\mu_{6}&\mu_{5}\\
\mu_{3}&0&-\mu_{1}&\mu_{6}&0&-\mu_{4}\\
-\mu_{2}&\mu_{1}&0&-\mu_{5}&\mu_{4}&0\\
0&-\mu_{6}&\mu_{5}&0&0&0\\
\mu_{6}&0&-\mu_{4}&0&0&0\\
-\mu_{5}&\mu_{4}&0&0&0&0\end{matrix}\right]\left[\begin{matrix}
\mu_{1}\\
\mu_{2}\\
0\\
1\\
0\\
0\\\end{matrix}\right]
\end{align*}
or, equivalently,
\begin{align}
\begin{cases}
\dot{\mu}_{1}=-\frac{1}{\sqrt{2}}\mu_{2}\mu_{3}, \\
\dot{\mu}_{2}=\frac{1}{\sqrt{2}}\mu_{1}\mu_{3}+\frac{1}{\sqrt{2}}\mu_{6}, \\
\dot{\mu}_{3}=-\frac{1}{\sqrt{2}}\mu_{5}\\
\dot{\mu}_{4}=-\frac{1}{\sqrt{2}}\mu_{2}\mu_{6},\\
\dot{\mu}_{5}=\frac{1}{\sqrt{2}}\mu_{1}\mu_{6},\\
\dot{\mu}_{6}=-\frac{1}{\sqrt{2}}\mu_{1}\mu_{5}+\frac{1}{\sqrt{2}}\mu_{2}\mu_{4}.\label{SE(3)_group_equation_one_particle}
\end{cases}
\end{align}
One verifies that $h$ is a conserved quantity, as we have
\begin{align*}
\dot{h}={}&\dot{\mu}_{4}+\mu_{1}\dot{\mu}_{1}+\mu_{2}\dot{\mu}_{2}=\\
&=-\frac{1}{\sqrt{2}}\mu_{2}\mu_{6}+\mu_{1}\left(-\frac{1}{\sqrt{2}}\mu_{2}\mu_{3}\right)
+\mu_{2}\left(\frac{1}{\sqrt{2}}\mu_{1}\mu_{3}+\frac{1}{\sqrt{2}}\mu_{6}\right)=0.
\end{align*}
Hence, the Hamiltonian and both Casimir functions are conserved quantities. No further simplification of equations is possible. 

However, if we were to consider a slightly different Hamiltonian with $q=6$ (drift in the 3rd linear momentum coordinate), so that $h=\mu_{6}+\frac{1}{2}\mu_{1}^{2}+\frac{1}{2}\mu_{2}^{2}$, then Lie-Poisson equations \eqref{LP_reduced_dynamics_equations} can be simplified and solved exactly for particular initial conditions. For this single particle problem the Lie--Poisson reduced dynamics equations $\dot{\breve{\mu}}=\Lambda(\breve{\mu})\nabla h$  \eqref{LP_reduced_dynamics_equations}, where $\nabla h=( \mu_{1} , \mu_{2} , 0 , 0 , 0 , 1)^T$ are given by
\begin{align*}
\frac{d}{dt}
\left[\begin{matrix}
{\mu}_{1}\\
{\mu}_{2}\\
{\mu}_{3}\\
{\mu}_{4}\\
{\mu}_{5}\\
{\mu}_{6}\\
\end{matrix}\right]=\frac{1}{\sqrt{2}}\left[\begin{matrix}
0&-\mu_{3}&\mu_{2}&0&-\mu_{6}&\mu_{5}\\
\mu_{3}&0&-\mu_{1}&\mu_{6}&0&-\mu_{4}\\
-\mu_{2}&\mu_{1}&0&-\mu_{5}&\mu_{4}&0\\
0&-\mu_{6}&\mu_{5}&0&0&0\\
\mu_{6}&0&-\mu_{4}&0&0&0\\
-\mu_{5}&\mu_{4}&0&0&0&0\end{matrix}\right]\left[\begin{matrix}
\mu_{1}\\
\mu_{2}\\
0\\
0\\
0\\
1\\\end{matrix}\right]
\end{align*}
or, equivalently,
\begin{align}
\begin{cases}
\dot{\mu}_{1}=-\frac{1}{\sqrt{2}}\mu_{2}\mu_{3}+\frac{1}{\sqrt{2}}\mu_{5}, \\
\dot{\mu}_{2}=\frac{1}{\sqrt{2}}\mu_{1}\mu_{3}-\frac{1}{\sqrt{2}}\mu_{4}, \\
\dot{\mu}_{3}=0\\
\dot{\mu}_{4}=-\frac{1}{\sqrt{2}}\mu_{2}\mu_{6},\\
\dot{\mu}_{5}=\frac{1}{\sqrt{2}}\mu_{1}\mu_{6},\\
\dot{\mu}_{6}=-\frac{1}{\sqrt{2}}\mu_{1}\mu_{5}+\frac{1}{\sqrt{2}}\mu_{2}\mu_{4}.\label{SE(3)_group_equation_one_particle_reduced}
\end{cases}
\end{align}
One verifies that $h$ is a conserved quantity, as we have
\begin{align*}
\dot{h}={}&\dot{\mu}_{6}+\mu_{1}\dot{\mu}_{1}+\mu_{2}\dot{\mu}_{2}=\\
&=-\frac{1}{\sqrt{2}}\mu_{1}\mu_{5}+\frac{1}{\sqrt{2}}\mu_{2}\mu_{4}+\mu_{1}\biggl(-\frac{1}{\sqrt{2}}\mu_{2}\mu_{3}+\frac{1}{\sqrt{2}}\mu_{5}\biggr)+\mu_{2}\biggl(\frac{1}{\sqrt{2}}\mu_{1}\mu_{3}-\frac{1}{\sqrt{2}}\mu_{4}\biggr)=0.
\end{align*}

From the third equation of \eqref{SE(3)_group_equation_one_particle_reduced} we obtain that $\mu_{3}=const$. Let us suppose the initial condition $\mu_{3}(0)=0$, that is, $\mu_{3}\equiv0$. In this case the system of equations \eqref{SE(3)_group_equation_one_particle} simpifies as follows:
\begin{align}
\begin{cases}
\dot{\mu}_{1}=\frac{1}{\sqrt{2}}\mu_{5}, \\
\dot{\mu}_{2}=-\frac{1}{\sqrt{2}}\mu_{4}, \\
\dot{\mu}_{3}=0 \quad (\mu_{3}=0),\\
\dot{\mu}_{4}=-\frac{1}{\sqrt{2}}\mu_{2}\mu_{6},\\
\dot{\mu}_{5}=\frac{1}{\sqrt{2}}\mu_{1}\mu_{6},\\
\dot{\mu}_{6}=-\frac{1}{\sqrt{2}}\mu_{2}\mu_{5}+\frac{1}{\sqrt{2}}\mu_{2}\mu_{4}.\label{SE(3)_group_equation_one_particle_mu(3)=0}
\end{cases}
\end{align}

Differentiating the fourth equation of \eqref{SE(3)_group_equation_one_particle_mu(3)=0} we obtain the second-order equation
\begin{align*}
\ddot{\mu}_{6}=-\frac{1}{2}\mu_{6}\left(\mu_{1}^{2}+\mu_{2}^{2}\right)-\frac{1}{2}\left(\mu_{4}^{2}+\mu_{5}^{2}\right).
\end{align*}
Substituting expressions for the Hamiltonian $h$ and the first Casimir $c_{1}$, we obtain:
\begin{gather}
\ddot{\mu}_{6}=-h\mu_{6}+\frac{3}{2}\mu_{6}^{2}-\frac{c_{1}}{2}.\label{equation1_for_mu(4)}
\end{gather}
Notice that $\frac{1}{2}(\dot{\mu}_{6}^{2})'=\dot{\mu}_{6}\ddot{\mu}_{6}$. Integration of this equality yields
\begin{gather*}
\frac{1}{2}(\dot{\mu}_{6}^{2})=\int \ddot{\mu}_{6}\dot{\mu}_{6} dt=\int \ddot{\mu}_{6}d \mu_{6}.
\end{gather*}
Substituting $\ddot{\mu}_{6}$ from \eqref{equation1_for_mu(4)} and performing the integration explicitly, we derive
\begin{gather}
\frac{1}{2}\dot{\mu}_{6}^{2}=-\frac{h\mu_{6}^{2}}{2}+\frac{\mu_{6}^{3}}{2}-\frac{c_{1}}{2}\mu_{6}+D,\label{equation2_for_mu(4)}
\end{gather}
where $D$ is an arbitrary constant which appears due to integration.

Our next goal is to perform a linear exchange of variables $\mu_{6}=Au+B$ with such constants $A$ and $B$ (to be determined below) that the equation \eqref{equation2_for_mu(4)}
will turn into
\begin{gather}
\dot{u}^{2}=4u^{3}-g_{2}u-g_{3}\label{general_equation}
\end{gather}
which can be solved using Weierstrass elliptic functions ($g_{2}$ and $g_{3}$ are some constants).
Performing the above mentioned exchange of variables, the equation \eqref{equation2_for_mu(4)} is rewritten as
\begin{gather}
\dot{u}^{2}=Au^{3}+(-h+3B)u^{2}+\left(-\frac{2hB}{A}+\frac{3B^{2}}{A}-\frac{c_{1}}{A}\right)u\\
+\left(-\frac{hB^{2}}{A^{2}}+\frac{B^{3}}{A^{2}}-\frac{c_{1}B}{A^{2}}+\frac{2D}{A^{2}}\right).\label{equation1_for_u}
\end{gather}
Comparing \eqref{general_equation} and \eqref{equation1_for_u} we see that we must have $A=4$ and $B=\frac{h}{3}$. That is, the desired exchange of variables is $\mu_{6}=4u+\frac{h}{3}$. Then the equation \eqref{equation1_for_u} takes the form
\begin{gather}
\dot{u}^{2}=4u^{3}-\biggl(\frac{h^{2}}{12}+\frac{c_{1}}{4}\biggr)u-\biggl(\frac{h^{3}}{216}+\frac{c_{1}h}{48}-\frac{D}{8}\biggr).\label{equation2_for_u}
\end{gather}
That is, comparing \eqref{general_equation} and \eqref{equation2_for_u}, we obtain constants $g_{2}=\frac{h^{2}}{12}+\frac{c_{1}}{4}$ and $g_{3}=\frac{h^{3}}{216}+\frac{c_{1}h}{48}-\frac{D}{8}$.

\section{Computations of gradients of Hamiltonians}
\label{sec:Hamiltonian_gradients}
\subsection{${\rm SO}(3)$ group}
\label{app_sec:SO3_exact_deriv}
\paragraph{Dictatorship}
In the case of 'Dictatorship', the Hamiltonian \eqref{Reduced_ham_explicit} becomes
\begin{align*}
h={}&\sum_{k=1}^{N}\mu_{k2}+\frac{1}{2}\biggl(\frac{1+2\chi}{1+2N\chi}\mu_{11}^{2}
+\frac{1+2N\chi+4\chi^{2}}{(1+2N\chi)(1+2\chi)}(\mu_{21}^{2}+\dots+\mu_{N1}^{2})  \nonumber \\
&  +2\frac{2\chi}{1+2N\chi}\mu_{11}(\mu_{21}+\dots+\mu_{N1})+2\frac{4\chi^{2}}{(1+2N\chi)(1+2\chi)}\sum_{i,j=2, i<j}^{N}\mu_{i1}\mu_{j1}\biggr).
\end{align*}
We remind the reader that the total gradient is split into the gradients with respect to the momenta $\bmu_k$ of each particle: 
\begin{gather*}
\nabla h=\frac{\partial h}{\partial \breve{\mu}}=\left[\begin{matrix}
\partial h/\partial \bmu_{1}\\
\partial h/\partial \bmu_{2}\\[-2pt]
\vdots\\
\partial h/\partial \bmu_{N}\\\end{matrix}\right]=\left[\begin{matrix}
(\nabla h)_{1}\\
(\nabla h)_{2}\\[-2pt]
\vdots\\
(\nabla h)_{N}\\\end{matrix}\right]
\end{gather*}
with the gradients expressed as 
\begin{gather*}
(\nabla h)_{k}=\frac{\partial h}{\partial \boldsymbol{\mu}_{k}}=\left[\begin{matrix}
\partial h/\partial \mu_{k1}\\
\partial h/\partial \mu_{k2}\\
\partial h/\partial \mu_{k3}\\\end{matrix}\right]=
\left[\begin{matrix}
\partial h/\partial \mu_{k1}\\
1\\
0\\\end{matrix}\right], \quad k=1,\dots, N.
\end{gather*}
The partial derivatives are computed explicitly as  
\begin{align*}
\partial h/\partial \mu_{11}&=\frac{1+2\chi}{1+2N\chi}\mu_{11}+\frac{2\chi}{1+2N\chi}\sum_{j=2}^N \mu_{j1},\\
\partial h/\partial\mu_{k1}&=\frac{2\chi}{1+2N\chi}\mu_{11}+\frac{1+2N\chi+4\chi^{2}}{(1+2N\chi)(1+2\chi)}\mu_{k1}
+\frac{4\chi^{2}}{(1+2N\chi)(1+2\chi)}\sum_{\substack{i=2\\ i\neq k}}^{N} \mu_{i1}
\end{align*}
with $k=2,\dots, N$.
\paragraph{Democracy} Similarly, we compute gradients for each particle as 
\begin{gather*}
(\nabla h)_{k}=\frac{\partial h}{\partial \boldsymbol{\mu}_{k}}=\left[\begin{matrix}
\partial h/\partial \mu_{k1}\\
\partial h/\partial \mu_{k2}\\
\partial h/\partial \mu_{k3}\\\end{matrix}\right]=
\left[\begin{matrix}
\partial h/\partial \mu_{k1}\\
1\\
0\\\end{matrix}\right], \quad k=1,\dots, N. 
\end{gather*}
We see that the derivatives with respect to $\mu_{k1}$ are computed as 
\begin{gather*}
\partial h/\partial \mu_{k1}=\frac{1+2\chi}{1+2N\chi}\mu_{k1}+\frac{2\chi}{1+2N\chi}\sum_{i=1, i\neq k}^{N}\mu_{i1} \quad \forall k=1,\dots, N.
\end{gather*}
\subsection{${\rm SE}(3)$ group}
\label{app_sec:SE3_exact_deriv}
\paragraph{Dictatorship}
The matrix $\Psi$ in \eqref{Reduced_ham_explicit} can be computed explicitly as 
\begin{gather*}
\Psi=(\mathbb{I}_{N}+2\chi B)^{-1}\otimes \mathbb{I}_{2}\\
=\left[\begin{matrix}
\frac{1+2\chi}{1+2N\chi}\!&\!0\!&\!\frac{2\chi}{1+2N\chi}\!&\!0\!&\dots&\!\frac{2\chi}{1+2N\chi}\!&\!0\\
0\!&\!\frac{1+2\chi}{1+2N\chi}\!&\!0\!&\!\frac{2\chi}{1+2N\chi}\!&\dots&\!0\!&\!\frac{2\chi}{1+2N\chi}\\
\frac{2\chi}{1+2N\chi}\!&\!0\!&\!\frac{1+2N\chi+4\chi^{2}}{(1+2N\chi)(1+2\chi)}\!&\!0\!&\!\dots&\!\frac{4\chi^{2}}{(1+2N\chi)(1+2\chi)}\!&\!0\\
0\!&\!\frac{2\chi}{1+2N\chi}\!&\!0&\!\frac{1+2N\chi+4\chi^{2}}{(1+2N\chi)(1+2\chi)}\!&\dots&\!0&\!\frac{4\chi^{2}}{(1+2N\chi)(1+2\chi)}\\
\dots&\dots&\dots&\dots&\dots&\dots&\dots\\
\frac{2\chi}{1+2N\chi}\!&\!0&\!\frac{4\chi^{2}}{(1+2N\chi)(1+2\chi)}\!&\!0\!&\dots&\!\frac{4\chi^{2}}{(1+2N\chi)(1+2\chi)}\!&\!0\\
0\!&\!\frac{2\chi}{1+2N\chi}\!&\!0\!&\!\frac{4\chi^{2}}{(1+2N\chi)(1+2\chi)}\!&\dots&\!0&\!\frac{4\chi^{2}}{(1+2N\chi)(1+2\chi)}\\
\frac{2\chi}{1+2N\chi}\!&\!0\!&\!\frac{4\chi^{2}}{(1+2N\chi)(1+2\chi)}\!&\!0\!&\dots&\!\frac{1+2N\chi+4\chi^{2}}{(1+2N\chi)(1+2\chi)}\!&\!0\\
0\!&\!\frac{2\chi}{1+2N\chi}\!&\!0\!&\!\frac{4\chi^{2}}{(1+2N\chi)(1+2\chi)}\!&\dots&\!0\!&\!\frac{1+2N\chi+4\chi^{2}}{(1+2N\chi)(1+2\chi)}\end{matrix}\right].
\end{gather*}
Using that expression, we arrive at \eqref{h_SE3_dictatorship}. 
\rem{ the Hamiltonian looks like
\begin{gather*}
h=\sum_{k=1}^{N}\mu_{k4}+\frac{1}{2}\biggl(\frac{1+2\chi}{1+2N\chi}\mu_{11}^{2}+\frac{1+2\chi}{1+2N\chi}\mu_{12}^{2}
+\frac{1+2N\chi+4\chi^{2}}{(1+2N\chi)(1+2\chi)}\mu_{21}^{2}+ \nonumber \\
+\frac{1+2N\chi+4\chi^{2}}{(1+2N\chi)(1+2\chi)}\mu_{22}^{2}
+\dots+\frac{1+2N\chi+4\chi^{2}}{(1+2N\chi)(1+2\chi)}\mu_{N1}^{2}+\frac{1+2N\chi+4\chi^{2}}{(1+2N\chi)(1+2\chi)}\mu_{N2}^{2}+ \nonumber \\
+2\frac{2\chi}{1+2N\chi}\mu_{11}(\mu_{21}+\dots+\mu_{N1}) 
+2\frac{2\chi}{1+2N\chi}\mu_{12}(\mu_{22}+\dots+\mu_{N2})+ \nonumber \\
2\frac{4\chi^{2}}{(1+2N\chi)(1+2\chi)}\sum_{i,j=2, i<j}^{N}\mu_{i1}\mu_{j1} +2\frac{4\chi^{2}}{(1+2N\chi)(1+2\chi)}\sum_{i,j=2, i<j}^{N}\mu_{i2}\mu_{j2}\biggr).
\end{gather*}
} 
Taking partial derivatives of \eqref{h_SE3_dictatorship} leads to 
\begin{gather*}
\nabla h=\frac{\partial h}{\partial \breve{\mu}}=\left[\begin{matrix}
\partial h/\partial \mu_{1}\\
\partial h/\partial \mu_{2}\\
\dots\\
\partial h/\partial \mu_{N}\\\end{matrix}\right]=\left[\begin{matrix}
(\nabla h)_{1}\\
(\nabla h)_{2}\\
\dots\\
(\nabla h)_{N}\\\end{matrix}\right]
\end{gather*}
with
\begin{gather*}
(\nabla h)_{k}=\frac{\partial h}{\partial \boldsymbol{\mu}_{k}}=\left[\begin{matrix}
\partial h/\partial \mu_{k1}\\
\partial h/\partial \mu_{k2}\\
\partial h/\partial \mu_{k3}\\
\partial h/\partial \mu_{k4}\\
\partial h/\partial \mu_{k5}\\
\partial h/\partial \mu_{k6}\\\end{matrix}\right]=
\left[\begin{matrix}
\partial h/\partial \mu_{k1}\\
\partial h/\partial \mu_{k2}\\
0\\
1\\
0\\
0\\\end{matrix}\right], \quad k=1,\dots, N.
\end{gather*}
We need to find the derivatives of $h$ with respect to $\mu_{k1}$ and $\mu_{k2}$. We start with $\mu_{k1}$. 
We see that
\begin{equation*}
\begin{aligned} 
\partial h/\partial \mu_{11} &=\frac{1+2\chi}{1+2N\chi}\mu_{11}+\frac{2\chi}{1+2N\chi}\left(\mu_{21}+\dots+\mu_{N1}\right); \nonumber \\
\partial h/\partial\mu_{k1}&=\frac{2\chi}{1+2N\chi}\mu_{11}+\frac{1+2N\chi+4\chi^{2}}{(1+2N\chi)(1+2\chi)}\mu_{k1}
+\frac{4\chi^{2}}{(1+2N\chi)(1+2\chi)}\sum_{i=2, i\neq k}^{N} \mu_{i1} \nonumber \\
& k=2,\dots, N.
\end{aligned} 
\end{equation*}
\paragraph{Democracy}
The matrix $\Psi$ can be again computed explicitly as 
\begin{gather*}
\Psi=(\mathbb{I}_{N}+2\chi B)^{-1}\otimes \mathbb{I}_{2}\\
=\left[\begin{matrix}
\frac{1+2\chi}{1+2N\chi}&0&\!\!\!\!\!\frac{2\chi}{1+2N\chi}\!\!\!\!\!&0&\!\!\!\!\!\dots&\frac{2\chi}{1+2N\chi}\!\!\!\!\!&0&\!\!\!\!\!\frac{2\chi}{1+2N\chi}\!\!\!\!\!&0\\
0&\!\!\!\!\!\frac{1+2\chi}{1+2N\chi}\!\!\!\!\!&\!\!\!\!\!0\!\!\!\!\!&\!\!\!\!\!\frac{2\chi}{1+2N\chi}\!\!\!\!\!&\!\!\!\!\!\dots\!\!\!\!\!&0&\!\!\!\!\!\frac{2\chi}{1+2N\chi}\!\!\!\!\!&0&\!\!\!\!\!\frac{2\chi}{1+2N\chi}\!\!\!\!\!\\
\frac{2\chi}{1+2N\chi}&0&\!\!\!\frac{1+2\chi}{1+2N\chi}\!\!\!&0&\!\!\!\dots&\!\!\!\!\!\frac{2\chi}{1+2N\chi}\!\!\!&0&\!\!\!\frac{2\chi}{1+2N\chi}\!\!\!&0\\
0&\!\!\!\!\!\frac{2\chi}{1+2N\chi}\!\!\!\!\!&0&\!\!\!\!\!\frac{1+2\chi}{1+2N\chi}\!\!\!\!\!&\dots&0&\!\!\!\!\!\frac{2\chi}{1+2N\chi}\!\!\!\!\!&0&\!\!\!\!\!\frac{2\chi}{1+2N\chi}\!\!\!\!\!\\
\dots&\dots&\dots&\dots&\dots&\dots&\dots&\dots&\dots\\
\frac{2\chi}{1+2N\chi}&0&\!\!\!\!\!\frac{2\chi}{1+2N\chi}\!\!\!\!\!&0&\!\!\!\!\!\dots\!\!\!\!\!&\!\!\!\!\!\frac{1+2\chi}{1+2N\chi}\!\!\!\!\!&0&\!\!\!\!\!\frac{2\chi}{1+2N\chi}\!\!\!\!\!&0\\
0&\!\!\!\!\!\frac{2\chi}{1+2N\chi}\!\!\!\!\!&\!\!\!\!\!0\!\!\!\!\!&\!\!\!\!\!\frac{2\chi}{1+2N\chi}\!\!\!\!\!&\!\!\!\!\!\dots\!\!\!\!\!&0&\!\!\!\!\!\frac{1+2\chi}{1+2N\chi}\!\!\!\!\!&0&\!\!\!\!\!\frac{2\chi}{1+2N\chi}\!\!\!\!\!\\
\frac{2\chi}{1+2N\chi}&0&\!\!\!\frac{2\chi}{1+2N\chi}\!\!\!&0&\!\!\!\dots&\!\!\!\!\!\frac{2\chi}{1+2N\chi}\!\!\!&0&\!\!\!\frac{1+2\chi}{1+2N\chi}\!\!\!&0\\
0&\!\!\!\!\!\frac{2\chi}{1+2N\chi}\!\!\!\!\!&0&\!\!\!\!\!\frac{2\chi}{1+2N\chi}\!\!\!\!\!&\dots&0&\!\!\!\!\!\frac{2\chi}{1+2N\chi}\!\!\!\!\!&0&\!\!\!\!\!\frac{1+2\chi}{1+2N\chi}\end{matrix}\right].
\end{gather*}
Using that matrix $\Psi$ and the expression \eqref{Reduced_ham_explicit}, we arrive to the expression for the Hamiltonian given by \eqref{h_SE3_Democracy}. We then compute the gradients of that Hamiltonian as follows: 
\begin{gather*}
\nabla h=\frac{\partial h}{\partial \breve{\mu}}=\left[\begin{matrix}
\partial h/\partial \mu_{1}\\
\partial h/\partial \mu_{2}\\
\dots\\
\partial h/\partial \mu_{N}\\\end{matrix}\right]=\left[\begin{matrix}
(\nabla h)_{1}\\
(\nabla h)_{2}\\
\dots\\
(\nabla h)_{N}\\\end{matrix}\right]
\end{gather*}

\begin{gather*}
h=\sum_{k=1}^{N}\mu_{k2}+\frac{1}{2}\left(\frac{1+2\chi}{1+2N\chi}\mu_{11}^{2}+\dots+\frac{1+2\chi}{1+2N\chi}\mu_{N1}^{2}+2\frac{2\chi}{1+2N\chi}\sum_{i,j=1, i<j}^{N}\mu_{i1}\mu_{j1}\right).
\end{gather*}
Hence
\begin{gather*}
\nabla h=\frac{\partial h}{\partial \breve{\mu}}=\left[\begin{matrix}
\partial h/\partial \mu_{1}\\
\partial h/\partial \mu_{2}\\
\dots\\
\partial h/\partial \mu_{N}\\\end{matrix}\right]=\left[\begin{matrix}
(\nabla h)_{1}\\
(\nabla h)_{2}\\
\dots\\
(\nabla h)_{N}\\\end{matrix}\right]\, , 
\end{gather*}
with the Hamiltonian given by the following explicit formulas:  
\begin{gather*}
h=\sum_{k=1}^{N}\mu_{k4}+\frac{1}{2}\biggl(\frac{1+2\chi}{1+2N\chi}\mu_{11}^{2}+\frac{1+2\chi}{1+2N\chi}\mu_{12}^{2}+\dots+\frac{1+2\chi}{1+2N\chi}\mu_{N1}^{2}+\frac{1+2\chi}{1+2N\chi}\mu_{N2}^{2} \nonumber \\
+2\frac{2\chi}{1+2N\chi}\sum_{i,j=1, i<j}^{N}\mu_{i1}\mu_{j1}+2\frac{2\chi}{1+2N\chi}\sum_{i,j=1, i<j}^{N}\mu_{i2}\mu_{j2}\biggr).
\end{gather*}
Hence
\begin{gather*}
\nabla h=\frac{\partial h}{\partial \breve{\mu}}=\left[\begin{matrix}
\partial h/\partial \bmu_{1}\\
\partial h/\partial \bmu_{2}\\
\dots\\
\partial h/\partial \bmu_{N}\\\end{matrix}\right]=\left[\begin{matrix}
(\nabla h)_{1}\\
(\nabla h)_{2}\\
\dots\\
(\nabla h)_{N}\\\end{matrix}\right]
\end{gather*}
We compute the gradients of $h$ with respect to $\mu_{k1}$ for each particle as 
\begin{gather*}
(\nabla h)_{k}=\frac{\partial h}{\partial \boldsymbol{\mu}_{k}}=\left[\begin{matrix}
\partial h/\partial \mu_{k1}\\
\partial h/\partial \mu_{k2}\\
\partial h/\partial \mu_{k3}\\
\partial h/\partial \mu_{k4}\\
\partial h/\partial \mu_{k5}\\
\partial h/\partial \mu_{k6}\\\end{matrix}\right]=
\left[\begin{matrix}
\partial h/\partial \mu_{k1}\\
\partial h/\partial \mu_{k2}\\
0\\
1\\
0\\
0\\\end{matrix}\right], \quad k=1,\dots, N.
\end{gather*}
We see that the derivative with respect to first component $\mu_{k1}$ for each particle is given by 
\begin{gather*}
\partial h/\partial \mu_{k1}=\frac{1+2\chi}{1+2N\chi}\mu_{k1}+\frac{2\chi}{1+2N\chi}\sum_{i=1, i\neq k}^{N}\mu_{i1} \quad \forall k=1,\dots, N.
\end{gather*}
\rem{ 
These formulas yield
\begin{gather*}
\left[\begin{matrix}
\partial h/\partial \mu_{11}\\
\partial h/\partial \mu_{21}\\
\dots\\
\partial h/\partial \mu_{N1}\\\end{matrix}\right]=(\mathbb{I}_{N}+2\chi B)^{-1}
\left[\begin{matrix}
\mu_{11}\\
\mu_{21}\\
\dots\\
\mu_{N1}\\\end{matrix}\right].
\end{gather*}
} 
The derivative with respect to second component $\mu_{k2}$ is given by 
\begin{gather*}
\partial h/\partial \mu_{k2}=\frac{1+2\chi}{1+2N\chi}\mu_{k2}+\frac{2\chi}{1+2N\chi}\sum_{i=1, i\neq k}^{N}\mu_{i2} \quad \forall k=1,\dots, N.
\end{gather*}
\rem{ 
These formulas yield
\begin{gather*}
\left[\begin{matrix}
\partial h/\partial \mu_{12}\\
\partial h/\partial \mu_{22}\\
\dots\\
\partial h/\partial \mu_{N2}\\\end{matrix}\right]=(\mathbb{I}_{N}+2\chi B)^{-1}
\left[\begin{matrix}
\mu_{12}\\
\mu_{22}\\
\dots\\
\mu_{N2}\\\end{matrix}\right].
\end{gather*}
} 
All in all, we see that the Hamiltonian $h$, and, hence, partial derivatives $\partial h/\partial\mu_{k1}$ and $\partial h/\partial\mu_{k2}$ are symmetric to each other with respect to the first and second components of all particles. This fact makes sense intuitively as well, since we suppose that control is applied equivalently in both components.
\section{Derivatives of the transformation matrices with respect to parameters}
\label{app_sec:Partial_deriv_transformations}
The partial derivatives $\frac{\partial \mathbb{R}(\mathbf{e}_{a}, w_{k_{0}a}t)}{\partial w_{k_{0}a}}, a \in {1,2,3}$ of matrices, which are Poisson transformations for ${\rm SO}(3)$ group:

\begin{gather*}
\frac{{\partial \mathbb{R}(\mathbf{e}_{1}, w_{k_{0}1}t)}}{\partial w_{k_{0}1}}=\left[\begin{matrix}
0&0&0\\
0&-t^{*}\sin(w_{k_{0}1}t^{*})&t^{*}\cos(w_{k_{0}1}t^{*})\\
0&-t^{*}\cos(w_{k_{0}1}t^{*})&-t^{*}\sin(w_{k_{0}1}t^{*})\\\end{matrix}\right];
\end{gather*}
\begin{gather*}
\frac{\partial \mathbb{R}(\mathbf{e}_{2}, w_{k_{0}2}t)}{\partial w_{k_{0}2}}=\left[\begin{matrix}
-t^{*}\sin(w_{k_{0}2}t^{*})&0&-t^{*}\cos(w_{k_{0}2}t^{*})\\
0&0&0\\
t^{*}\cos(w_{k_{0}2}t^{*})&0&-t^{*}\sin(w_{k_{0}2}t^{*})\\\end{matrix}\right];
\end{gather*}
\begin{gather*}
\frac{\partial \mathbb{R}(\mathbf{e}_{3}, w_{k_{0}3}t)}{\partial w_{k_{0}3}}=\left[\begin{matrix}
-t^{*}\sin(w_{k_{0}3}t^{*})&t^{*}\cos(w_{k_{0}3}t^{*})&0\\
-t^{*}\cos(w_{k_{0}3}t^{*})&-t^{*}\sin(w_{k_{0}3}t^{*})&0\\
0&0&0\\\end{matrix}\right].
\end{gather*}

The partial derivatives $\frac{\partial A(\mathbf{e}_{a}, w_{k_{0}a}t)}{\partial w_{k_{0}a}}, i \in {1, \dots, 6}$ of matrices, which are Poisson transformations for ${\rm SE}(3)$ group:
\begin{gather*}
\frac{\partial A(\mathbf{e}_{1}, w_{k_{0}1}t)}{\partial w_{k_{0}1}}=\left[\begin{matrix}
0&0&0&0&0&0\\
0&-t^{*}\sin(w_{k_{0}1}t^{*})&t^{*}\cos(w_{k_{0}1}t^{*})&0&0&0\\
0&-t^{*}\cos(w_{k_{0}1}t^{*})&-t^{*}\sin(w_{k_{0}1}t^{*})&0&0&0\\
0&0&0&0&0&0\\
0&0&0&0&-t^{*}\sin(w_{k_{0}1}t^{*})&t^{*}\cos(w_{k_{0}1}t^{*})\\
0&0&0&0&-t^{*}\cos(w_{k_{0}1}t^{*})&-t^{*}\sin(w_{k_{0}1}t^{*})\\\end{matrix}\right];
\end{gather*}
\begin{gather*}
\frac{\partial A(\mathbf{e}_{2}, w_{k_{0}2}t)}{\partial w_{k_{0}2}}=\left[\begin{matrix}
-t^{*}\sin(w_{k_{0}2}t^{*})&0&-t^{*}\cos(w_{k_{0}2}t^{*})&0&0&0\\
0&0&0&0&0&0\\
t^{*}\cos(w_{k_{0}2}t^{*})&0&-t^{*}\sin(w_{k_{0}2}t^{*})&0&0&0\\
0&0&0&-t^{*}\sin(w_{k_{0}2}t^{*})&0&-t^{*}\cos(w_{k_{0}2}t^{*})\\
0&0&0&0&0&0\\
0&0&0&t^{*}\cos(w_{k_{0}2}t^{*})&0&-t^{*}\sin(w_{k_{0}2}t^{*})\\\end{matrix}\right];
\end{gather*}
\begin{gather*}
\frac{\partial A(\mathbf{e}_{3}, w_{k_{0}3}t)}{\partial w_{k_{0}3}}=\left[\begin{matrix}
-t^{*}\sin(w_{k_{0}3}t^{*})&t^{*}\cos(w_{k_{0}3}t^{*})&0&0&0&0\\
-t^{*}\cos(w_{k_{0}3}t^{*})&-t^{*}\sin(w_{k_{0}3}t^{*})&0&0&0&0\\
0&0&0&0&0&0\\
0&0&0&-t^{*}\sin(w_{k_{0}3}t^{*})&t^{*}\cos(w_{k_{0}3}t^{*})&0\\
0&0&0&-t^{*}\cos(w_{k_{0}3}t^{*})&-t^{*}\sin(w_{k_{0}3}t^{*})&0\\0&0&0&0&0&0\\\end{matrix}\right];
\end{gather*}
Explicitly:
\begin{gather*}
A(\mathbf{e}_{4}, w_{k_{0}4}t)=\left[\begin{matrix}
1&0&0&0&0&0\\
0&1&0&0&0&w_{k_{0}4}t^{*}\\
0&0&1&0&-w_{k_{0}4}t^{*}&0\\
0&0&0&1&0&0\\
0&0&0&0&1&0\\
0&0&0&0&0&1\\\end{matrix}\right];
\end{gather*}
\begin{gather*}
A(\mathbf{e}_{5}, w_{k_{0}5}t)=\left[\begin{matrix}
1&0&0&0&0&-w_{k_{0}5}t^{*}\\
0&1&0&0&0&0\\
0&0&1&w_{k_{0}5}t^{*}&0&0\\
0&0&0&1&0&0\\
0&0&0&0&1&0\\
0&0&0&0&0&1\\\end{matrix}\right];
\end{gather*}
\begin{gather*}
A(\mathbf{e}_{6}, w_{k_{0}6}t)=\left[\begin{matrix}
1&0&0&0&w_{k_{0}6}t^{*}&0\\
0&1&0&-w_{k_{0}6}t^{*}&0&0\\
0&0&1&0&0&0\\
0&0&0&1&0&0\\
0&0&0&0&1&0\\
0&0&0&0&0&1\\\end{matrix}\right].
\end{gather*}
Then, the partial derivatives are:
\begin{gather*}
\frac{\partial A(\mathbf{e}_{4}, w_{k_{0}4}t)}{\partial w_{k_{0}4}}=\left[\begin{matrix}
0&0&0&0&0&0\\
0&0&0&0&0&t^{*}\\
0&0&0&0&-t^{*}&0\\
0&0&0&0&0&0\\
0&0&0&0&0&0\\
0&0&0&0&0&0\\\end{matrix}\right];
\end{gather*}
\begin{gather*}
\frac{\partial A(\mathbf{e}_{5}, w_{k_{0}5}t)}{\partial w_{k_{0}5}}=\left[\begin{matrix}
0&0&0&0&0&-t^{*}\\
0&0&0&0&0&0\\
0&0&0&t^{*}&0&0\\
0&0&0&0&0&0\\
0&0&0&0&0&0\\
0&0&0&0&0&0\\\end{matrix}\right];
\end{gather*}
\begin{gather*}
\frac{\partial A(\mathbf{e}_{6}, w_{k_{0}6}t)}{\partial w_{k_{0}6}}=\left[\begin{matrix}
0&0&0&0&t^{*}&0\\
0&0&0&-t^{*}&0&0\\
0&0&0&0&0&0\\
0&0&0&0&0&0\\
0&0&0&0&0&0\\
0&0&0&0&0&0\\\end{matrix}\right].
\end{gather*}

\section{Prompts for LLM}
\label{app:sec_LLM}
We used \emph{Gemini} Pro version, which provides a good combination of analytical skills and programming abilities. 
Our first prompt was as follows. 
\\  
\begin{verbatim}
Consider a set of N particles, each evolving on A Lie group G. We will take
three Lie groups: G =SE(2), SO(3), SE(3). Suppose there are N particles and
the dimension of G is n. Suppose Gamma_{ij}^s are the structure constants for
the corresponding Lie algebra of G. Take assembly of N vectors of length n,
we denote it mu_{ks}, where k is the particle number and s is the component
of the n-dimensional vector mu_k. For each k=1...N, define the nxn matrix
\widehat{mu}_k = \sum_s mu_{ks} \Gamma^s_{ij}, where (i,j) are in range 1...N. 
Now, stack all \widehat{mu}_k in the diagonal of a large Nn x Nn matrix, 
change the sign and divide by \sqrt{2}. This Lambda will be the Poisson matrix
for our Poisson bracket. Choose your own basis and structure function with the
same normalization for SE(2). The basis vectors should be normalized in trace
norm. Next, define the Hamiltonian h(mu_vec), where mu_vec consists of vectors 
mu_k stacked on top of each other. Define an N \times N adjacency matrix A.
Consider two cases of matrix A: dictatorship and Democracy. The dictatorship
case has the matrix A that has 1 in row 1 except for (1,1) element, and 1 in
column 1(except for 1,1 element) and 0 everywhere else. Democracy has 1
everywhere and 0 on diagonal. The Degree D matrix = a diagonal matrix, where
the sum of elements over a given column k is put on k-th element of diagonal.
So D_dictatorship = Diagonal (N-1,1,...,1) and
D_Democracy = Diagonal(N-1,N-1,...,N-1). 
Graph Laplacian B = D-A. For a given parameter \chi, define
Psi = (Id + \chi B)^{-1}.
For SE(2) and SO(3), define Psi1= Psi 
for SE(3), define Psi1 = Psi1 \otimes Identity(2x2) which is a 2N \times 2N
dimensional matrix which is obtained from matrix Psi as follows. Take the 1st
row of Psi and expand it to 2N by inserting 0 in all the even positions (i.e.
in 2,4,...). Next, take the same row and insert 0 in all the odd positions,
insert below the first expanded row. Repeat with all the rows of Psi (N times),
you will get 2N x 2N matrix Psi1. Now, consider the N-dimensional vector
mu_tilde consisting of parts of mu_vector: (mu_11, mu_21, ..., mu_N1) (i.e.,
N first component of all n-dimensional vectors mu_k). 
For SE(2) and SO(3) cases, define the Hamiltonian 
h = \sum_{k=1}^N \mu_{k2} + 1/2*\mu_tilde \cdot \Psi @ mu_tilde 
SE(3): 
For SE(3), take mu_tilde= (mu_11,mu_12, mu_21,mu_22, ..., mu_N1, mu_N2) (take
first two components of mu out of 6). Define the Hamiltonian 
h = \sum{k=1}^N \mu_{k4} + 1/2* \mu_tilde \cdot \Psi1 @ mu_tilde 
 
Now, write the equations of motion in the Poisson form 
\dot mu_vec =  \Lambda \nabla h, where \nabla is taken over all variables
mu_vec (there will be 0s and 1s in this gradient). Compute this gradient
analytically. 
 
Now, write a Python program using Numpy to compute N_trajectories=20
trajectories of this system in N_points=51, depending on the choice of
the group. 
 
Make a user switch controlling which group to make the computations for,
say switch = 0 -> SE(2), switch = 1 -> SO(3), switch = 2 -> SE(3).
Make another switch separating dictatorship from Democracy. 
 
Use a second order Lie-Poisson integrator to compute each trajectory for
each case (since the equations are Lie-Poisson). Use the integrator
preserving the Casimirs with machine precision. Choose the internal time
step for the integrator to be h/100. Output the values of the trajectories
each time point t_0,t_1,... t_{N_points} = 0,h,2*h,... (i.e. every h).
Choose h = 0.1 but keep it as a parameter. Plot the Hamiltonian and Casimir
for each case. 
 
For each trajectory, choose the first N_points-1 points as Begin_points,
and last N_points-1 as End_points. Stack the values for each trajectories.
Make Begin_points an array (N_points-1)*N_trajectories x n, the same for
End_points. 
 
Save these Begin_points, End_points in an .npz file with the name depending
on the case. 
 
Plot the Hamiltonian and Casimir invariants (one for SE(2) and SO(3), two for
SE(3) case) for each case for a single trajectory, and for each particle. 
\end{verbatim}
The LLM produced a running code, but the code did not conserve either the Hamiltonian or Casimirs. The case of the ${ \rm SE}(3)$ group took several attempts of pointing out that fact to the LLM, before it could find the error in the code. Interestingly, for quite some time, LLM claimed that large variations in energy were due to the non-compact nature of ${\rm SE}(3)$ group, before finding the error. 

Note that we have originally asked \emph{Gemini} to define $\Psi = (I+\chi B)^{-1}$, not $\Psi = (I+2 \chi B)^{-1}$ as in \cite{justh2010extremal,justh2015optimality} and also \eqref{matrix_PSI}.  We have then corrected the formula in the program by hand to match \eqref{matrix_PSI}.
The code produced by LLM was thoroughly checked by hand once the conservation of energy and Casimirs was done. 

The next prompt for generating the code was as follows. 

\begin{verbatim}
// Insert previous correct code here // 
    This program defines data structures and notations. Continue the files and
    notations as above. 

Now, write a Tensorflow program which does the following. 
It reads Begin_points and End_points according to the name convention defined
in the program above. It should determine the number of particles (from the
second dimension, by dividing by n) and the number of intervals (first
dimension of data array). 

Convert Begin_points and End_points arrays into appropriate TensorFlow data
structures. 

Set up a TensorFlow program defining a series of K neural networks with
input N*n vectors (n = dimension of the group, N = number of matrices)
and one output which is going to be a_k, k=1...K . Each network has depth
of hidden layer Depth = 1, and width W (take it to be int(N*n)/10). Let us
call this mapping A_k(\mu). 

Next, create a series of K transformations, where K = N*n, repeated P times
(take P=1 for now). Each transformation takes matrix V = (M,n) of data,
(where M is an arbitrary number >=1) and applies the transformation according
to SWITCH_GROUP (=0 -> SE(2), =1 -> SO(3) and =2 -> SE(3)), with a_k = A_k(\mu)
defined above, for a sequence k=1,2,...,K, with time step \Delta_t = 0.1,
equal to the Delta_t in the data. Neural network time step approximating
NN_step is the sequence of all K compositions. 

The loss function is defined as follows. Take V = Begin_points, apply the
sequence of transformations as defined above, and compare it to End_points
in the L2 norm. 

Now, optimize the loss function with respect to the parameters of K networks
using Adam algorithm, over Num_epochs = 1000.  Print optimization progress
every Num_epochs/10. Save the loss values during optimization and plot it
on the semiology scale. 

Next, use the found optimized parameters for prediction. Take a set of M = 10
random initial conditions in a cube of [-1,1]^n. Use Lie-Poisson integrator
defined above to obtain the ground truth by iterating the solutions from these
initial conditions over the time interval \Delta_t = 0.1 (with 100 subsets over
each step as above) over Num_points_evaluate = 100. Next, use the NN_step
approximations over the same number of time points starting with the same
initial conditions. 

Compare the solutions obtained by the neural network and Lie-Poisson
integrators. Compute the error in solution and compute the average error.
Next, compute the energy and Casimirs for Lie-Poisson integrator and NN_step
approximation. Plot the conservation of energy and Casimirs (2 plots for SE(2)
and SO(3) cases, 3 plots for SE(3) since there are two Casimirs for SE(3)). 
Plot the solutions for sample trajectories for all components of mu
(3*N in SE(2) and SO(3), 6*N for SE(3)). Finally, save the trajectories for
ground truth and NN solutions. Also save Casimirs and energy values in the
same file. Finally, save learned network parameters.
\end{verbatim}
Interestingly, LLM could produce a working code for CO-LPNets without substantial mistakes. While some work on the code was necessary, it could correctly use the Poisson tensors found in the previous case. The code was thoroughly checked 'by hand' to verify its correctness. 

\paragraph{Conclusion} LLMs are a great way to speed up writing code. However, one has to be extremely careful in applying programs written by LLMs. It seems that commercially available LLMs, at this point, are better at programming, while a combination of programming and analytic computation causes potential problems. These problems may appear randomly in different places, and may be contained in either computing the analytical results or actual programming errors. Thus, at this point, we can recommend the use of LLM only after the analytical derivation and the algorithm of the solution are understood. 
\end{document}